\documentclass[10pt,journal,compsoc]{IEEEtran}

\usepackage[utf8]{inputenc}

\usepackage{color}
\usepackage{epsfig}
\usepackage{graphicx}
\usepackage[table, svgnames]{xcolor}

\usepackage{adjustbox}
\usepackage{array}
\usepackage{booktabs}
\usepackage{colortbl}
\usepackage{float,wrapfig}
\usepackage{hhline}
\usepackage{multirow}
\usepackage[caption=false,font=normalsize,labelfont=sf,textfont=sf]{subfig}

\usepackage{amsmath,amsfonts,amsthm,amssymb}
\usepackage{bm}
\usepackage{nicefrac}
\usepackage{microtype}
\usepackage{inconsolata}

\usepackage{changepage}
\usepackage{extramarks}
\usepackage{fancyhdr}
\usepackage{lastpage}
\usepackage{setspace}
\usepackage{soul}
\usepackage{xspace}
\usepackage{paralist}
\usepackage{ragged2e}
\usepackage{tabu}

\usepackage{hyperref}
\usepackage[nocompress]{cite}
\usepackage{url}

\usepackage[ruled,vlined]{algorithm2e}
\usepackage{enumerate}

\usepackage{titlesec}
\usepackage{listings}
\usepackage{enumitem}
\usepackage{lipsum}
\newcolumntype{L}[1]{>{\raggedright\let\newline\\\arraybackslash\hspace{0pt}}m{#1}}
\newcolumntype{C}[1]{>{\centering\let\newline\\\arraybackslash\hspace{0pt}}m{#1}}
\newcolumntype{R}[1]{>{\raggedleft\let\newline\\\arraybackslash\hspace{0pt}}m{#1}}

\newcommand{\sect}[1]{Section~\ref{#1}}

\newcommand{\fig}[1]{Fig.~\ref{#1}}
\newcommand{\tab}[1]{Table~\ref{#1}}

\newcommand{\ignorethis}[1]{}
\newcommand{\norm}[1]{\lVert#1\rVert}

\makeatletter
\DeclareRobustCommand\onedot{\futurelet\@let@token\@onedot}
\def\@onedot{\ifx\@let@token.\else.\null\fi\xspace}

\def\eg{\emph{e.g}\onedot} 
\def\ie{\emph{i.e}\onedot} 
 
 \def\vs{\emph{vs}\onedot}
\def\wrt{w.r.t\onedot} 
\def\etal{\emph{et al}\onedot}
\makeatother

\definecolor{citecolor}{RGB}{34,139,34}
\definecolor{mydarkblue}{rgb}{0,0.08,1}
\definecolor{mydarkgreen}{rgb}{0.02,0.6,0.02}
\definecolor{mydarkred}{rgb}{0.8,0.02,0.02}
\definecolor{mydarkorange}{rgb}{0.40,0.2,0.02}
\definecolor{mypurple}{RGB}{111,0,255}
\definecolor{myred}{rgb}{1.0,0.0,0.0}
\definecolor{mygold}{rgb}{0.75,0.6,0.12}
\definecolor{mydarkgray}{rgb}{0.66, 0.66, 0.66}

\newcommand{\myparagraph}[1]{\textit{#1}}

\begin{document}

\title{PVNAS: 3D Neural Architecture Search \\ with Point-Voxel Convolution}

\author{
Zhijian Liu*, Haotian Tang*, Shengyu Zhao, Kevin Shao, and Song Han
\IEEEcompsocitemizethanks{
\IEEEcompsocthanksitem Z. Liu, H. Tang, S. Zhao, K. Shao, and S. Han are with Department of Electrical Engineering and Computer Science, Massachusetts Institute of Technology, Cambridge, MA 02139, USA. The first two authors contributed equally to this work.\\
E-mails: \texttt{\{zhijian,kentang,shengyuz,kshao23,songhan\}@mit.edu}.}
}

\markboth{IEEE TRANSACTIONS ON PATTERN ANALYSIS AND MACHINE INTELLIGENCE, VOL. X, NO. X, MMMMMMM YYYY}{Liu \MakeLowercase{\textit{et al.}}: PVNAS: 3D Neural Architecture Search with Point-Voxel Convolution}

\IEEEtitleabstractindextext{

\begin{abstract}
    
3D neural networks are widely used in real-world applications (\eg, AR/VR headsets, self-driving cars). They are required to be fast and accurate; however, limited hardware resources on edge devices make these requirements rather challenging. Previous work processes 3D data using either voxel-based or point-based neural networks, but both types of 3D models are not hardware-efficient due to the large memory footprint and random memory access. In this paper, we study 3D deep learning from the efficiency perspective. We first systematically analyze the bottlenecks of previous 3D methods. We then combine the best from point-based and voxel-based models together and propose a novel hardware-efficient 3D primitive, \emph{Point-Voxel Convolution (PVConv)}. We further enhance this primitive with the sparse convolution to make it more effective in processing large (outdoor) scenes. Based on our designed 3D primitive, we introduce \emph{3D Neural Architecture Search (3D-NAS)} to explore the best 3D network architecture given a resource constraint. We evaluate our proposed method on six representative benchmark datasets, achieving state-of-the-art performance with \textbf{1.8-23.7$\times$} measured speedup. Furthermore, our method has been deployed to the autonomous racing vehicle of MIT Driverless, achieving larger detection range, higher accuracy and lower latency.

\end{abstract}


}

\maketitle

\IEEEdisplaynontitleabstractindextext
\IEEEpeerreviewmaketitle

\IEEEraisesectionheading{\section{Introduction}}

\begin{figure*}[t]
\centering
\captionsetup[subfigure]{position=top}
\subfloat[Operation: Memory \vs Arithmetic]{\includegraphics[width=0.49\linewidth]{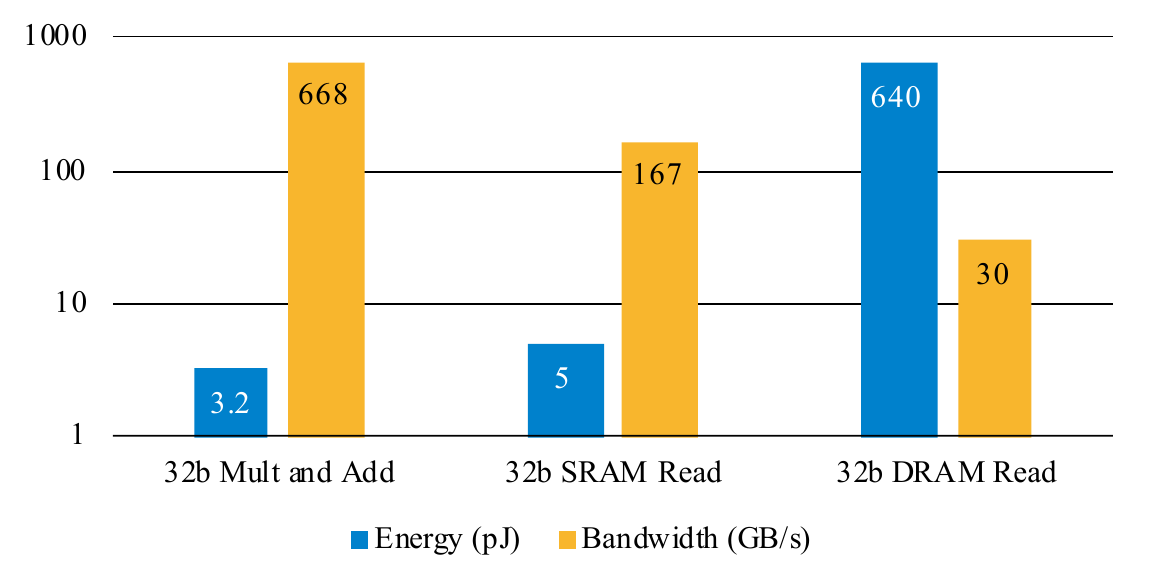}
\label{fig:teaser:a}}
\hfil
\subfloat[Memory Access: Random \vs Sequential]{\includegraphics[width=0.49\linewidth]{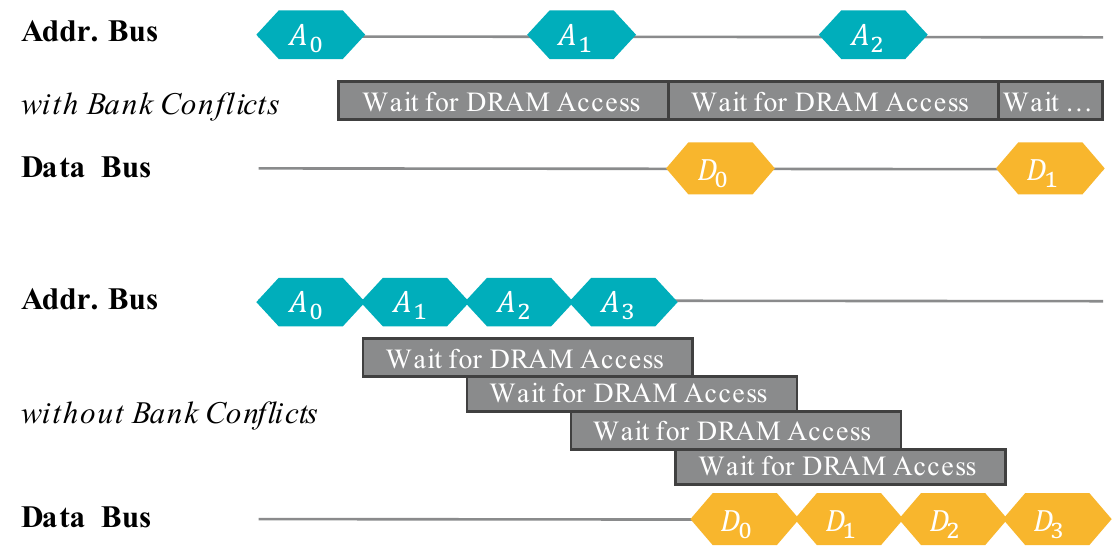}
\label{fig:teaser:b}}
\caption{Efficient 3D models should reduce memory footprint and avoid random memory accesses. \textbf{(a)} Off-chip DRAM accesses take two orders of magnitude more energy than arithmetic operations (640pJ \vs 3pJ), whereas the bandwidth is two orders of magnitude lower (30GB/s \vs 668GB/s). Efficient 3D model should \textbf{reduce the memory footprint}, which is the bottleneck of \textit{voxel-based} methods. \textbf{(b)} Random memory access is inefficient since it cannot take advantage of the DRAM burst and will cause bank conflicts, whereas contiguous memory access does not suffer from the above issue. Efficient 3D model should \textbf{avoid random memory accesses}, which is the bottleneck of \textit{point-based} methods.}
\label{fig:teaser}
\end{figure*}

\IEEEPARstart{3}{D} deep learning has received increased attention thanks to its wide applications. It has been applied in AR/VR headsets to understand the layout of indoor scenes; it has also been used in the LiDAR perception that serves as the eyes of autonomous driving systems to understand the semantics of outdoor scenes to parse the drivable area (\eg, roads, parking areas). These real-world applications require high accuracy and low latency at the same time: \ie, AR/VR headsets aim to offer an instant and accurate response for better user experience, and self-driving cars are expected to drive safely even at a relatively high speed. However, the computational resources on these devices are tightly constrained by the form factor (since we do not want a full backpack of hardware or a whole trunk of workstations) and heat dissipation. Thus, it is crucial to design efficient and effective 3D neural network models with limited hardware resources.

Collected by LiDAR sensors, 3D data usually comes in the format of point clouds. Conventionally, researchers rasterize the point cloud into voxel grids and process them using 3D volumetric convolutions~\cite{riegler2017octnet}. With low resolutions, there will be information loss during voxelization: multiple points will be merged together if they lie in the same grid. Therefore, a high-resolution representation is needed in order to preserve the fine details in the input data. However, the computational cost and memory requirement both increase \textit{cubically} with voxel resolution. Thus, it is infeasible to train a voxel-based model with high-resolution inputs: \eg, 3D-UNet~\cite{cicek20163d} requires more than 10 GB of GPU memory on 64$\times$64$\times$64 inputs with batch size of 16, and the large memory footprint makes it rather difficult to scale beyond this resolution.

Recently, another stream of models attempt to directly process the 3D point clouds~\cite{qi2017pointnet,qi2017pointnet++,klokov2017escape,li2018pointcnn}. These point-based models require much lower GPU memory than voxel-based models thanks to the sparse data representation. However, they neglect the fact that the \textit{random memory access} is also very inefficient. As the input points are scattered over the entire 3D space in a very irregular manner, processing them introduces many random memory accesses. Most point-based models~\cite{li2018pointcnn} mimic the 3D volumetric convolution: \ie, they compute the feature of each point by aggregating its neighboring features. However, neighbors are not stored contiguously in the point representation; therefore, indexing them requires the costly nearest neighbor search. To trade space for time, previous methods replicate the entire point cloud for each center point in the nearest neighbor search, and the memory cost will be $\mathcal{O}(n^2)$, where $n$ is the number of input points. Another very large overhead is introduced by the dynamic kernel computation. Since the relative positions of neighbors are not fixed, these point-based models have to generate the convolution kernels dynamically based on different offsets.

Designing efficient 3D neural networks needs to take the hardware into consideration. Compared with arithmetic operations, memory operations are much more expensive: \ie, they consume two orders of magnitude \textit{higher} energy and have two orders of magnitude \textit{lower} bandwidth (\fig{fig:teaser:a}). Another very important aspect is the memory access pattern: \ie, the random access will introduce memory bank conflicts and decrease the throughput (\fig{fig:teaser:b}). From the hardware perspective, conventional 3D models are inefficient due to large memory footprint and random memory access.

This paper studies 3D deep learning from the perspective of hardware efficiency. After analyzing the bottlenecks of previous methods, we introduce a novel hardware-efficient 3D primitive, \emph{Point-Voxel Convolution (PVConv)}, that brings the best from previous point-based and voxel-based models. It is composed of a fine-grained point-based branch that keeps the 3D data in high resolution without large memory footprint, and a coarse-grained voxel-based branch which aggregates the neighboring features without random memory accesses. However, as the resolution of its voxel-based branch is still constrained by the memory, this primitive is not effective in processing large scenes. To this end, we propose \emph{Sparse Point-Voxel Convolution (SPVConv)} that enhances our PVConv with the sparse convolution to enable higher resolutions in the voxel-based branch. Based on these efficient 3D primitives, we then propose \emph{3D Neural Architecture Search (3D-NAS)} to automatically explore the optimal 3D network architecture given a resource constraint.

As 3D deep learning has been used in various real-world scenarios (\eg, indoor scenes for AR/VR, and outdoor scenes for autonomous driving), we demonstrate the effectiveness of our proposed method on extensive benchmarks including 3D part segmentation (for objects), 3D semantic segmentation (for indoor and outdoor scenes) as well as 3D object detection (for outdoor scenes). Across all these datasets, our method consistently achieves the state-of-the-art performance with \textbf{1.8-23.7$\times$} measured speedup. Furthermore, our method has been deployed into the autonomous racing vehicle of MIT Driverless, achieving larger detection range, lower latency and higher accuracy. We hope that our research can bring inspirations to further explorations in this direction.
\begin{figure*}[t]
\centering
\captionsetup[subfigure]{position=top}
\subfloat[Voxel-Based Models: Memory Grows Cubically]{%
\includegraphics[width=0.49\linewidth]{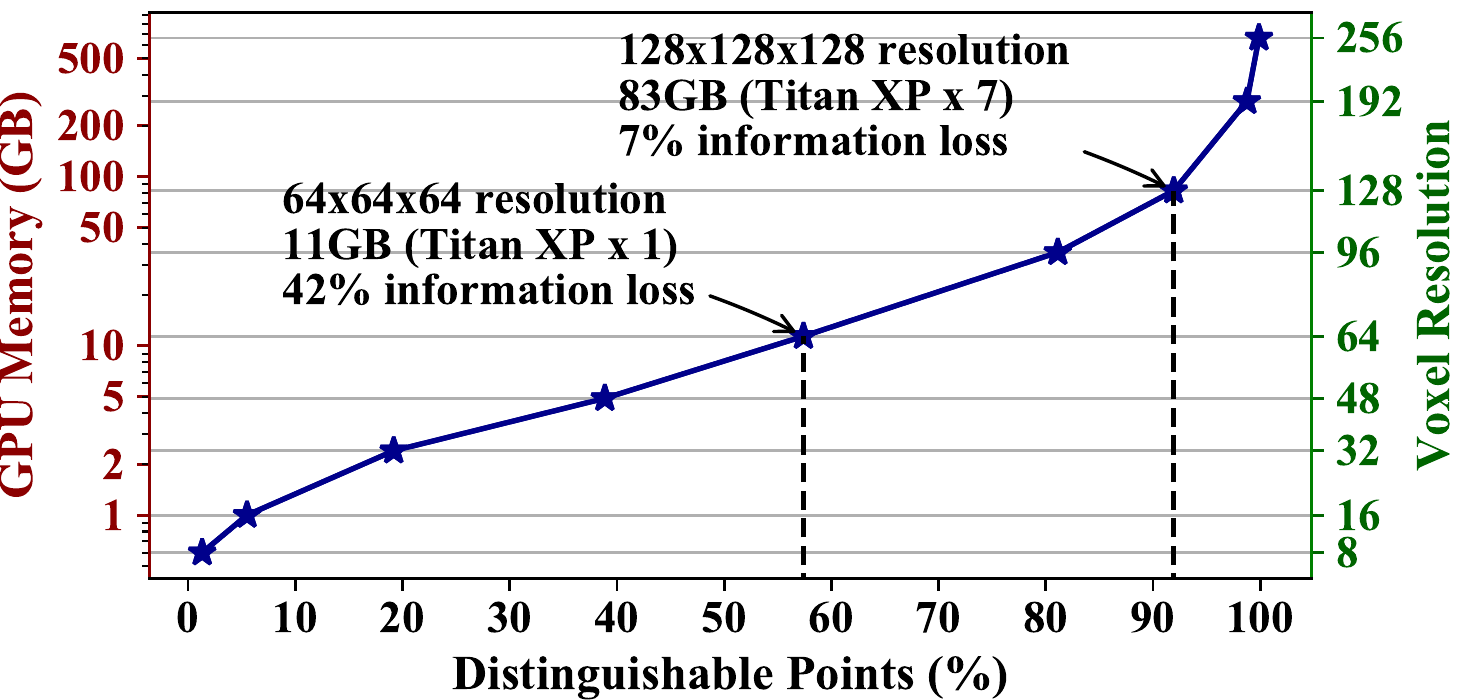}%
\label{fig:primitive:bottlenecks:a}}
\hfil
\subfloat[Point-Based Models: Large Memory Overheads]{%
\includegraphics[width=0.49\linewidth]{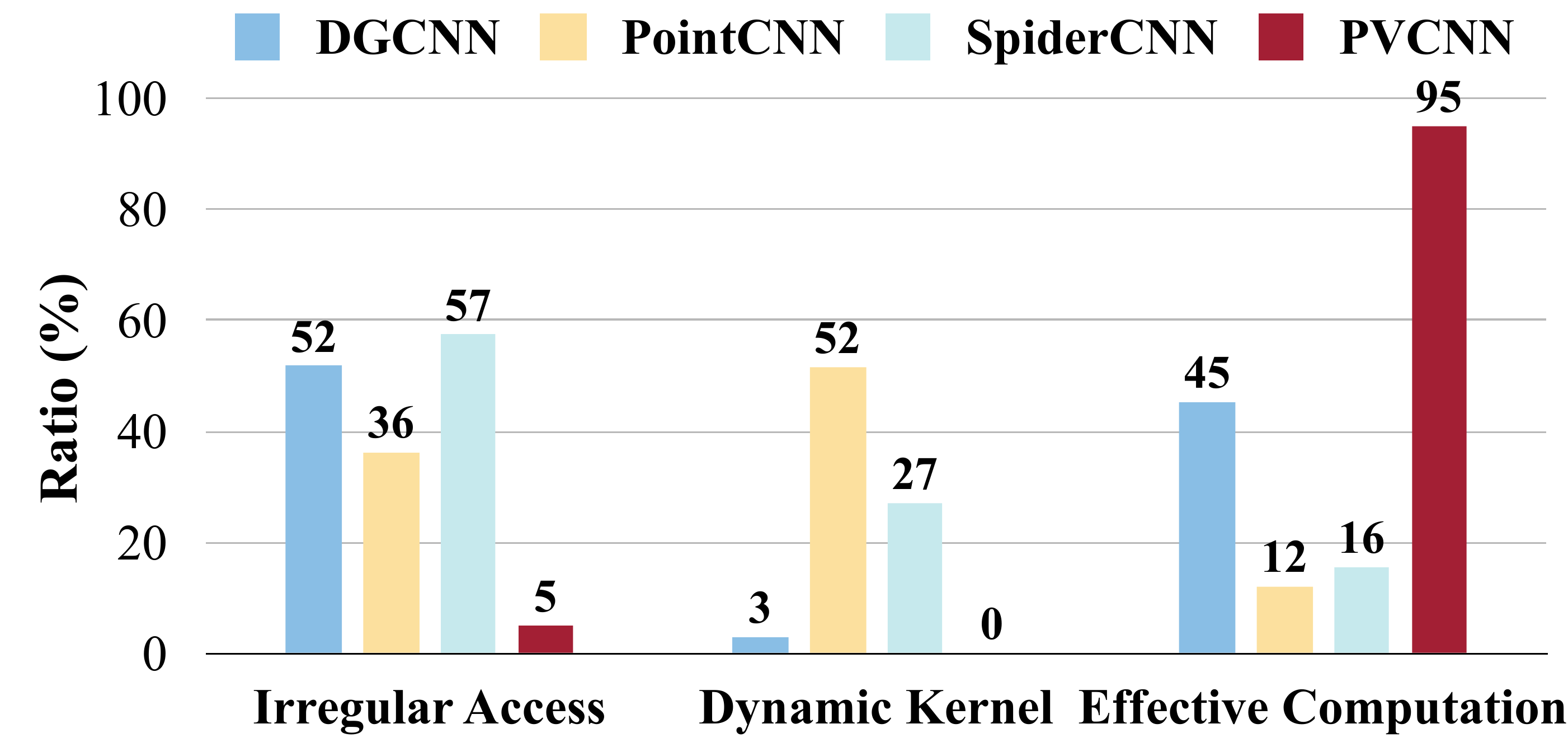}%
\label{fig:primitive:bottlenecks:b}}
\caption{Both conventional voxel-based and point-based models are inefficient. \textbf{(a)} Voxel-based models suffer from the large information loss at acceptable GPU memory consumption. \textbf{(b)} Point-based model suffer from large irregular memory access and dynamic kernel computation overheads.}
\label{fig:primitive:bottlenecks}
\end{figure*}

\section{Related Work}

\subsection{3D Neural Networks}

Conventionally, researchers relied on the volumetric representation and applied the convolution to process 3D data~\cite{wu20153d,chang2015shapenet,maturana2015voxnet,qi2016volumetric,wang2019voxsegnet,zhou2018voxelnet}. To name a few, Maturana~\etal~\cite{maturana2015voxnet} proposed the vanilla volumetric CNN; Qi~\etal~\cite{qi2016volumetric} extended 2D CNNs to 3D and systematically analyzed the relationship between 3D CNNs and multi-view CNNs; Wang~\etal~\cite{wang2017cnn} incorporated the octree into the volumetric CNN to reduce the memory consumption. Recent studies suggest that the volumetric representation can be used in 3D shape segmentation~\cite{tatarchenko2018tangent,wang2019voxsegnet,le2018pointgrid} and 3D object detection~\cite{zhou2018voxelnet} as well. Due to the sparse nature of 3D data, the dense volumetric representation is inherently inefficient and also inevitably introduces information loss.

PointNet~\cite{qi2017pointnet} takes advantage of the symmetric function to directly process the unordered point sets in 3D. Later research~\cite{qi2017pointnet++,klokov2017escape, wang2018dynamic} proposed to stack PointNets hierarchically to model neighborhood information and increase model capacity. Instead of stacking PointNets as basic blocks, another type of methods~\cite{li2018pointcnn,lan2019modeling,xu2018spidercnn} abstract away the symmetric function using dynamically generated convolution kernels or learned neighborhood permutation function. Some other research, such as SPLATNet~\cite{su2018splatnet} that naturally extends the 2D image SPLAT to 3D, and SONet~\cite{li2018so} which uses the self-organization mechanism with the theoretical guarantee of invariance to point order, also show great potential in general-purpose 3D modeling with point clouds as input. Apart from these general-purpose models, there are also some attempts for specific 3D tasks. SegCloud~\cite{tchapmi2017segcloud}, SGPN~\cite{wang2018sgpn}, SPGraph~\cite{landrieu2018large}, ParamConv~\cite{wang2018deep}, SSCN~\cite{graham20183d} and RSNet~\cite{huang2018recurrent} are specialized for 3D semantic and instance segmentation. As for 3D object detection, F-PointNet~\cite{qi2018frustum} is based on the RGB detector and point-based regional proposal networks; PointRCNN~\cite{shi2019pointrcnn} follows the similar idea while removing the RGB detector.

Recently, researchers started to pay attention to the efficiency of 3D models. Hu~\etal~\cite{hu2020rand} proposed to aggressively downsample the point cloud to reduce the computation cost. Riegler~\etal~\cite{riegler2017octnet}, Wang~\etal~\cite{wang2017cnn,wang2018adaptive} and later Lei~\etal~\cite{lei2019octree} proposed to reduce the memory of volumetric representation using octrees where areas with lower density occupy fewer voxel grids. Graham~\etal~\cite{graham20183d} and Choy~\etal~\cite{choy20194d} proposed the sparse convolution to accelerate the vanilla volumetric convolution by keeping the activation sparse and skipping the computations in the inactive regions. However, all these methods still require 
considerable random memory accesses, which are very inefficient.

\subsection{Hardware-Efficient Deep Learning}

Extensive attention has been paid to hardware-efficient deep learning for real-world applications. For instance, researchers have proposed to reduce the memory access cost by pruning and quantizing the models~\cite{han2015learning,han2016deep,he2018amc,zhou2017incremental,wang2019haq} or directly designing the compact models~\cite{iandola2016squeezenet,howard2017mobilenets,sandler2018mobilenetv2,howard2019searching,zhang2018shufflenet,ma2018shufflenet}. These approaches are general-purpose and suitable for any neural networks. In this paper, we instead accelerate 3D neural networks based on some domain-specific properties: \eg, 3D point clouds are highly sparse and spatially structured.

\subsection{Neural Architecture Search}

To alleviate the burden of manually designing neural networks~\cite{howard2017mobilenets,sandler2018mobilenetv2,ma2018shufflenet,zhang2018shufflenet,iandola2016squeezenet}, researchers have introduced neural architecture search (NAS) to automatically architect the neural network with high accuracy using reinforcement learning~\cite{zoph2017neural,zoph2018learning} and evolutionary search~\cite{liu2019progressive}. A new wave of research started to design efficient models with neural architecture search~\cite{tan2019mnasnet,wu2019fbnet,tan2019efficientnet,wang2020hat} for edge deployment. However, conventional frameworks require high computation cost and considerable carbon footprint~\cite{strubell2019energy}. In order to tackle these, researchers have introduced different techniques to reduce the search cost, including differentiable architecture search~\cite{liu2019darts}, path-level binarization~\cite{cai2019proxylessnas}, single-path one-shot sampling~\cite{guo2019single,chen2019detnas,cai2020once}, and weight sharing~\cite{stamoulis2019single,cai2020once,wang2020hat}. Furthermore, neural architecture search has also been used in compressing and accelerating neural networks, including pruning~\cite{he2018amc,yang2018netadapt,liu2019metapruning,cai2019automl,li2020gan} and quantization~\cite{wang2019haq,guo2019single,wang2020hardware,wang2020apq}. Most of these methods are tailored for 2D visual recognition, which has many well-defined search spaces~\cite{radosavovic2019on}. Lately, researchers have applied neural architecture search to 3D medical image segmentation~\cite{zhu2019vnas,kim2019scalable,yang2019searching,bae2019resource,wong2019segnas3d,yu2020c2fnas} as well as 3D shape classification~\cite{ma2020auto,li2020sgas}. However, they are not directly applicable to 3D scene understanding since 3D medical data are still in the similar format as 2D images (which are entirely different from 3D scenes), and 3D objects are of much smaller scales than 3D scenes (which makes them less sensitive to the resolution).

\section{Analysis of Efficiency Bottlenecks}

3D data usually comes in the format of point clouds:
\begin{equation}
    \bm{x} = \{\bm{x}_k\} = \{(\bm{p}_k, \bm{f}_k)\},
\end{equation}
where $\bm{p}_k$ is the coordinate of the $k$\textsuperscript{th} point, and $\bm{f}_k$ is the feature corresponding to $\bm{p}_k$. The voxelized representation can also be unified into this formulation, where $\bm{p}_k$ stands for the coordinate of the $k$\textsuperscript{th} voxel grid. Based on this, voxel-based and point-based convolution can be formulated as
\begin{equation}
    \bm{y}_k = \sum_{\bm{x}_i \in \mathcal{N}(\bm{x}_k)} \mathcal{K}(\bm{x}_k, \bm{x}_i) \times \mathcal{F}(\bm{x}_i).
\end{equation}
During the convolution, we iterate $\bm{x}_k$ over the entire input. For each center $\bm{x}_k$, we first index its neighbors $\bm{x}_i$ in $\mathcal{N}(\bm{x}_k)$, then convolve the neighboring features $\mathcal{F}(\bm{x}_i)$ with the kernel $\mathcal{K}(\bm{x}_k, \bm{x}_i)$, and produces the corresponding output $\bm{y}_k$.

\subsection{Voxel-Based Models: Large Memory Footprint}

Conventionally, researchers rasterize the point cloud into voxel grids and process them with 3D volumetric convolutions~\cite{riegler2017octnet}. Voxel-based representation is regular and has good memory locality. However, it requires very high resolution in order not to lose much information. When the resolution is low, multiple points are bucketed into the same voxel grid, and these points will no longer be \emph{distinguishable}. A point is kept only when it exclusively occupies one voxel grid. In \fig{fig:primitive:bottlenecks:a}, we investigate the number of distinguishable points and the memory consumption (during training with batch size of 16) with different resolutions. On a single GPU (with 12 GB of memory), the largest affordable resolution is 64, which will lead to \textbf{42\%} of information loss. To keep more than 90\% of the information, we then need to double the resolution to 128, consuming 7.2$\times$ GPU memory (\textbf{82.6 GB}), which is prohibitive in constrained scenarios. Although the GPU memory increases cubically with the resolution, the number of distinguishable points has a diminishing return. Therefore, the voxel-based solution is not scalable.

\subsection{Point-Based Models: Sparse Data Organization}

Recently, another stream of models process the point cloud directly~\cite{qi2017pointnet,qi2017pointnet++,li2018pointcnn,wang2018dynamic,xu2018spidercnn}. These point-based models require much lower GPU memory than voxel-based models thanks to the sparse representation. Among them, PointNet~\cite{qi2017pointnet} is also computation efficient, but it lacks the local context modeling capability. All the other models~\cite{qi2017pointnet++,li2018pointcnn,wang2018dynamic,xu2018spidercnn} improve the expressiveness of PointNet by aggregating the neighborhood information in the point-based domain. However, this will lead to the irregular memory access pattern and introduce the dynamic kernel computation overhead, which becomes the efficiency bottleneck.

\myparagraph{Irregular Memory Access.}
Different from the voxel-based representation, neighboring points $\bm{x}_i \in \mathcal{N}(\bm{x}_k)$ in the point-based representation will not be laid out contiguously in the memory. Besides, 3D points are scattered in $\mathbb{R}^3$; therefore, we need to explicitly identify who are in the neighboring set $\mathcal{N}(\bm{x}_k)$, rather than by direct indexing. Point-based methods often define $\mathcal{N}(\bm{x}_k)$ as nearest neighbors in the coordinate space~\cite{li2018pointcnn,xu2018spidercnn} or the feature space~\cite{wang2018dynamic}. Either requires explicit and expensive KNN computation. After KNN, gathering all neighbors $\bm{x}_i$ in $\mathcal{N}(\bm{x}_k)$ will require large amount of random memory accesses, which is not cache friendly. Combining the cost of neighbor indexing and data movement, the state-of-the-art point-based models spend \textbf{36}\%~\cite{li2018pointcnn}, \textbf{52}\%~\cite{wang2018dynamic} and \textbf{57}\%~\cite{xu2018spidercnn} of the total runtime on structuring the irregular data and accessing the memory randomly (\fig{fig:primitive:bottlenecks:b}).

\myparagraph{Dynamic Kernel Computation.}
For 3D volumetric convolutions, the kernel value $\mathcal{K}(\bm{x}_k, \bm{x}_i)$ can be directly indexed because the relative positions of the neighbor $\bm{x}_i$ are fixed for different center $\bm{x}_k$: \eg, each axis of the offset $\bm{p}_i - \bm{p}_k$ can only be 0, $\pm$1 for the convolution with size of 3. For the point-based convolution, the points are scattered over the entire 3D space irregularly; therefore, the relative positions of neighbors become unpredictable. In this case, we will have to calculate the kernel $\mathcal{K}(\bm{x}_k, \bm{x}_i)$ for each neighbor $\bm{x}_i$ \textit{on the fly}. For instance, SpiderCNN~\cite{xu2018spidercnn} leverages the third-order Taylor expansion as a continuous approximation of the kernel $\mathcal{K}(\bm{x}_k, \bm{x}_i)$; PointCNN~\cite{li2018pointcnn} permutes the neighboring points into a canonical order with the feature transformer $\mathcal{F}(\bm{x}_i)$. Both will introduce additional matrix multiplications. Empirically, the overhead of dynamic kernel computation for PointCNN can be more than \textbf{50}\% (\fig{fig:primitive:bottlenecks:b}).

The combined overhead of irregular memory access and dynamic kernel computation ranges from \textbf{55}\% (DGCNN) to \textbf{88}\% (PointCNN). Thus, most computations are wasted on dealing with the irregularity of point-based representation.
\section{Designing Efficient 3D Primitives}

\begin{figure}[t]
\centering
\includegraphics[width=\linewidth]{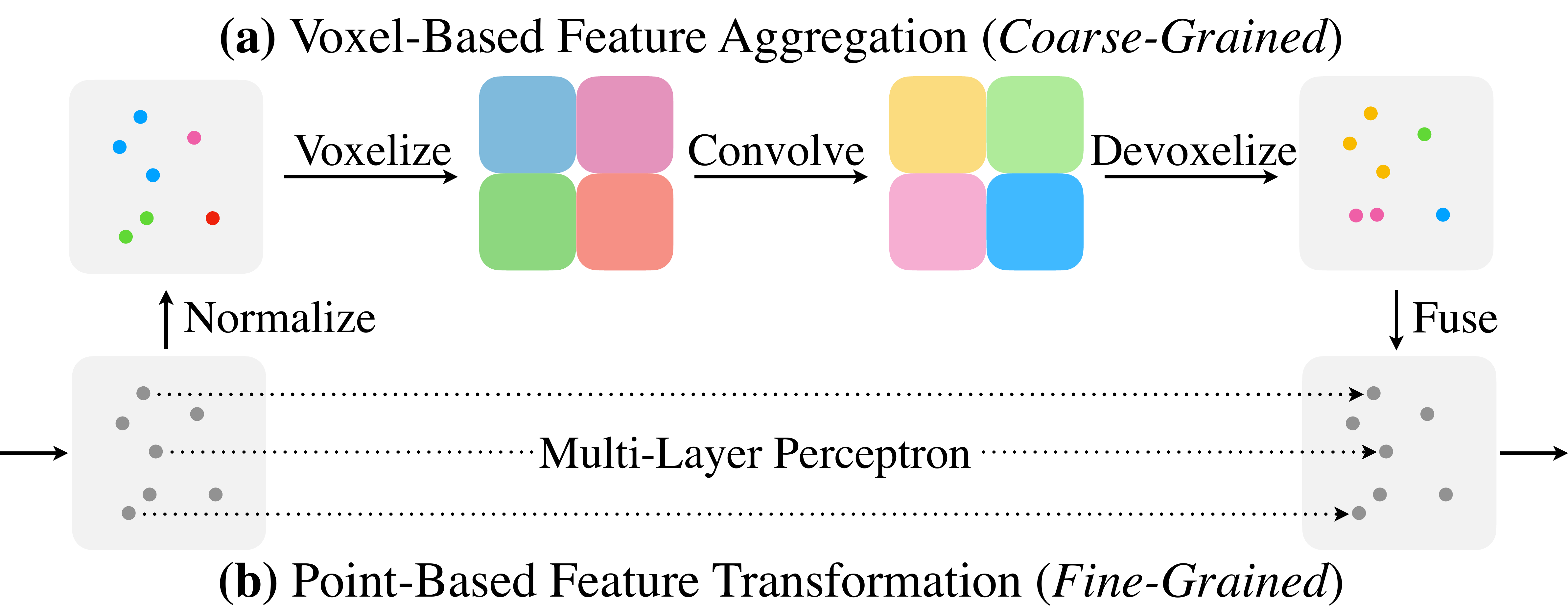}
\caption{Point-Voxel Convolution (PVConv) is composed of a \emph{low-resolution} voxel-based branch and a \emph{high-resolution} point-based branch. The voxel-based branch extracts \emph{coarse-grained} neighborhood information, which is supplemented by \emph{fine-grained} individual point features extracted from the point-based branch.}
\label{fig:primitive:pvconv}
\end{figure}

Based on our analysis of efficiency bottlenecks, we introduce a novel hardware-efficient 3D primitive for small 3D objects as well as indoor scenes: \emph{Point-Voxel Convolution (PVConv)}. It combines the advantages of point-based methods (small memory footprint) and voxel-based methods (good data locality and regularity). For large outdoor scenes, we further propose \emph{Sparse Point-Voxel Convolution (SPVConv)} that enhances PVConv with the sparse convolution to enable higher resolutions in the voxel-based branch.

\subsection{Point-Voxel Convolution (PVConv)}

Our PVConv disentangles the \emph{fine-grained} feature transformation and the \emph{coarse-grained} neighbor aggregation so that each branch can be implemented very efficiently and effectively. As in \fig{fig:primitive:pvconv}, the upper voxel-based branch first transforms the points into \emph{low-resolution} voxel grids, then it aggregates the neighboring points with voxel-based convolutions, followed by the devoxelization to convert them back to points. Either voxelization or devoxelization requires a single scan over all points, making the memory cost low. The lower point-based branch extracts the features for each individual point. As it does not aggregate the neighbor's information, it is able to afford a very \emph{high resolution}.

\subsubsection{Voxel-Based Feature Aggregation}

A key component of convolution is to aggregate the neighboring information in order to extract the local features. We choose to perform this feature aggregation in the volumetric domain due to its regularity.

\myparagraph{Normalization.}
The scale of different point clouds might be different. Thus, we normalize the coordinates $\{\bm{p}_k\}$ before converting the point cloud into the volumetric domain. First, we translate all points into the local coordinate system with the gravity center as the origin. After that, we normalize the points into the unit sphere by dividing all coordinates by $\max\norm{\bm{p}_k}_2$, and then scale and translate the points to $[0, 1]$. We denote the normalized coordinates as $\{\hat{\bm{p}}_k\}$.

\myparagraph{Voxelization.}
We transform the normalized point cloud $\{(\hat{\bm{p}}_k, \bm{f}_k)\}$ into the dense voxelized representation $\{\bm{V}_{u, v, w}\}$ by averaging all of the features $\bm{f}_k$ whose coordinate $\hat{\bm{p}}_k = (\hat{\bm{x}}_k, \hat{\bm{y}}_k, \hat{\bm{z}}_k)$ falls into the voxel grid $(u, v, w)$:
\begin{align}
  \begin{split}
    \bm{V}_{u, v, w, c} = \sum_{k=1}^n ~&\mathbb{I}[\lfloor \hat{\bm{x}}_k r \rfloor = u, \lfloor \hat{\bm{y}}_k r \rfloor = v, \lfloor \hat{\bm{z}}_k r \rfloor = w] \\ &\times \bm{f}_{k,c} / N_{u, v, w},
  \end{split}
\end{align}
where $r$ denotes the voxel resolution, $\mathbb{I}[\cdot]$ is the binary indicator of whether the coordinate $\hat{\bm{p}}_k$ belongs to the voxel grid $(u, v, w)$, $\bm{f}_{k,c}$ denotes the $c$\textsuperscript{th} channel feature corresponding to $\hat{\bm{p}}_k$, and $N_{u,v,w}$ is the normalization factor (\ie, the number of points that fall in that voxel grid).

\myparagraph{Feature Aggregation.}
After converting the points into the voxel grids, we apply a stack of 3D volumetric convolutions to aggregate the features. Similar to conventional 3D models, we apply batch normalization~\cite{ioffe2015batch} and nonlinear activation function~\cite{maas2013rectifier} after each 3D volumetric convolution.

\myparagraph{Devoxelization.}
As we need to fuse the information with the point-based feature transformation branch, we transform the voxel-based features back to the domain of point cloud. A simple implementation of the voxel-to-point mapping is the nearest-neighbor interpolation (\ie, assign the feature of a grid to all points in it). However, this implementation will make the points in the same voxel grid always share the same feature values. Therefore, we instead leverage the trilinear interpolation to transform the voxel grids to points to make sure that the features mapped to each point are distinct.

\subsubsection{Point-Based Feature Transformation}

The voxel-based feature aggregation branch fuses the neighborhood information in a coarse granularity. However, in order to model finer-grained individual point features, low-resolution voxel-based methods alone might not be sufficient. Hence, we directly operate on each point to extract individual point features using an MLP. Though simple, the MLP outputs distinct and discriminative features for each point. Such high-resolution individual point information is very critical to supplement the coarse-grained voxel-based information. With individual point features and aggregated neighborhood information, we can fuse two branches efficiently with an addition as they are providing complementary information.

\subsubsection{Analysis}

PVConv is much better than previous voxel-based and point-based models in terms of both efficiency and effectiveness.

\myparagraph{Better Locality and Regularity.}
Our PVConv is more efficient than conventional point-based convolutions due to its better data locality and regularity. Our voxelization and devoxelization both require only $\mathcal{O}(n)$ random memory accesses, where $n$ is the number of points, since we only need to iterate over all points once to scatter them to their corresponding grids. However, for conventional point-based methods, gathering neighbors for all points will require at least $\mathcal{O}(kn)$ random memory accesses, where $k$ is the number of neighbors. Thus, our PVConv is $k\times$ more efficient from this viewpoint. As the typical value for $k$ is 32/64 in PointNet++~\cite{qi2017pointnet++} and 16 in PointCNN~\cite{li2018pointcnn}, PVConv empirically reduces the number of incontiguous memory accesses by 16-64$\times$, achieving better data locality. Besides, as our convolutions are done in the voxel domain, which is regular, our PVConv does not require KNN computation and dynamic kernel computation, which are usually quite expensive.

\myparagraph{Higher Resolution.}
As our point-based feature extraction branch is implemented as MLP, a natural advantage is that we are able to maintain the same number of points throughout the whole network while still having the capability to model neighborhood information. Let us make a comparison between our PVConv and the set abstraction (SA) module in PointNet++~\cite{qi2017pointnet++}. Suppose we have a batch of 2048 points with 64-channel features (with batch size of 16), and we then aggregate information from 125 neighbors of each point and transform the aggregated feature to output the features with the same size. In this case, the SA module requires 75.2 ms of latency and 3.6 GB of memory, while our PVConv only requires 25.7 ms of measured latency and 1.0 GB of memory. The SA module will have to downsample to 685 points (\ie, around 3$\times$ downsampling) to match up with the latency of our PVConv, while the memory consumption will still be 1.5$\times$ higher. Therefore, with the same latency, our PVConv is capable of modeling the full point cloud, while the SA module has to downsample the input aggressively, which will inevitably induce information loss.

\subsection{Sparse Point-Voxel Convolution (SPVConv)}

PVConv is very efficient and effective especially for small 3D objects and indoor scenes as their scales are fairly small; however, it is less suitable for large outdoor scenes due to the coarse voxelization. PVConv-based model can afford the resolution of at most 128 in its voxel-based branch on a single GPU (with 12 GB of memory). For a large outdoor scene (with size of 100m$\times$100m$\times$10m), each voxel grid will correspond to a large area (with size of 0.8m$\times$0.8m$\times$0.1m). In this case, small instances (\eg, pedestrians) will only occupy very few voxel grids. From such few points, PVConv can hardly learn any useful information from the voxel-based branch, leading to a relatively low performance (see \tab{tab:primitive:limitation}). Alternatively, we can process the large 3D scenes piece by piece so that each sliding window is of smaller scale. In order to preserve the fine-grained information, we found empirically that the voxel size needs to be lower than 0.05m. In this case, we have to run the model once for each of the 244 sliding windows, which will take 35 seconds to process a single scene. Such a large latency is not affordable for most real-time applications (\eg, autonomous driving).

\begin{table}[!t]
\renewcommand{\arraystretch}{1.3}
\setlength{\tabcolsep}{4pt}
\small\centering

\caption{Comparison between PVConv and SPVConv in Large Outdoor Scenes}
\label{tab:primitive:limitation}

\vspace{-6pt}
\scalebox{0.85}{
\begin{tabular}{lcccc}
\toprule
& Input & Voxel Size (m) & Latency (ms) & Mean IoU \\
\midrule
\multirow{2}{*}{PVConv} & Sliding Window & 0.05 & 35640 & -- \\
& Entire Scene & 0.80 & 146 & 39.0 \\
\midrule
SPVConv & Entire Scene & 0.05 & \textbf{85} & \textbf{58.8} \\
\bottomrule
\end{tabular}
}

\vspace{8pt}\justifying\noindent
{\footnotesize\itshape{PVConv is not suitable for large scenes. If processing with sliding windows, its latency is not affordable for deployment. If taking the whole scene, its resolution is too coarse to capture useful information.}}
\end{table}
\begin{figure*}[t]
\centering
\captionsetup[subfigure]{position=top}
\subfloat[Stage 1: Super Network Training]{\includegraphics[width=0.49\linewidth]{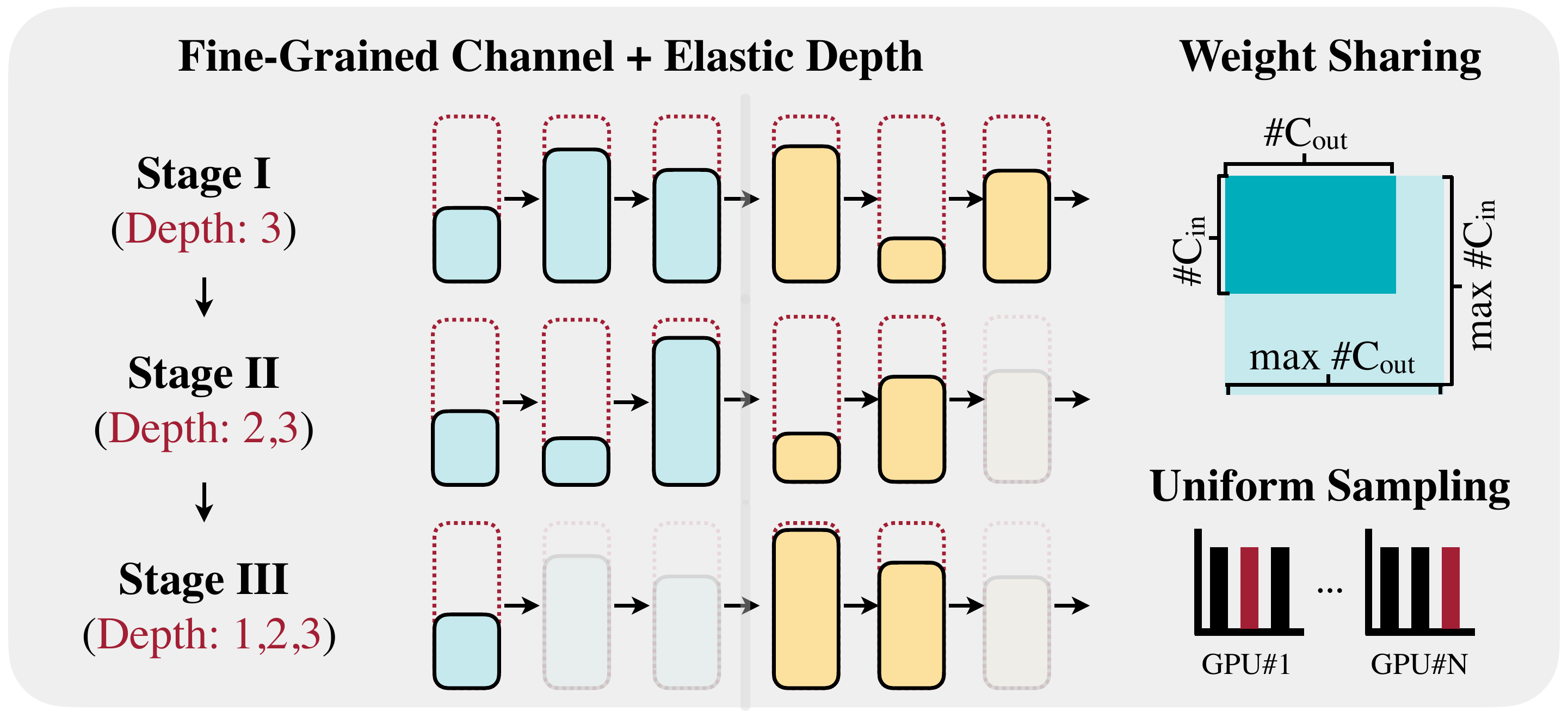}}
\hfill
\subfloat[Stage 2: Evolutionary Architecture Search]{\includegraphics[width=0.49\linewidth]{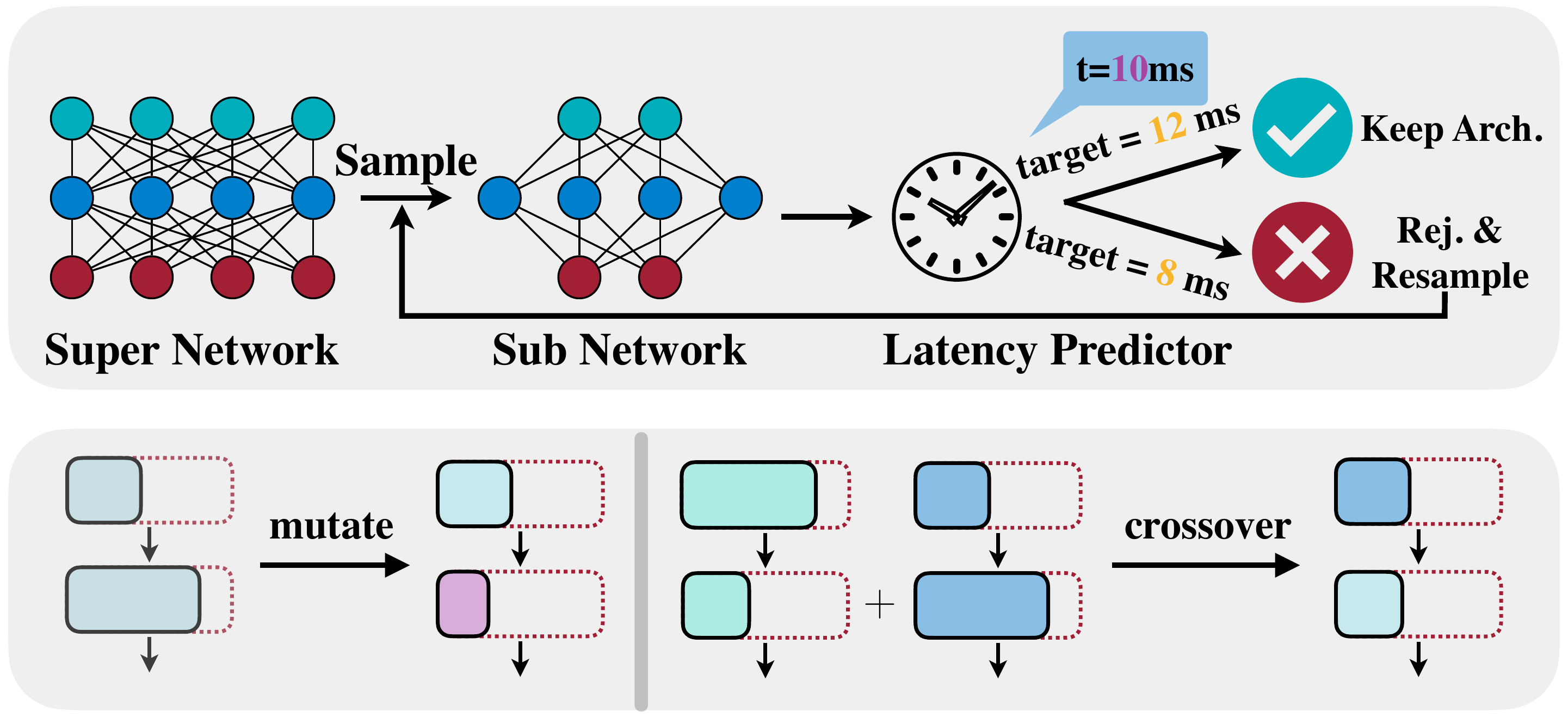}}
\caption{We propose a two-stage 3D Neural Architecture Search (3D-NAS) framework to automatically design efficient 3D deep learning architectures. \textbf{(a)} In the first stage, we train a \emph{super network} that supports all candidate networks within the design space. \textbf{(b)} In the second stage, we perform \emph{evolutionary architecture search} to find the best candidate network given a specific resource constraint.}
\label{fig:nas:overview}
\end{figure*}

To this end, we introduce \emph{Sparse Point-Voxel Convolution (SPVConv)} to effectively and efficiently process the large 3D scene. It follows a similar two-branch architectural design as PVConv except that the volumetric convolution in the voxel-based branch is replaced with the sparse convolution~\cite{choy20194d}. As point clouds are intrinsically very sparse, this modification can significantly reduce the memory consumption if we use a higher resolution in the voxel-based branch. Furthermore, SPVConv can also be considered as adding a high-resolution point-based branch to the vanilla sparse convolution so that the fine details (\ie, small instances) can be preserved.

Most of the operations in PVConv can be directly adapted to the sparse voxelized representation. However, the voxelization and devoxelization is not as trivial since we cannot easily index a given 3D coordinate in the sparse voxelized representation. A straightforward implementation based on the iterative coordinate comparison will require $\mathcal{O}(mn)$ of time, where $m$ is the number of points in the point cloud representation, and $n$ is the number of activated points in the sparse voxelized representation. As $m$ and $n$ are typically at the order of $10^5$, this naive implementation is not practical for real-time applications. Instead, we propose to use the parallel hash tables on GPUs to accelerate the sparse voxelization and devoxelization. Note that existing implementations~\cite{graham20183d,choy20194d} use CPU-based hash tables for kernel map construction in the sparse convolution, which is much slower. Concretely, we first construct a hash table for all activated points in the sparse voxelized representation, which can be finished in $\mathcal{O}(n)$ of time. After that, we iterate over all points, and for each point, we then query its coordinate in the hash table to obtain its corresponding index in the sparse voxelized representation. As the lookup over the hash table requires $\mathcal{O}(1)$ time in the worst case, this query step will in total take $\mathcal{O}(m)$ time. Therefore, the total time of coordinate indexing will be reduced from $\mathcal{O}(mn)$ to $\mathcal{O}(m+n)$.

\section{Searching Efficient 3D Architectures}

Even with the efficient 3D primitive, designing an efficient neural network model is still challenging. We need to carefully adjust the network architecture (\eg, channel numbers and kernel sizes of all layers) to meet the requirements for real-world applications (\eg, latency, energy, and accuracy). In this section, we propose \emph{3D Neural Architecture Search (3D-NAS)} to automatically design efficient 3D models (\fig{fig:nas:overview}). We first carefully design a search space tailored for 3D and then introduce our training paradigm that supports a large number of neural networks within a single super network. Finally, we use the evolutionary search to explore the best candidate in the design space given a resource constraint.

\subsection{Design Space}

The performance of neural architecture search is impacted by the design space quality. In our search space, we incorporate fine-grained channel numbers and elastic network depths; however, we do not support different kernel sizes.

\subsubsection{{Primitive Selection}}
\label{sect:method:nas:primitive}

\begin{figure}[t]
\centering
\includegraphics[width=0.95\linewidth]{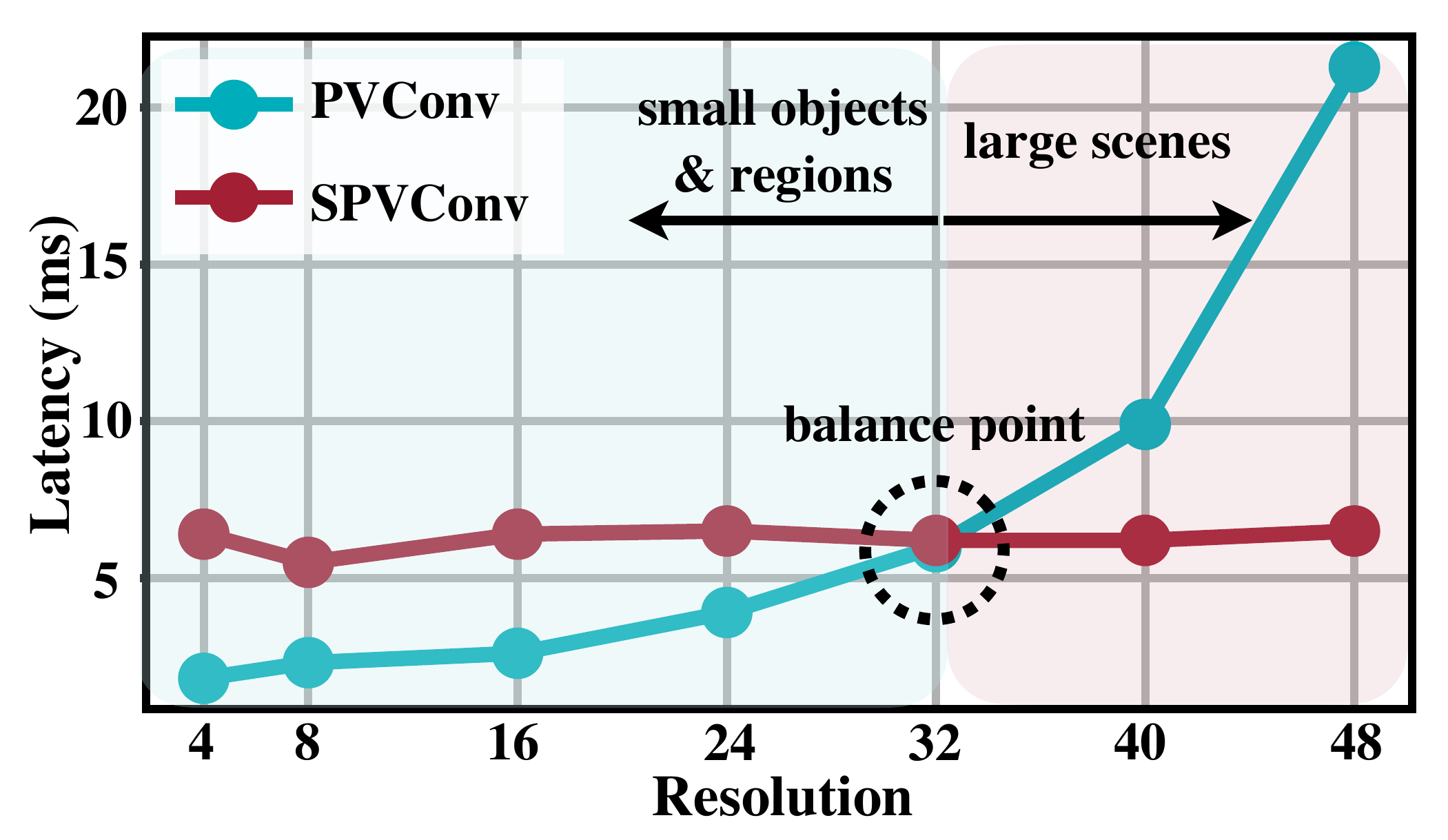}
\caption{{PVConv is more efficient and effective at smaller resolutions while SPVConv is more efficient at larger resolutions. Here, the GPU latency is measured on NVIDIA GTX 1080 Ti.}}
\label{fig:primitive:resolution}
\end{figure}

{PVConv accesses the memory contiguously and has fairly small memory footprint for small objects and indoor scans. However, it does not scale up very well to large-scale outdoor scenes (\tab{tab:primitive:limitation}) as the resolution of its voxel-based branch is still constrained by the memory. SPVConv, on the other hand, can efficiently scale up to large scans but falls short in modeling small objects and regions due to the irregular overhead introduced by sparse operations. As in \fig{fig:primitive:resolution}, the latency of SPVConv almost stays constant when the voxel resolution increases from 4 to 48, but the latency of PVConv quickly grows when the resolution scales up. As a result, at smaller resolutions, PVConv can be up to 3.6$\times$ faster than SPVConv, while SPVConv becomes 3.3$\times$ faster than PVConv once the resolution is larger than 48. As the spatial range of single objects/indoor scans is smaller than 1.5m$\times$1.5m$\times$1.5m, it is sufficient to use voxel resolution smaller than 32, and therefore, PVConv is favored over SPVConv. On the other hand, SPVConv is favored when the spatial range scales far beyond 1.5m$\times$1.5m$\times$1.5m. We will validate this in \sect{sect:exp}.}

\subsubsection{Fine-Grained Channel Numbers}

The computation cost increases quadratically with the number of channels; therefore, the channel number selection has a large influence on the network efficiency. Most existing neural architecture frameworks~\cite{cai2019proxylessnas} only support the coarse-grained channel number selection: \eg, searching the expansion ratio of the ResNet/MobileNet blocks over a few (2-3) choices. In this case, only intermediate channel numbers of the blocks can be changed; while the input and output channel numbers will still remain the same. Empirically, we observe that this limits the variety of the search space. To this end, we enlarge the search space by allowing all channel numbers to be selected from a large collection of choices (with size of $O(n)$). This fine-grained channel number selection largely increases the number of candidates for each block: \eg, from 2-3 to $\mathcal{O}(n^2)$ for a block with two convolutions.

\subsubsection{Elastic Network Depth}

For 3D CNNs, reducing the channel numbers alone cannot achieve significant measured speedup, which is different from normal 2D CNNs. For example, by shrinking all channel numbers in MinkowskiNet by 4$\times$ and 8$\times$, the number of MACs is reduced to 7.5 G and 1.9 G, respectively. However, although the number of MACs is drastically reduced, their measured latency on the GPU is very similar: 105 ms and 96 ms (on NVIDIA GTX 1080 Ti GPU). This suggests that scaling down the number of channels alone is not able to offer us with very efficient models, even though the number of MACs is very small. This might be because 3D modules are usually more memory-bounded than 2D modules; the number of MACs decreases quadratically with channel number, while memory decreases linearly. Thus, we incorporate the elastic network depth into our design space so that layers with very small computation (and large memory cost) can be removed and merged into their neighboring layers.

\subsubsection{Small Kernel Matters}

Kernel sizes are usually included into the search space of 2D CNNs. This is because a single convolution with larger kernel size can be more efficient than multiple convolutions with smaller kernel sizes on GPUs. However, it is not the case for 3D CNNs. From the perspective of computation cost, a single 2D convolution with kernel size of 5 requires only 1.4$\times$ more MACs than two 2D convolutions with kernel sizes of 3; while a single 3D convolution with kernel size of 5 requires 2.3$\times$ more MACs than two 3D convolutions with kernel sizes of 3 (if applied to dense voxel grids). This larger computation cost makes it less suitable to use large kernel sizes in 3D CNNs. Furthermore, the computation overhead of 3D modules is also related to the kernel sizes. For example, the sparse convolution requires $\mathcal{O}(k^3n)$ time to build the kernel map, where $k$ is the kernel size and $n$ is the number of points, which indicates that its cost grows cubically with respect to the kernel size. Based on these reasons, we decide to keep the kernel size of all convolutions to be 3 and do not allow the kernel size to change in our search space. Even with the small kernel size, we can still achieve a large receptive field by changing the network depth, which can achieve the same effect as changing the kernel size.

\subsection{Training Paradigm}

Searching over a fine-grained design space is very challenging since it is impossible to train every sampled candidate network randomly from scratch~\cite{tan2019mnasnet}. Motivated by Guo~\etal~\cite{guo2019single}, we incorporate all candidate networks into a single super network, and after training this super network once, each candidate network can then be extracted with inherited weights. The total training cost {during the neural architecture search} can then be reduced from $\mathcal{O}(n)$ to $\mathcal{O}(1)$, where $n$ is the number of candidate networks.

\subsubsection{Uniform Sampling}

At each training iteration, we randomly sample a candidate network from the super network: \ie, randomly select the channel number for each layer, and then randomly select the network depth (\ie the number of blocks to be used) for each stage. The total number of candidate networks to be sampled during training is very limited; thus, we choose to sample different candidate networks on different GPUs and average their gradients at each step so that more candidate networks can be sampled. For 3D, this is more critical because the 3D datasets usually contain fewer samples than the 2D datasets: \eg, 20K on SemanticKITTI~\cite{behley2019semantickitti} \vs 1M on ImageNet~\cite{deng2009imagenet}.

\subsubsection{Weight Sharing}

As the number of candidate networks is enormous, every candidate network will only be optimized for a small fraction of the total schedule. Therefore, uniform sampling alone is not enough to train all candidate networks sufficiently (\ie, achieving the same level of performance as being trained from scratch). To tackle this, we adopt the weight sharing technique so that every candidate network can be optimized at each iteration even if it is not sampled. Specifically, given the input channel number $C_{\text{in}}$ and output channel number $C_{\text{out}}$ of each convolution layer, we simply index the first $C_{\text{in}}$ and $C_{\text{out}}$ channels from the weight tensor accordingly to perform the convolution~\cite{guo2019single}. For each batch normalization layer, we similarly crop the first $c$ channels from the weight tensor based on the sampled channel number $c$. Finally, with the sampled depth $d$ for each stage, we choose to keep the first $d$ layers, instead of randomly sampling $d$ of them. This ensures that each layer will always correspond to the same depth index within the stage.

\subsubsection{Progressive Depth Shrinking}

Suppose we have $n$ stages, each of which has $m$ different depth choices ranging from 1 to $m$. If we sample the depth $d_k$ for each stage $k$ randomly, the expected total depth of the network will be
$\mathbb{E}[d] = \sum_{k=1}^n \mathbb{E}[d_k] = n(m+1)/2$,
which is much smaller than the maximum depth $nm$. Furthermore, the probability of the largest candidate network (with the maximum depth) being sampled is extremely small: $m^{-n}$. Therefore, the largest candidate networks are poorly trained due to the small possibility of being sampled. To this end, we introduce progressively depth shrinking to alleviate this issue. We divide the training epochs into $m$ segments for $m$ different depth choices. During the $k^{\text{th}}$ training segment, we only allow the depth of each stage to be selected from $m-k+1$ to $m$. This is essentially designed to enlarge the search space gradually so that these large candidate networks can be sampled more frequently.

\subsection{Search Algorithm}

After the super network is fully trained, we use the evolutionary architecture search to find the best network architecture under a certain resource constraint.

\subsubsection{Resource Constraint}

We support \#MACs and measured latency as our resource constraint for candidate networks.

\myparagraph{\#MACs Constraint.}
\#MACs is an implementation- and hardware-agnostic efficiency metric. Unlike dense 2D CNNs, \#MACs of sparse 3D CNNs cannot be simply determined by the input size and network architecture. This is because the sparse convolution only performs the computation over the active synapses; thus, its computation cost is also related to the size of kernel map, which is determined by the input sparsity pattern. To address this, we first estimate the average kernel map size over the entire dataset for each convolution layer and then sum them up to compute the average \#MACs.

\myparagraph{Latency Constraint.}
We support to directly use the measured latency on the target hardware as our resource constraint as well. We first encode each candidate network into a 1D architecture vector. We then randomly sample 50,000 candidate networks from the design space and measure their latency on the target hardware. With the collected pairs of architecture vector and measured latency, we finally train an MLP regressor to estimate the latency based on the network architecture. The resulting predictor can accurately predict the latency of different candidate networks with a relative error of less than 2\% on both edge and cloud GPUs.

\subsubsection{Evolutionary Search}

We automate the architecture search with the evolutionary algorithm~\cite{guo2019single}. We initialize the starting population with $n$ randomly sampled candidate networks. At every iteration, we evaluate all candidate networks in the population and select the $k$ models with the highest accuracy (\ie, the fittest individuals). The population for the next iteration is then generated with $(n/2)$ mutations and $(n/2)$ crossovers. For each mutation, we randomly select one among the top-$k$ candidates and modify each of its architectural parameters (\eg, channel numbers, network depths) with a pre-defined probability; for each crossover, we select two from the top-$k$ candidates and produce a new network by fusing them together randomly. Finally, the best network architecture is obtained from the population of the last iteration. During the evolutionary search, we ensure that all candidate networks in the population always meet the given resource constraint (we will resample another candidate network until the resource constraint is satisfied).

\begin{figure*}[!t]
\centering
\captionsetup[subfigure]{position=top}
\subfloat[Top Row: Features Extracted from \emph{Coarse-Grained} Voxel-Based Branch (Large, Continuous)]{%
\includegraphics[width=0.115\linewidth]{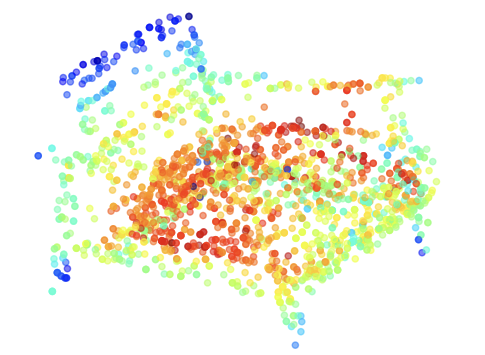}%
\includegraphics[width=0.115\linewidth]{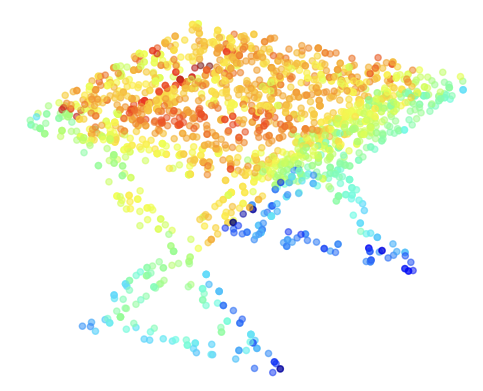}%
\includegraphics[width=0.115\linewidth]{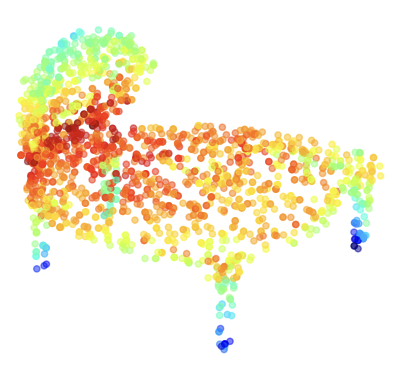}%
\includegraphics[width=0.115\linewidth]{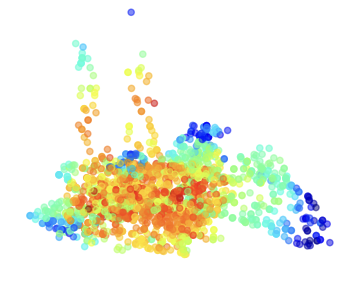}%
\includegraphics[width=0.115\linewidth]{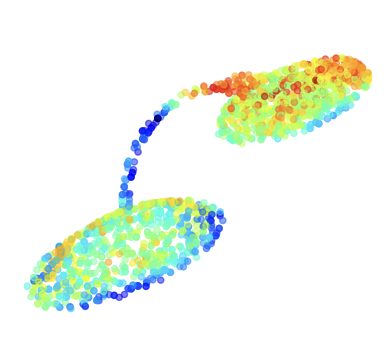}%
\includegraphics[width=0.115\linewidth]{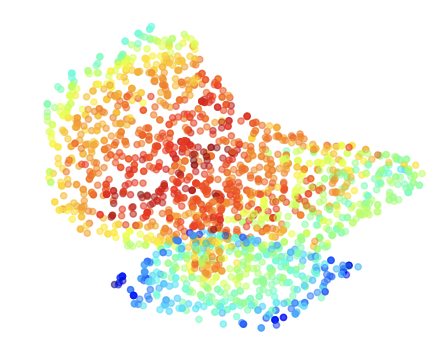}%
\includegraphics[width=0.115\linewidth]{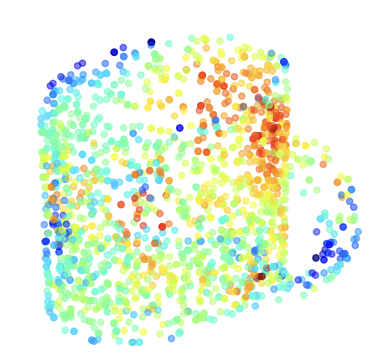}%
\includegraphics[width=0.115\linewidth]{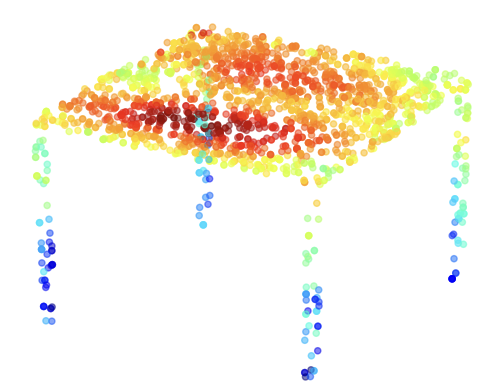}%
}
\\
\captionsetup[subfigure]{position=bottom}
\subfloat[Bottom Row: Features Extracted from \emph{Fine-Grained} Point-Based Branch (Isolated, Discontinuous)]{%
\includegraphics[width=0.115\linewidth]{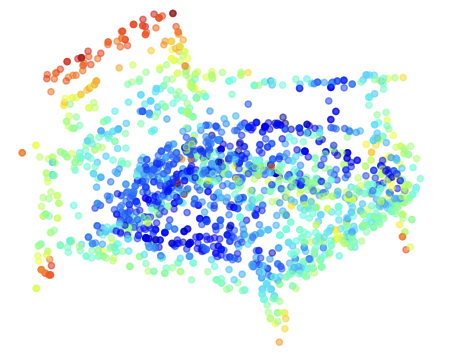}%
\includegraphics[width=0.115\linewidth]{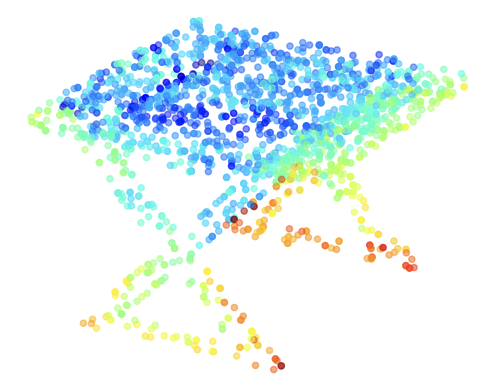}%
\includegraphics[width=0.115\linewidth]{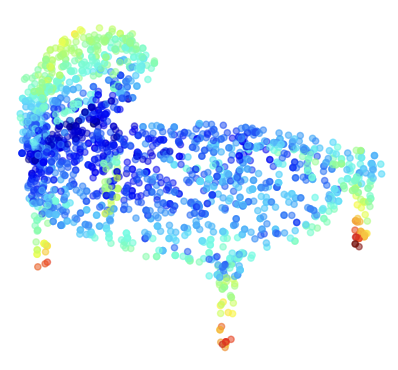}%
\includegraphics[width=0.115\linewidth]{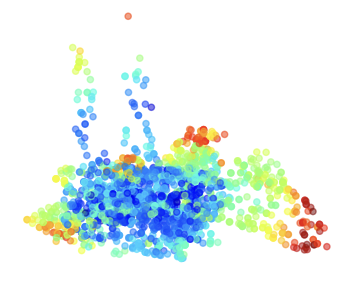}%
\includegraphics[width=0.115\linewidth]{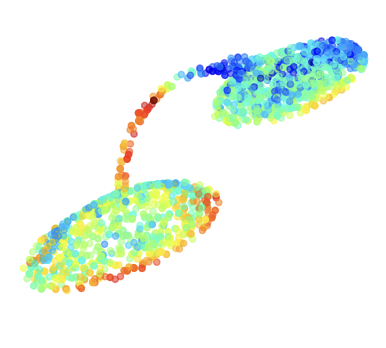}%
\includegraphics[width=0.115\linewidth]{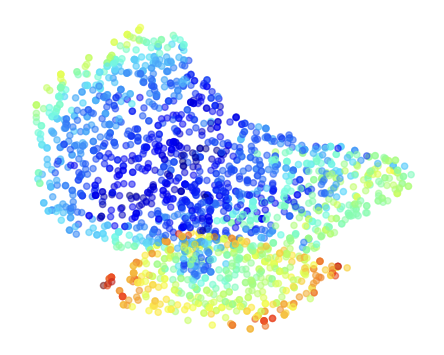}%
\includegraphics[width=0.115\linewidth]{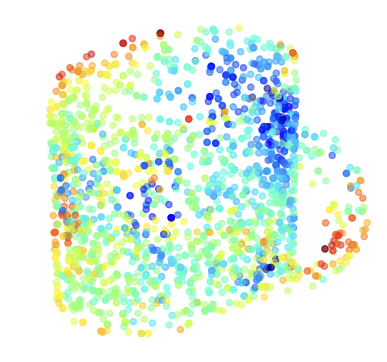}%
\includegraphics[width=0.115\linewidth]{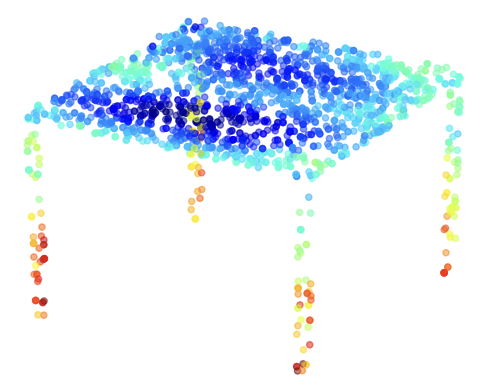}%
}
\caption{Two branches are providing complementary information: the voxel-based branch focuses on the large, continuous parts, while the point-based focuses on the isolated, discontinuous parts.}
\label{fig:results:shapenet:features}
\end{figure*}

\begin{table}[!t]
\renewcommand{\arraystretch}{1.3}
\setlength{\tabcolsep}{4.5pt}
\small\centering

\caption{Results of Object Part Segmentation on ShapeNet Part}
\label{tab:exp:shapenet:results}

\vspace{-6pt}
\scalebox{0.825}{
\begin{tabular}{lcccccc}
    \toprule
     & \#Par. (M) & \#MACs (G) & Mem. (G) & Lat. (ms) & mIoU \\
    \midrule
    PointNet~\cite{qi2017pointnet} &  2.5 & 5.3 & 1.5 & 15.1 & 83.7\\
    3D-UNet~\cite{cicek20163d} & 8.1 & 2996.9 & 8.8 & 682.1 & 84.6\\
    RSNet~\cite{huang2018recurrent} & 6.9 & 1.4 & 0.8 & 74.6 & 84.9\\
    PointNet++~\cite{qi2017pointnet++} & 1.8 & 4.9 & 2.0 & 77.9 & 85.1\\
    DGCNN~\cite{wang2018dynamic} & 1.5 & 18.5 & 2.4 & 87.8 & 85.1\\
    {\textbf{SPVCNN} (0.25$\times$C)} & {0.3} & {0.3} & {1.1} & {20.2} & {84.4}\\
    \textbf{PVCNN} (0.25$\times$C) & \textbf{0.3} & 1.0 & \textbf{0.8} & 8.3 & \textbf{85.2}\\
    {\textbf{SPVNAS}} & {1.7} & {1.0} & {1.2} & {18.4} & {85.2}\\
    \textbf{PVNAS-A} & 0.4 & \textbf{0.9} & 0.9 & \textbf{7.0} & \textbf{85.2}\\
    \midrule
    SpiderCNN~\cite{xu2018spidercnn} & 2.6 & 10.6 & 6.5 & 170.7 & 85.3 \\
    {\textbf{SPVCNN} (0.5$\times$C)} & {1.1} & {1.3} & {1.4} & {24.7} & {85.1}\\
    \textbf{PVCNN} (0.5$\times$C) & 1.1 & 3.9 & \textbf{1.0} & 16.0 & 85.5\\
    \textbf{PVNAS-B} & \textbf{0.6} & \textbf{2.2} & \textbf{1.0} & \textbf{11.6} & 85.5\\
    \textbf{PVNAS-C} & 0.7 & 2.9 & \textbf{1.0} & 14.0 & \textbf{85.6}\\
    \midrule
    PointConv~\cite{wu2019pointconv} & 21.6 & \textbf{11.6} & 6.5 & 163.7 & 85.7\\
    PointCNN~\cite{li2018pointcnn} &  8.3 & 26.9 & 2.5 & 135.8 & 86.1\\
    {\textbf{SPVCNN} (1$\times$C)} & {4.2} & {5.3} & {2.0} & {38.0} & {85.6}\\
    \textbf{PVCNN} (1$\times$C) & \textbf{4.2} & 15.3 & \textbf{1.6} & \textbf{41.4} & \textbf{86.2}\\
    \bottomrule
\end{tabular}
}

\vspace{8pt}\justifying\noindent
{\footnotesize\itshape{On average, PVCNN outperforms point-based models with \textbf{5.5$\times$} speedup and \textbf{3$\times$} memory reduction, and outperforms the voxel-based baseline with \textbf{59$\times$} measured speedup and \textbf{11$\times$} memory reduction.}}
\end{table}

\section{Experiments}
\label{sect:exp}

In this section, we evaluate our models on six representative 3D benchmark datasets:
\begin{itemize}
    \item ShapeNet~\cite{chang2015shapenet} (coarse-grained object part segmentation),
    \item PartNet~\cite{mo2019partnet} (fine-grained object part segmentation),
    \item S3DIS~\cite{armeni20163d,armeni2017joint} (indoor scene segmentation),
    \item SemanticKITTI~\cite{behley2019semantickitti} (outdoor scene segmentation),
    \item nuScenes~\cite{caesar2020nuscenes} (outdoor scene segmentation),
    \item KITTI~\cite{geiger2012kitti} (outdoor object detection).
\end{itemize}
On these datasets, we evaluate four variants of our method:
\begin{itemize}
    \item \textbf{PVCNN} (manually-designed model with PVConv),
    \item \textbf{PVNAS} (3D-NAS applied to PVCNN),
    \item \textbf{SPVCNN} (manually-designed model with SPVConv),
    \item \textbf{SPVNAS} (3D-NAS applied to SPVCNN).
\end{itemize}

\subsection{3D Object Part Segmentation}

\subsubsection{ShapeNet}

We first evaluate our method on 3D part segmentation and conduct experiments on the large-scale 3D object dataset, ShapeNet~\cite{chang2015shapenet}. As 3D objects are very small, it is suitable to process them using PVConv. Therefore, we build our PVCNN by replacing the MLP layers in PointNet~\cite{qi2017pointnet} with PVConv's.

\myparagraph{Baselines.}
We use PointNet~\cite{qi2017pointnet}, RSNet~\cite{huang2018recurrent}, PointNet++~\cite{qi2017pointnet++} (with multi-scale grouping), DGCNN~\cite{wang2018dynamic}, SpiderCNN~\cite{xu2018spidercnn} and PointCNN~\cite{li2018pointcnn} as point-based baselines. We reimplement 3D-UNet~\cite{cicek20163d} as voxel-based baseline. For a fair comparison, we follow the same evaluation protocol as in Li~\etal~\cite{li2018pointcnn} and Graham~\etal~\cite{graham20183d}. The evaluation metric is mean intersection-over-union (mIoU): we first calculate the part-averaged IoU for each of the 2,874 test models and average the values as the final metrics. Besides, we report the measured GPU latency and GPU memory consumption on a single NVIDIA GTX 1080 Ti GPU. We ensure the input data to have the same size with 2048 points and batch size of 8.

\begin{table*}[!t]
\renewcommand{\arraystretch}{1.3}
\setlength{\tabcolsep}{2.5pt}
\small\centering

\caption{Results of Fine-Grained Object Part Segmentation on PartNet}
\label{tab:exp:partnet:results}

\vspace{-6pt}
\scalebox{0.825}{
\begin{tabular}{lcccc|ccccccccccccccccc}
\toprule
& {\#P (M)} & {\#M (G)} & {L (ms)} & {mIoU} & {Bed} & {Bott} & {Chair} & {Clock} & {Dish} & {Disp} & {Door} & {Ear} & {Fauc} & {Knife} & {Lamp} & {Micro} & {Frid} & {Stora} & {Table} & {Trash} & {Vase} \\
\midrule
PointNet~\cite{qi2017pointnet} & 2.5 & 25.1 & 9.4 & 35.6 & 13.4 & 29.5 & 27.8 & 28.4 & 48.9 & 76.5 & 30.4 & 33.4 & 47.6 & 32.9 & 18.9 & 37.2 & 33.5 & 38.0 & 29.0 & 34.8 & 44.4 \\
PointNet++~\cite{qi2017pointnet++} & 1.8 & 5.4 & 26.0 & 42.5 & 30.3 & 41.4 & 39.2 & 41.6 & 50.1 & 80.7 & 32.6 & 38.4 & 52.4 & 34.1 & 25.3 & 48.5 & 36.4 & 40.5 & 33.9 & 46.7 & 49.8 \\
Deep LPN~\cite{le2020going} & 2.7 & 3.0 & 206.8 & 38.6 & 29.5 & 42.1 & 41.8 & 34.7 & 33.2 & 81.6 & 34.8 & 49.6 & 53.0 & 44.8 & 28.4 & 33.5 & 32.3 & 41.1 & 36.3 & 43.1 & 57.8 \\
SpiderCNN~\cite{xu2018spidercnn} & 2.6 & 52.2 & 2170.8 & 37.0 & 36.2 & 32.2 & 30.0 & 24.8 & 50.0 & 80.1 & 30.5 & 37.2 & 44.1 & 22.2 & 19.6 & 43.9 & 39.1 & 44.6 & 20.1 & 42.4 & 32.4 \\
PointCNN~\cite{li2018pointcnn} & 8.3 & 71.9 & 106.6 & 46.4 & 41.9 & 41.8 & 43.9 & 36.3 & 58.7 & 82.5 & 37.8 & 48.9 & 60.5 & 34.1 & 20.1 & 58.2 & 42.9 & 49.4 & 21.3 & 53.1 & 58.9 \\
ResGCN~\cite{li2019deepgcns} & 3.8 & 55.9 & 771.3 & 45.1 & 35.9 & 49.3 & 41.1 & 33.8 & 56.2 & 81.0 & 31.1 & 45.8 & 52.8 & 44.5 & 23.1 & 51.8 & 34.9 & 47.2 & 33.6 & 50.8 & 54.2 \\
{\textbf{SPVCNN} (0.5$\times$C)} & {0.3} & {1.6} & {13.5} & {43.6} & {32.8} & {45.1} & {35.3} & {33.7} & {58.1} & {79.5} & {36.2} & {49.4} & {48.7} & {39.5} & {22.7} & {49.3} & {38.1} & {41.2} & {27.2} & {49.1} & {55.4}\\
\textbf{PVCNN} (0.5$\times$C) & 0.3 & 2.3 & 4.4 & 47.2 & 35.4 & 44.5 & 37.0 & 38.9 & 63.2 & 81.4 & 44.2 & 52.1 & 53.5 & 46.0 & 23.1 & 54.5 & 43.6 & 44.1 & 30.3 & 52.7 & 57.3 \\
{\textbf{SPVCNN} (0.72$\times$C)} & {0.6} & {3.3} & {15.4} & {44.3} & {33.3} & {47.2} & {36.9} & {36.0} & {59.6} & {79.5} & {34.9} & {49.0} & {50.4} & {37.4} & {23.9} & {48.9} & {40.4} & {42.6} & {27.8} & {50.1} & {54.5}\\
\textbf{PVCNN} (0.72$\times$C) & 0.6 & 4.7 & 5.5 & 47.3 & 35.6 & 43.5 & 38.8 & 39.5 & 62.9 & 81.0 & 42.6 & 53.0 & 54.5 & 42.9 & 24.4 & 53.5 & 42.5 & 45.6 & 32.0 & 52.1 & 59.7 \\
{\textbf{SPVCNN} (1$\times$C)} & {1.1} & {6.4} & {16.9} & {45.0} & {32.9} & {46.1} & {37.6} & {33.5} & {58.7} & {80.0} & {38.3} & {51.6} & {50.6} & {44.6} & {23.4} & {49.3} & {38.7} & {44.1} & {29.8} & {51.0} & {54.7}\\
\textbf{PVCNN} (1$\times$C) & 1.1 & 9.1 & 6.9 & 47.8 & 35.9 & 43.9 & 39.8 & 40.0 & 61.5 & 81.6 & 44.9 & 52.5 & 54.6 & 45.0 & 26.0 & 55.2 & 44.3 & 46.2 & 32.7 & 51.6 & 56.7 \\
\midrule
SGAS~\cite{li2019sgas} & -- & -- & 143$^*$ & 48.3 & 43.4 & 50.8 & 41.2 & 38.8 & 61.4 & 82.6 & 37.1 & 48.8 & 56.1 & 49.4 & 21.2 & 56.5 & 44.5 & 49.4 & 29.3 & 54.4 & 56.0 \\
LC-NAS-14~\cite{li2020lcnas} & -- & -- & 152$^*$ & 48.6 & 41.9 & 51.7 & 39.7 & 39.6 & 61.5 & 82.5 & 39.3 & 49.0 & 54.7 & 55.3 & 22.2 & 55.1 & 45.2 & 48.0 & 30.3 & 54.6 & 54.9 \\
LC-NAS-18~\cite{li2020lcnas} & -- & -- & 185$^*$ & 46.6 & 40.7 & 50.5 & 39.9 & 39.5 & 59.8 & 82.2 & 35.0 & 44.5 & 53.2 & 44.9 & 22.0 & 54.1 & 41.5 & 45.8 & 31.5 & 53.0 & 54.4 \\
{\textbf{SPVNAS}} & {0.4} & {1.2} & {13.2} & {46.8} & {36.7} & {48.2} & {39.4} & {35.7} & {61.9} & {81.8} & {40.0} & {54.2} & {55.1} & {40.4} & {25.3} & {50.8} & {41.3} & {45.9} & {31.3} & {51.5} & {56.7}\\
\textbf{PVNAS-A} & 0.3 & 2.7 & 4.2 & 47.5 & 35.2 & 44.1 & 39.5 & 36.6 & 59.6 & 81.8 & 44.8 & 52.4 & 54.7 & 47.3 & 23.2 & 54.1 & 45.0 & 45.8 & 32.5 & 53.4 & 57.6 \\
\textbf{PVNAS-B} & 0.4 & 3.9 & 4.9 & 48.0 & 37.3 & 44.5 & 40.2 & 36.0 & 60.6 & 82.1 & 43.8 & 52.3 & 55.8 & 47.3 & 23.8 & 52.2 & 47.8 & 47.8 & 34.1 & 52.7 & 57.5 \\
\bottomrule
\end{tabular}
}

\vspace{8pt}\justifying\noindent
{\footnotesize\itshape{Here, \textbf{\#P} denotes the number of parameters, \textbf{\#M} denotes the number of MACs, and \textbf{L} denotes the measured latency (on a single NVIDIA GTX 1080 Ti GPU). $^*$: numbers are from Li~\etal~\cite{li2020lcnas}, which are measured on a single NVIDIA RTX 2080 GPU.}}
\end{table*}
\begin{figure}[!t]
\centering
\captionsetup[subfigure]{position=top}
\subfloat[PVCNN \vs PointNet]{%
\includegraphics[width=0.49\linewidth]{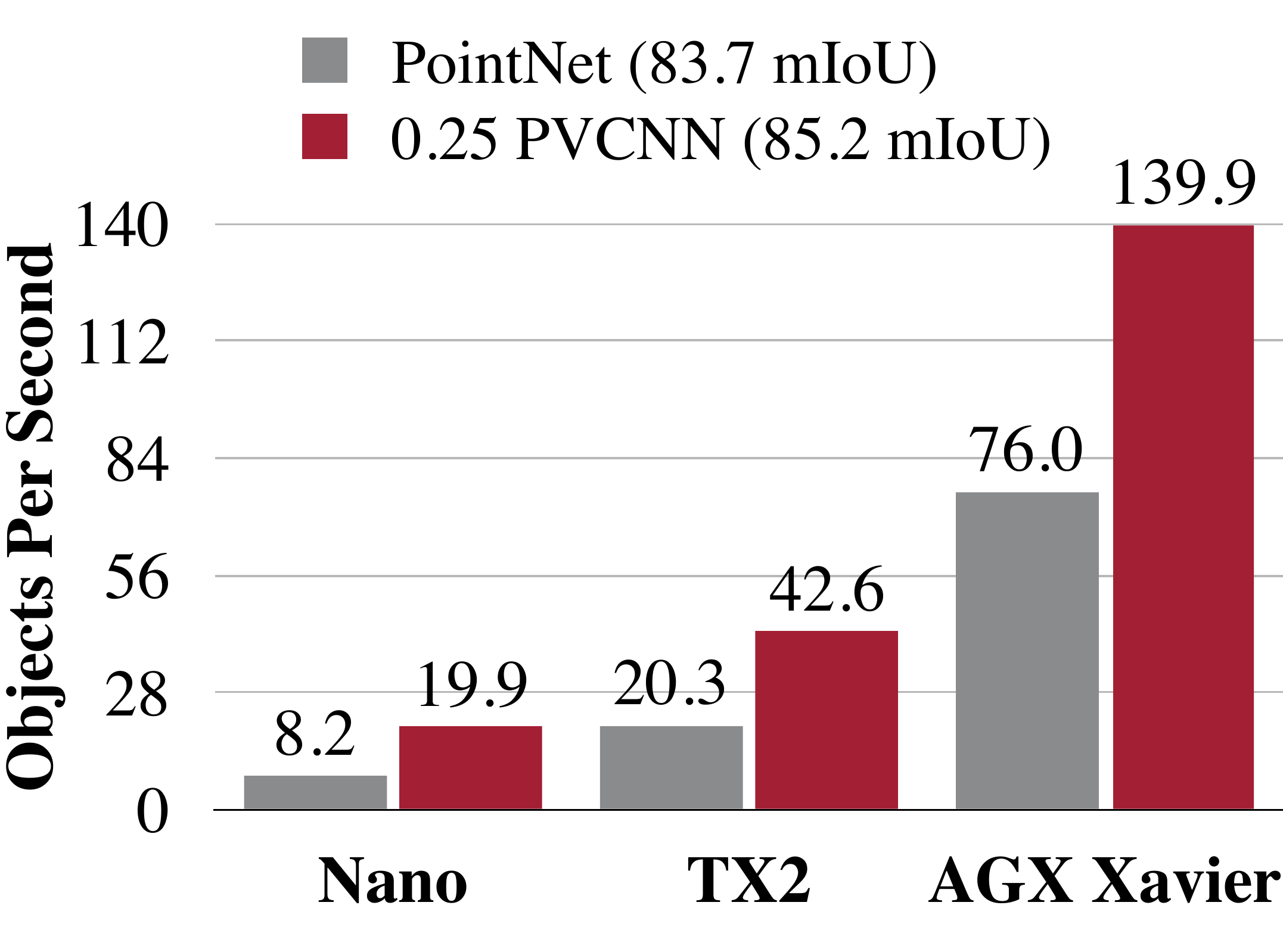}%
\label{fig:exp:edge:a}}
\hfil
\subfloat[PVNAS \vs PVCNN]{%
\includegraphics[width=0.49\linewidth]{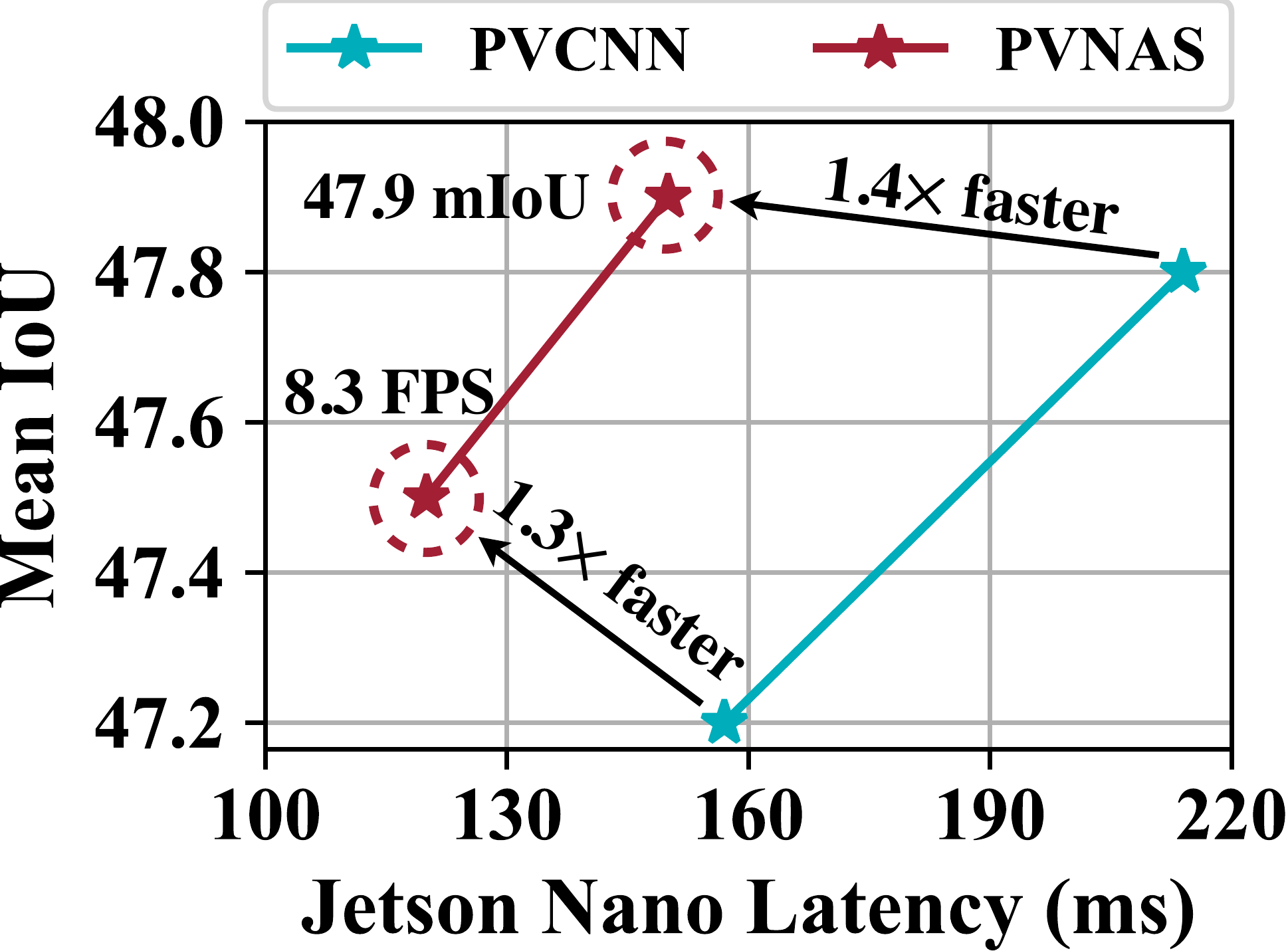}%
\label{fig:exp:edge:b}
}
\caption{PVCNN achieves \textbf{real-time} 3D object segmentation with 2,048 input points on edge devices. With PVNAS, we further boost the efficiency on NVIDIA Jetson Nano, achieving \textbf{8.3 FPS} with \textbf{10,000} input points.}
\label{fig:exp:edge}
\end{figure}

\myparagraph{Results.}
As in \tab{tab:exp:shapenet:results}, PVCNN outperforms all previous models. It directly improves the accuracy of its backbone (PointNet) by 2.5\% with even smaller overhead compared with PointNet++. Besides, we also design narrower versions of our PVCNN by reducing the number of channels to 25\% (0.25$\times$C) and 50\% (0.5$\times$C). The resulting model requires only 46.4\% latency of PointNet, and it still outperforms several point-based methods with sophisticated neighborhood aggregation including RSNet, PointNet++ and DGCNN, which are almost an order of magnitude slower. PVCNN achieves a much better accuracy \vs latency trade-off compared with all point-based methods. With similar accuracy, PVCNN is \textbf{24$\times$} faster than SpiderCNN and \textbf{3.3$\times$} faster than PointCNN. Our PVCNN also achieves a significantly better accuracy \vs memory trade-off compared with the voxel-based baseline. With better accuracy, PVCNN can save the GPU memory consumption by \textbf{10$\times$} compared with 3D-UNet. We further apply 3D-NAS to PVCNN under different latency constraints to obtain a family of PVNAS models. With the same accuracy as PVCNN (0.25$\times$C), PVNAS-A achieves 1.2$\times$ faster inference speed. It also outperforms PointNet by 1.5 mIoU with \textbf{3$\times$} measured speedup. In comparison with PVCNN (0.5$\times$C), PVNAS-B achieves 1.4$\times$ speedup with no loss of accuracy, and PVNAS-C achieves 1.1$\times$ speedup with better mIoU.

\myparagraph{PV/SPVConv.}
{On ShapeNet, PVCNN/PVNAS achieves far better efficiency-accuracy trade-offs compared with SPVCNN/SPVNAS. Specifically, PVCNN achieves similar or better mIoU than SPVCNN with \textbf{2.4}$\times$ measured speedup. Similarly, PVNAS-A achieves the same accuracy as SPVNAS with \textbf{2.6}$\times$ lower latency. These results validate our claim in \sect{sect:method:nas:primitive} that PVConv is more favorable for modeling small 3D objects.}

\myparagraph{Visualization.}
We visualize the voxel and point features from the final PVConv, where warmer color indicates larger magnitude. As in \fig{fig:results:shapenet:features}, the voxel branch tends to capture large, continuous parts (\eg, table top, lamp head) while the point branch captures isolated, discontinuous details (\eg, table legs, lamp neck). Two branches indeed provide complementary information. This is aligned with our design.

\subsubsection{PartNet}

We also conduct experiments on the more challenging fine-grained 3D object part segmentation benchmark, PartNet~\cite{mo2019partnet}. Different from ShapeNet, where each object is annotated with 2 to 6 parts, objects in PartNet can have as many as 50 parts. To perform fine-grained segmentation, the models usually take in a batch of 10,000 points instead of 2,048 points.

\myparagraph{Baselines.}
We compare PVCNN with state-of-the-art point-based methods including PointNet~\cite{qi2017pointnet}, PointNet++~\cite{qi2017pointnet++}, Deep LPN~\cite{le2020going}, SpiderCNN~\cite{xu2018spidercnn}, PointCNN~\cite{li2018pointcnn} and ResGCN~\cite{li2019deepgcns}. We also compare PVNAS with automatically-designed point cloud segmentation networks including SGAS~\cite{li2019sgas} and LC-NAS~\cite{li2020lcnas}. To ensure fair comparisons, we follow the original experiment setting in Mo~\etal~\cite{mo2019partnet} to train separate models for different object classes. The results of our method are averaged over at least three runs. For the other methods, we report the accuracy from their papers and measure \#Params, \#MACs and latency with their open-source implementation.

\myparagraph{Results.}
As in \tab{tab:exp:partnet:results}, PVCNN achieves superior results compared with all manually-designed point-based methods. Specifically, PVCNN (0.5$\times$C) outperforms PointCNN with \textbf{27.7$\times$} model size reduction, \textbf{31.3$\times$} computation reduction, and \textbf{23.7$\times$} measured speedup. This indicates that PVConv scales much better (\wrt the number of points) than other point-based primitives. Then, we apply 3D-NAS to PVCNN under 4.2/4.9 ms latency constraints on NVIDIA GTX 1080 Ti GPU. The resulting PVNAS outperforms PVCNN (0.72$\times$C) and PVCNN (1$\times$C) with 1.3$\times$ and 1.4$\times$ speedup, respectively. It also compares favorably with AutoML-based approaches. With more than \textbf{29$\times$} speedup, PVNAS achieves similar IoU compared with SGAS and LC-NAS. 

\myparagraph{PV/SPVConv.}
{On PartNet, PVCNN (0.5$\times$C) is \textbf{3.8}$\times$ faster than SPVCNN (1$\times$C), while achieving \textbf{2.2\%} higher accuracy. 3D-NAS improves the performance of SPVCNN significantly; however, SPVNAS is still \textbf{3.1}$\times$ slower than PVNAS-A. This again emphasizes the importance of primitive selection.}

\myparagraph{Deployment.}
We deploy our PVCNN and PVNAS on edge devices. As in \fig{fig:exp:edge:a}, PVCNN runs at real time (20 FPS) on NVIDIA Jetson Nano, which only consumes the power of a light bulb (5 W). Even when the number of input points increases by 5$\times$ on PartNet, PVNAS can still run at 8.3 FPS on NVIDIA Jetson Nano (\fig{fig:exp:edge:b}). Thus, PVNAS can empower efficient 3D vision on low-power devices, which becomes increasingly important with the rise of AR/VR applications.

\subsection{3D Indoor Scene Segmentation}

\begin{table}[t]
\renewcommand{\arraystretch}{1.3}
\setlength{\tabcolsep}{3pt}
\small\centering

\caption{Results of Indoor Scene Segmentation on S3DIS}
\label{tab:exp:s3dis:main}

\vspace{-6pt}
\scalebox{0.825}{
\begin{tabular}{lccccc}
    \toprule
    & \#Par. (M) & \#MACs (G) & Mem. (GB) & Lat. (ms) & mIoU \\
    \midrule
    PointNet~\cite{qi2017pointnet} & 1.2 & 3.6 & 1.0 & 12.1 & 43.0 \\
    \textbf{PVCNN} (0.125$\times$C) & 0.04 & 0.3 & \textbf{0.6} & \textbf{6.2} & \textbf{46.9} \\
    \midrule
    DGCNN~\cite{wang2018dynamic} & 1.0 & 36.9 & 2.4 & 168.1 & 48.0 \\
    RSNet~\cite{huang2018recurrent} & 6.9 & 2.2 & 1.1 & 111.5 & 52.0 \\
    \textbf{SPVCNN} (0.25$\times$C) & 0.2 & 0.4 & 1.1 & 21.1 & 50.4 \\
    \textbf{PVCNN} (0.25$\times$C) & 0.2 & 0.9 & \textbf{0.7} & \textbf{9.0} & \textbf{52.3} \\
    \midrule
    3D-UNet~\cite{cicek20163d} & 14.0 & 349.6 & 6.8 & 574.7 & 55.0 \\
    \textbf{SPVCNN} (1$\times$C) & 2.7 & 6.6 & 1.8 & 41.0 & 54.9 \\

    \textbf{PVCNN} (1$\times$C) & 2.6 & 13.0 & 1.3 & \textbf{39.1} & 56.1 \\
    \textbf{PVCNN++} (0.5$\times$C) & 3.4 & 6.6 & \textbf{0.7} & 40.9 & \textbf{57.6} \\
    \midrule
    PointCNN~\cite{li2018pointcnn} & 11.5 & 17.5 & 4.6 & 282.3 & 57.3 \\
    \textbf{PVCNN++} (1$\times$C) & 13.7 & 26.2 & \textbf{0.8} & \textbf{67.2} & \textbf{59.0} \\
    \bottomrule
\end{tabular}
}

\vspace{8pt}\justifying\noindent
{\footnotesize\itshape{On average, PVCNN and PVCNN++ outperform point-based models with \textbf{8$\times$} speedup and \textbf{3$\times$} memory reduction. They outperform the voxel-based baseline with \textbf{14$\times$} speedup and \textbf{10$\times$} memory reduction.}}
\end{table}

\begin{figure*}[!t]
\centering
\captionsetup[subfigure]{position=top}
\subfloat[Trade-Off: Mean IoU \vs GPU Latency]{\includegraphics[width=0.49\linewidth]{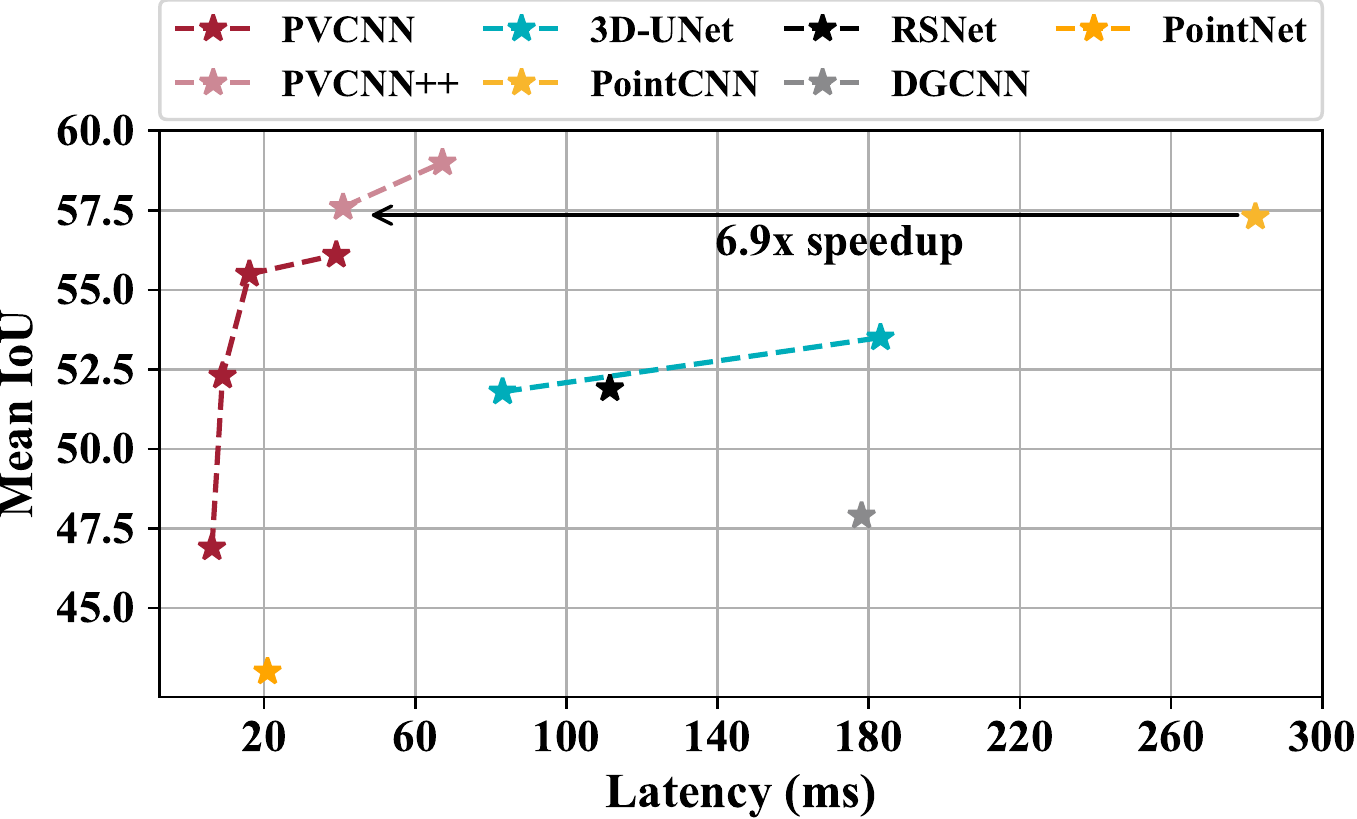}}
\hfil
\subfloat[Trade-Off: Mean IoU \vs GPU Memory]{\includegraphics[width=0.49\linewidth]{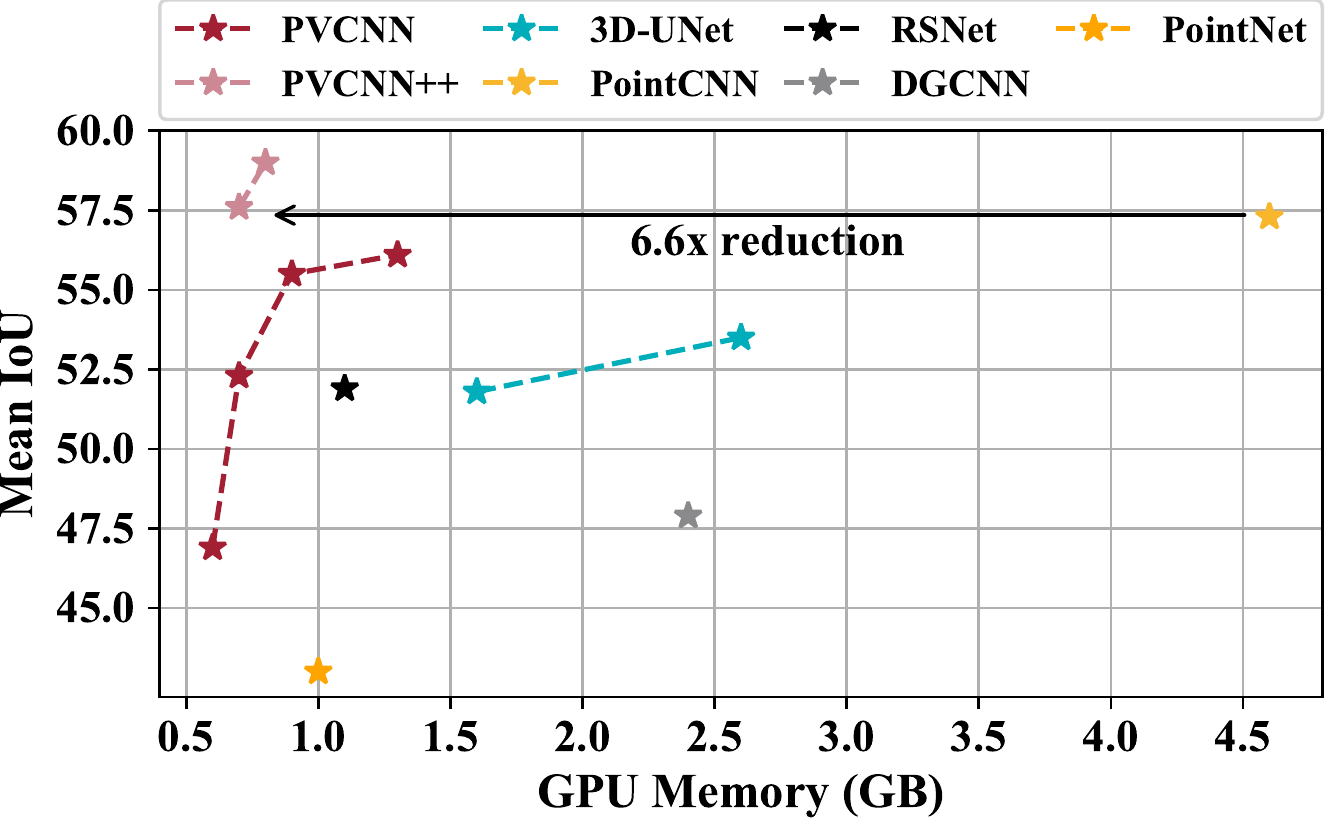}}
\caption{PVCNN achieves a much better trade-off between accuracy and efficiency than the point-based and voxel-based baselines on S3DIS.}
\label{fig:exp:s3dis:tradeoffs}
\end{figure*}

We then evaluate our method on 3D semantic segmentation and conduct experiments on the large-scale indoor scene dataset, S3DIS~\cite{armeni20163d,armeni2017joint}. Though indoor scenes are usually much larger (\eg, 6m$\times$6m$\times$3m) than single 3D objects, it is still affordable to process them using sliding windows (\eg, 1.5m$\times$1.5m$\times$3m). Note that the evaluation protocol is different from recent methods~\cite{choy20194d} that take in the entire scene. Our setting is closer to the real-world scenario since depth cameras only capture a small region in a single pass. Similar to object part segmentation, we build PVCNN based on PointNet~\cite{qi2017pointnet} and PVCNN++ based on PointNet++~\cite{qi2017pointnet++}.

\myparagraph{Baselines.}
We compare our models with the state-of-the-art point-based models~\cite{qi2017pointnet,huang2018recurrent,wang2018dynamic,li2018pointcnn} and the voxel-based baseline~\cite{cicek20163d}. Following Tchapmi~\etal~\cite{tchapmi2017segcloud} and Li~\etal~\cite{li2018pointcnn}, we train the models on area 1,2,3,4,6 and test them on area 5 because it is the only one that does not overlap with any other area. Both data processing and evaluation protocol are the same as PointCNN~\cite{li2018pointcnn} for fair comparison. We measure the latency and memory consumption with 32768 points per batch on a single NVIDIA GTX 1080 Ti GPU.

\myparagraph{Results.}
As in \tab{tab:exp:s3dis:main}, PVCNN improves its backbone (\ie, PointNet) by more than \textbf{13\%} in mIoU and outperforms DGCNN (which involves sophisticated graph convolutions) by a large margin in both accuracy and latency. Remarkably, our PVCNN++ outperforms the state-of-the-art point-based model (PointCNN) by 1.7\% in mIoU with \textbf{4}$\times$ lower latency, and the voxel-based baseline (3D-UNet) by 4\% in mIoU with more than \textbf{8$\times$} lower latency and GPU memory consumption. Similar to object part segmentation, we also design compact models by reducing the number of channels in our PVCNN to 12.5\%, 25\% and 50\% and our PVCNN++ to 50\%. Remarkably, the narrower version of our PVCNN outperforms DGCNN with \textbf{15$\times$} measured speedup, and RSNet with \textbf{9$\times$} measured speedup. Furthermore, it achieves 4\% improvement in mIoU over PointNet while still being \textbf{2.5$\times$} faster than this extremely efficient 3D model (which does not have any neighborhood aggregation). We refer the readers to \fig{fig:exp:s3dis:tradeoffs} for accuracy \vs latency and accuracy \vs memory trade-offs.

\myparagraph{PV/SPVConv.}
{With the same network structure, PVCNN is more effective than SPVCNN, achieving 1.2-2.1\% higher mIoU (\tab{tab:exp:s3dis:main}). Similar to our results on object segmentation, SPVCNN fails to achieve large speedups with small channel numbers: with 4$\times$ channel reduction, SPVCNN achieves only 1.9$\times$ speedup while PVCNN achieves 4.3$\times$ speedup. Thus, PVCNN is a superior choice for indoor scene segmentation.}

\subsection{3D Outdoor Scene Segmentation}

\subsubsection{SemanticKITTI}

\begin{table}[t]
\renewcommand{\arraystretch}{1.3}
\setlength{\tabcolsep}{5pt}
\small\centering

\caption{Results of Outdoor Scene Segmentation on SemanticKITTI (3D)}
\label{tab:semantickitti:results:3d}

\vspace{-6pt}
\scalebox{0.825}{
\begin{tabular}{lcccc}
    \toprule
     & \#Params (M) & \#MACs (G) & Latency (ms) & mIoU \\
    \midrule
    PointNet~\cite{qi2017pointnet} & 3.0$^{*}$ & -- & 500$^{*}$ & 14.6 \\
    SPGraph~\cite{landrieu2018large} & 0.3$^{*}$ & -- & 5200$^{*}$ & 17.4 \\
    PointNet++~\cite{qi2017pointnet++} & 6.0$^{*}$ & -- & 5900$^{*}$ & 20.1 \\
    TangentConv~\cite{tatarchenko2018tangent} & 0.4$^{*}$ & -- & 3000$^{*}$ & 40.9 \\
    RandLA-Net~\cite{hu2019randla} & 1.2 & 66.5 & 880 (256+624)$^\dag$ & 53.9 \\
    KPConv~\cite{thomas2019kpconv} & 18.3 & 207.3 & -- & 58.8 \\
    MinkowskiNet~\cite{choy20194d} & 21.7 & 114.0 & 294 & 63.1 \\
    \midrule
    PVCNN & 2.5 & 42.4 & 146 & 39.0 \\
    SPVCNN & 21.8 & 118.6 & 317.1 & 65.3 \\
    \midrule
    \textbf{SPVNAS-A} & 2.6 & 15.0 & 110 & 63.7 \\
    \textbf{SPVNAS-B} & 12.5 & 73.8 & 259 & \textbf{66.4} \\
    \bottomrule
\end{tabular}
}

\vspace{8pt}\justifying\noindent
{\footnotesize\itshape{SPVNAS outperforms MinkowskiNet with \textbf{2.7}$\times$ speedup. {$^\dag$: computation time + post-processing time.} $^{*}$: results from Behley~\etal~\cite{behley2019semantickitti}.}}
\end{table}

\begin{table}[t]
\renewcommand{\arraystretch}{1.3}
\setlength{\tabcolsep}{4pt}
\small\centering

\caption{Results of Outdoor Scene Segmentation on SemanticKITTI (2D)}
\label{tab:semantickitti:results:2d}

\vspace{-6pt}
\scalebox{0.825}{
\begin{tabular}{lcccc}
    \toprule
     & \#Params (M) & \#MACs (G) & Latency (ms) & mIoU \\
    \midrule
    DarkNet21Seg~\cite{behley2019semantickitti} & 24.7 & 212.6 & 73 (49+24)$^\dag$ & 47.4 \\
    DarkNet53Seg~\cite{behley2019semantickitti} & 50.4 & 376.3 & 102 (78+24)$^\dag$ & 49.9 \\
    SqueezeSegV3-21~\cite{xu2020squeezesegv3} & 9.4 & 187.5 & 97 (73+24)$^\dag$ & 51.6 \\
    SqueezeSegV3-53~\cite{xu2020squeezesegv3} & 26.2 & 515.2 & 238 (214+24)$^\dag$ & 55.9 \\
    3D-MiniNet~\cite{alonso20203d} & 4.0 & -- & -- & 55.8 \\
    PolarNet~\cite{zhang2020polarnet} & 13.6 & 135.0 & 62 & 57.2 \\
    SalsaNext~\cite{cortinhal2020salsanext} & 6.7 & 62.8 & 71 (47+24)$^\dag$ & 59.5 \\
    \midrule
    \textbf{SPVNAS} & \textbf{1.1} & \textbf{8.9} & 89 & \textbf{60.3} \\
    \bottomrule
\end{tabular}
}

\vspace{8pt}\justifying\noindent
{\footnotesize\itshape{SPVNAS} outperforms the 2D projection-based methods with at least \textbf{7.1$\times$} computation reduction. {$^\dag$: computation time + projection time.}}
\end{table}
\begin{figure*}[t]
\centering
\captionsetup[subfigure]{position=top}
\subfloat[Trade-Off: Mean IoU \vs \#MACs]{\includegraphics[width=0.49\linewidth]{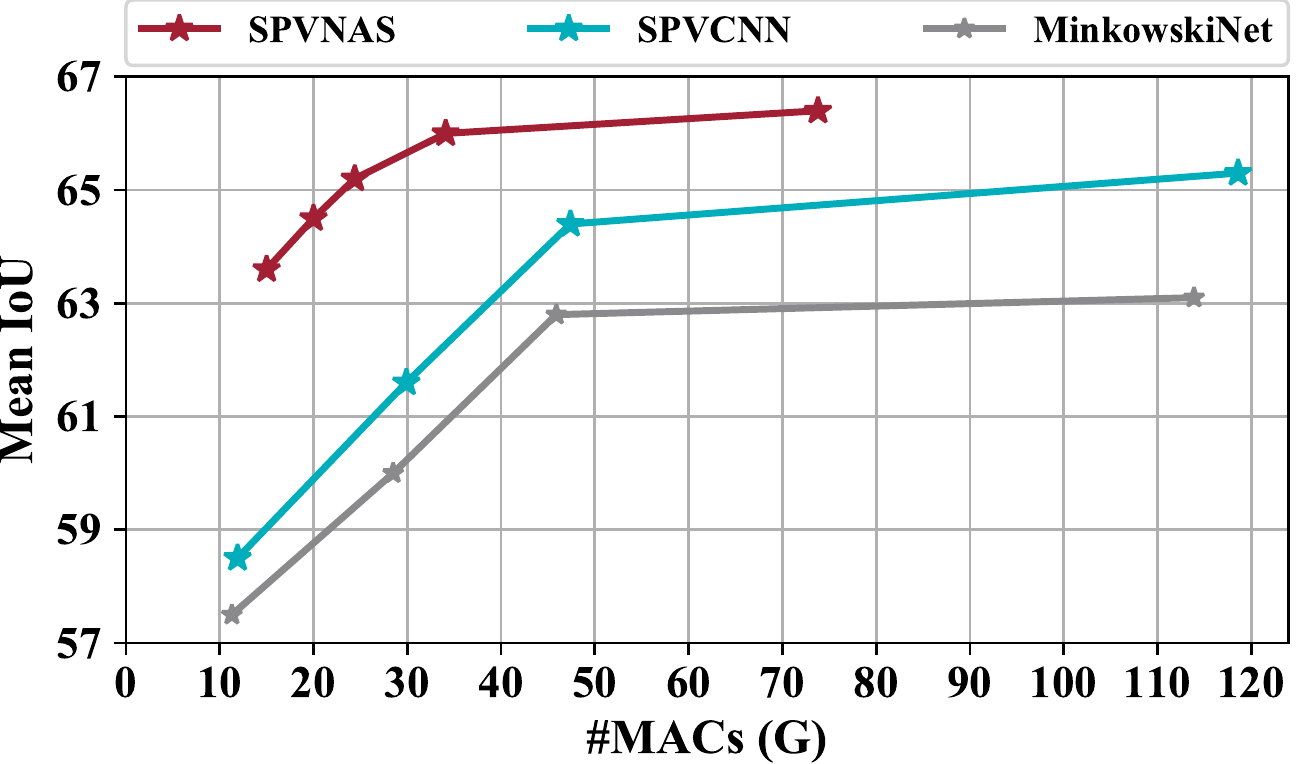}}
\hfil
\subfloat[Trade-Off: Mean IoU \vs GPU Latency]{\includegraphics[width=0.49\linewidth]{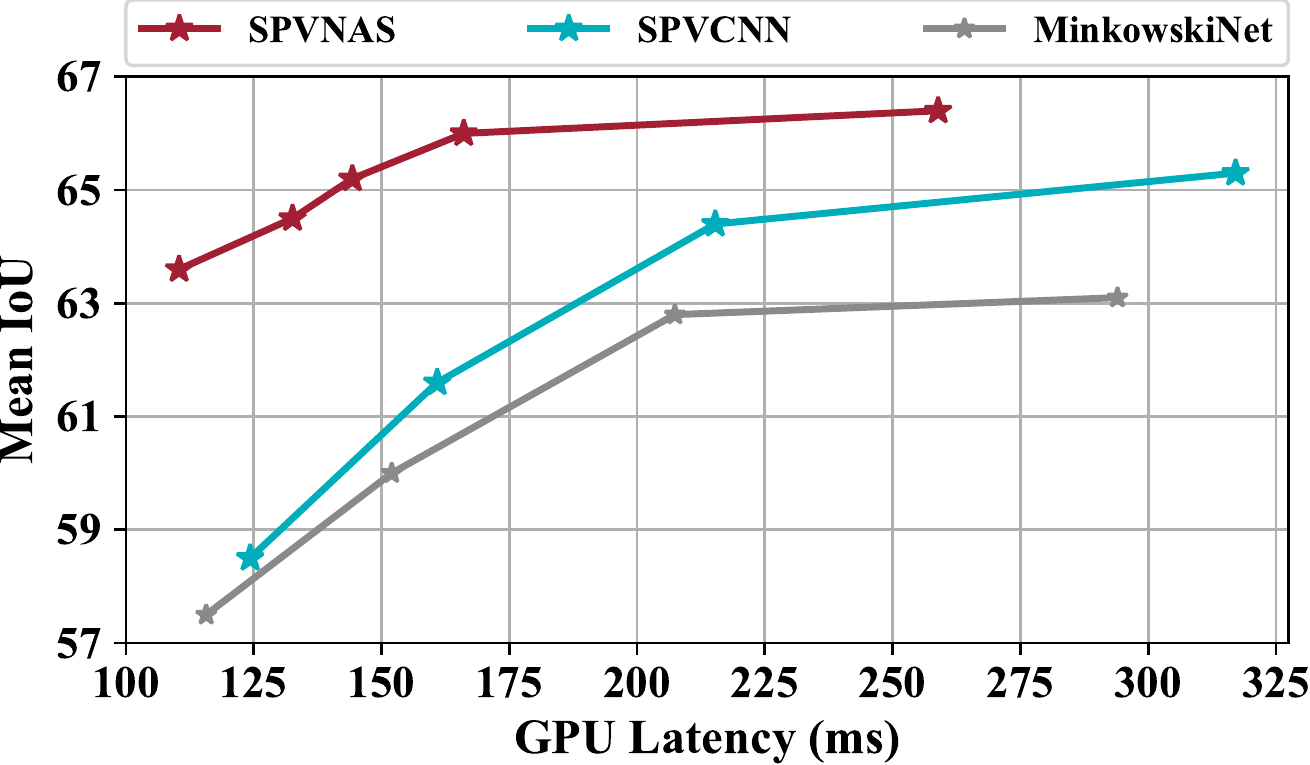}}
\caption{An efficient 3D primitive (SPVConv) and a well-designed network architecture (3D-NAS) are both important to the performance of SPVNAS.}
\label{fig:semantickitti:tradeoffs}
\end{figure*}

\begin{figure*}[!t]
\captionsetup[subfigure]{position=top}
\subfloat[Input Scene]{\begin{tabular}[b]{c}%
\includegraphics[width=0.22\linewidth]{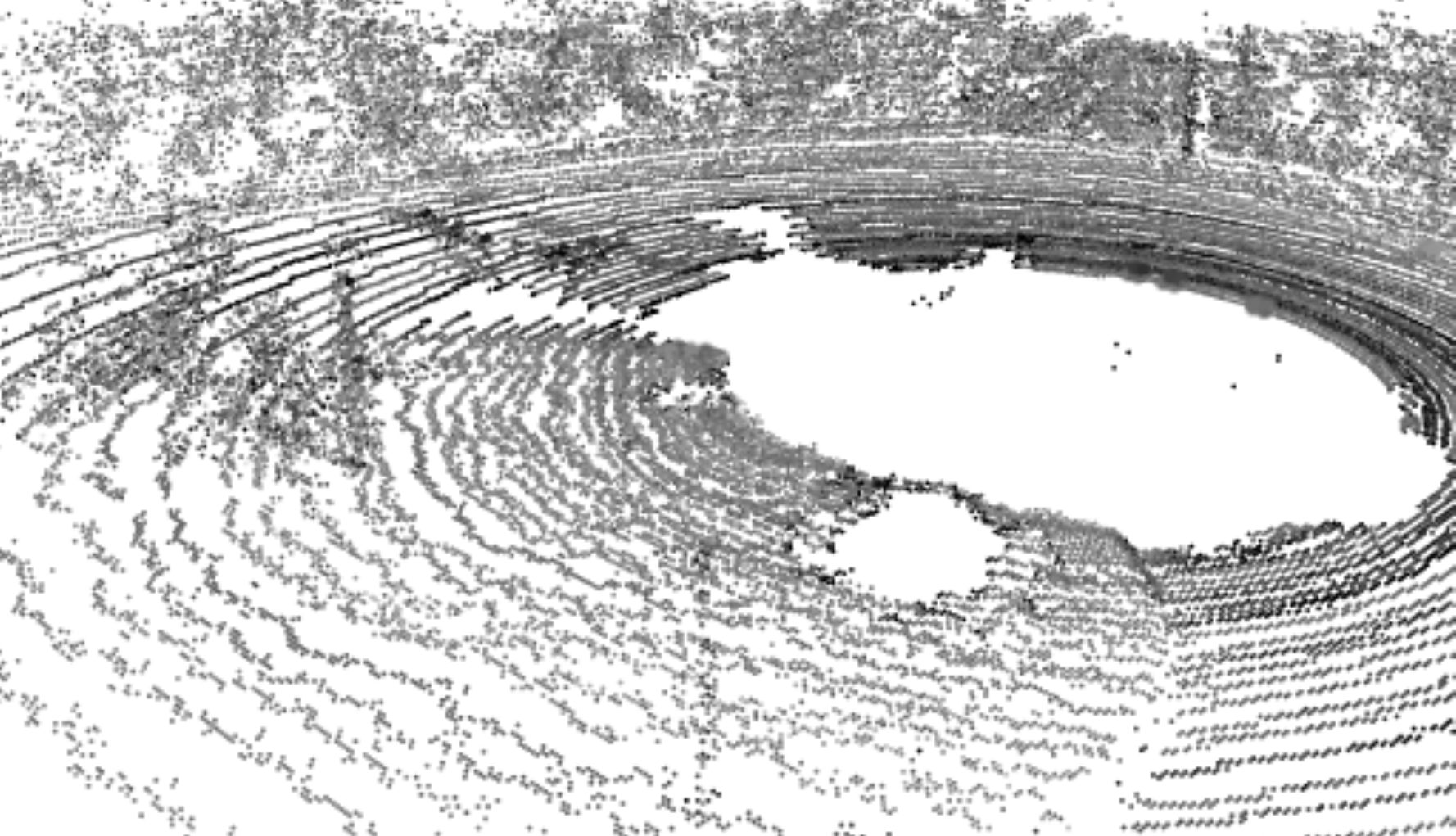}\\
\includegraphics[width=0.22\linewidth]{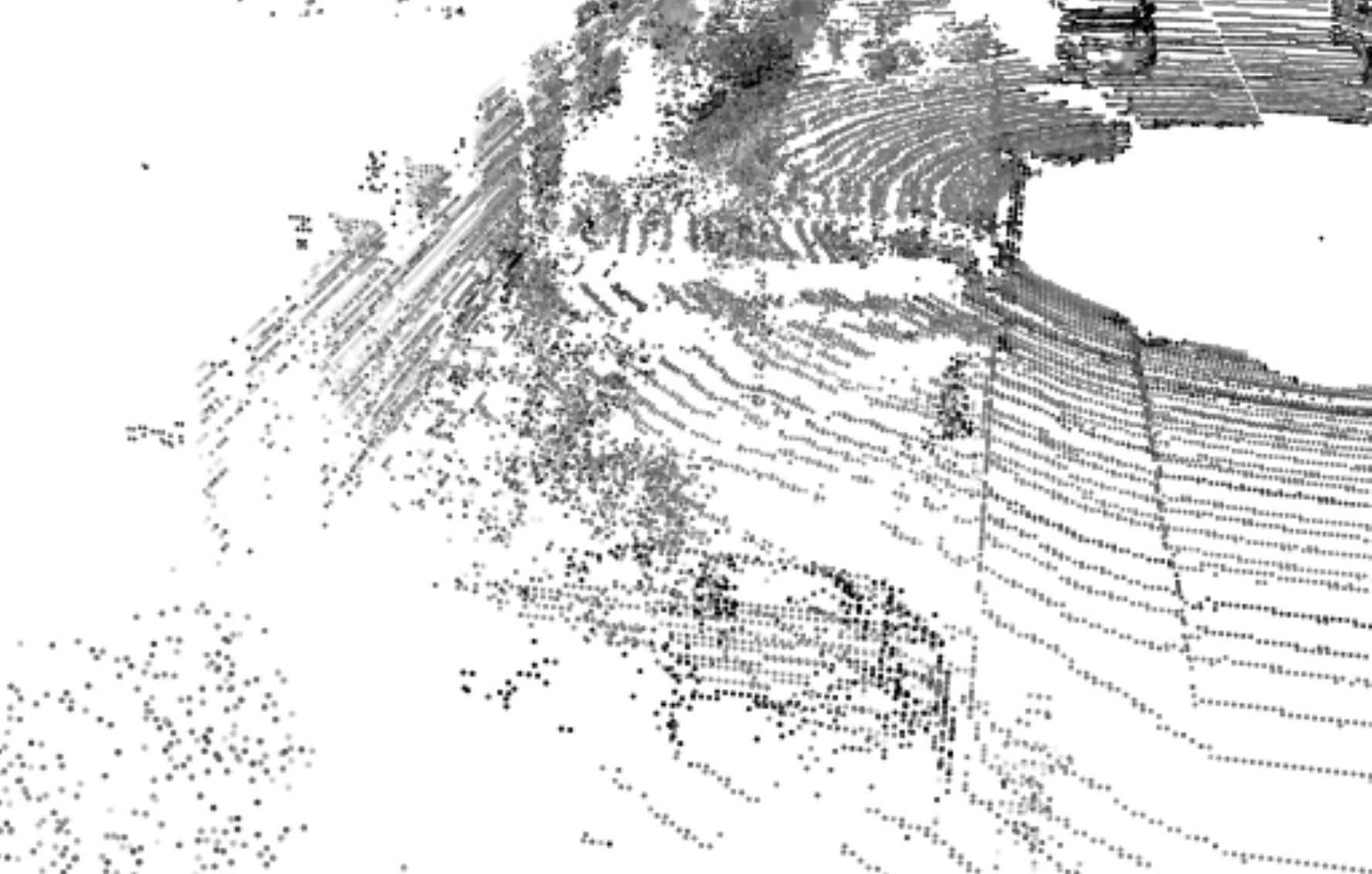}%
\end{tabular}}
\hfil
\subfloat[Error by MinkowskiNet]{\begin{tabular}[b]{c}%
\includegraphics[width=0.22\linewidth]{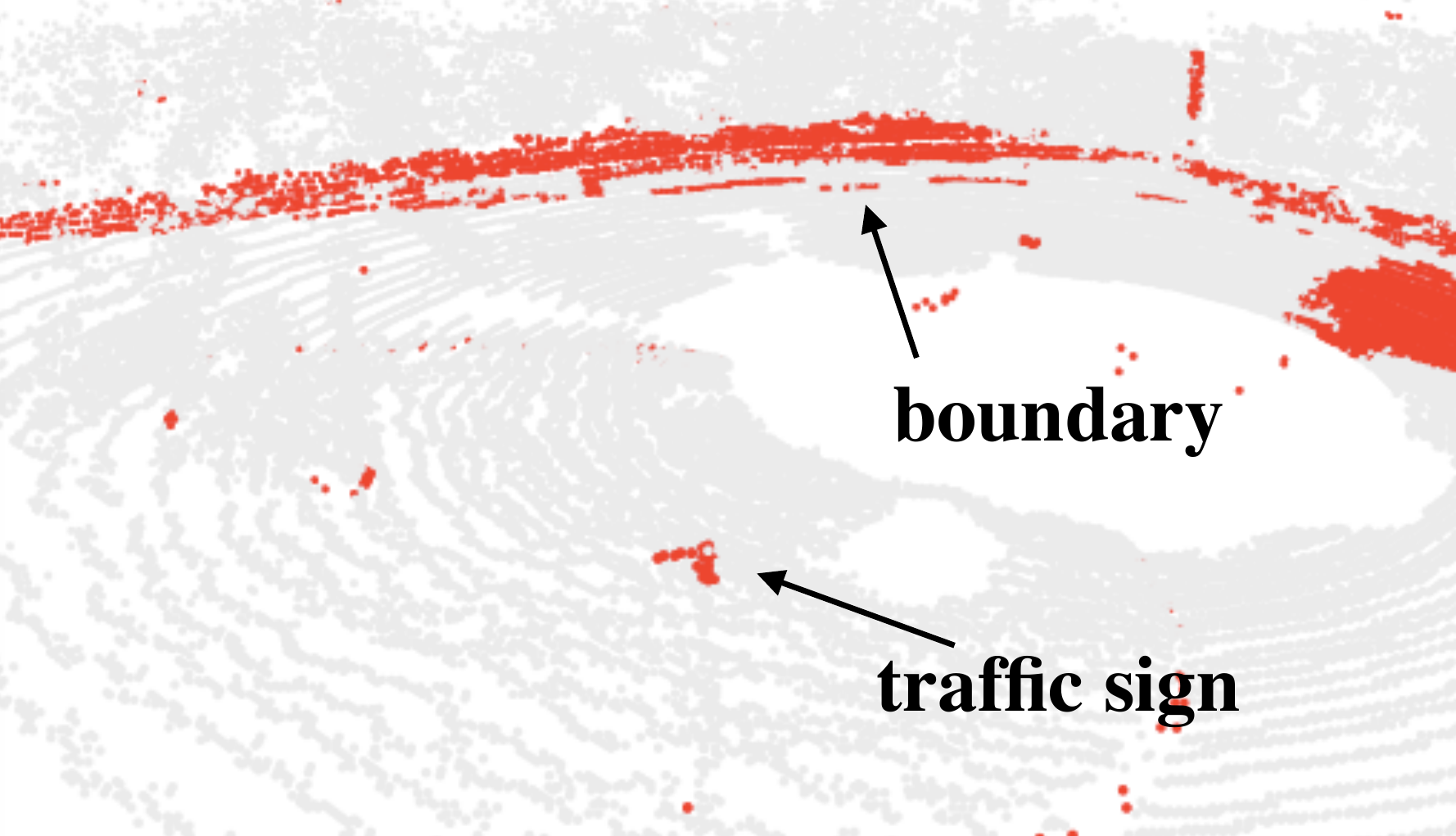}\\
\includegraphics[width=0.22\linewidth]{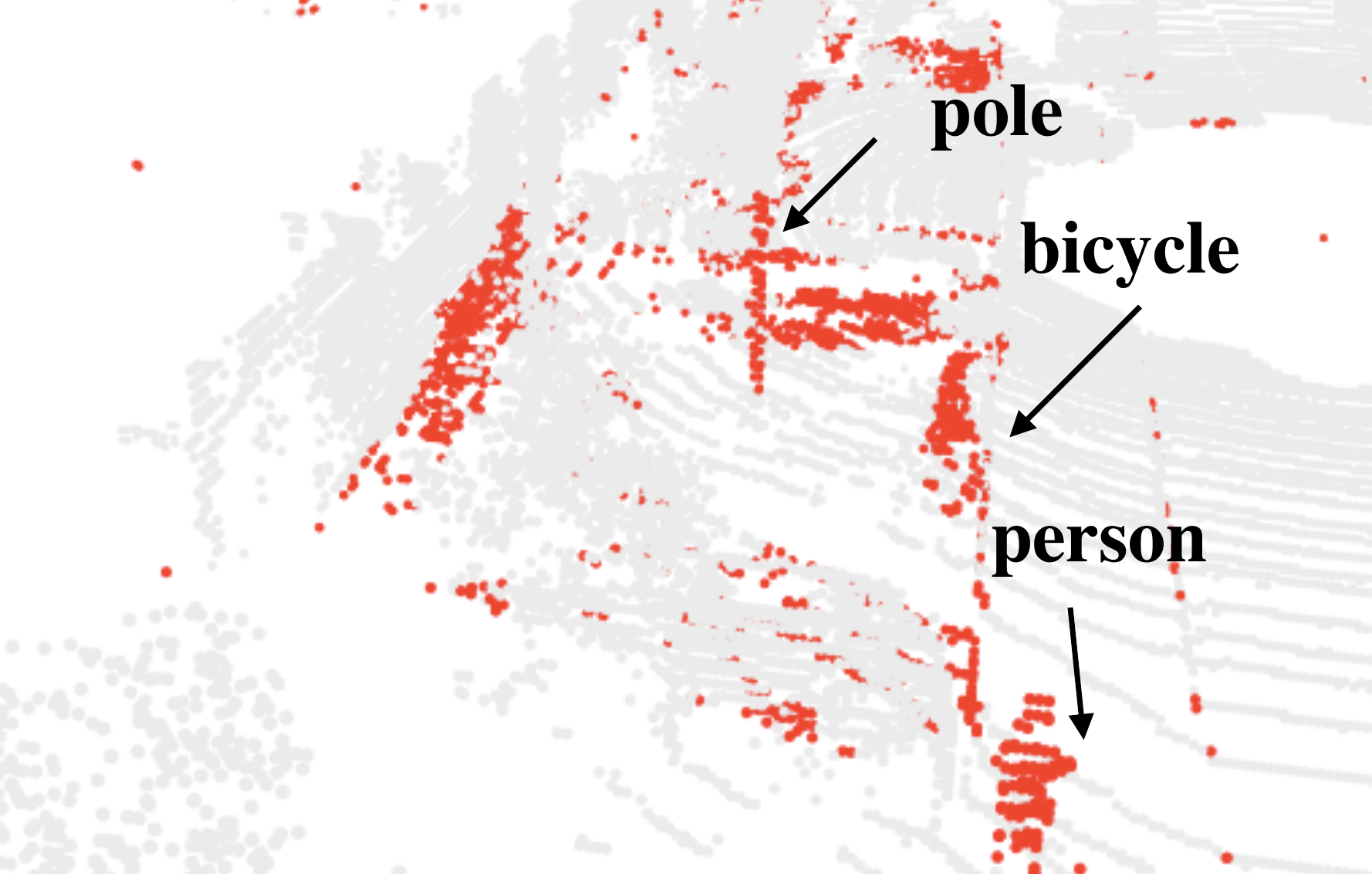}%
\end{tabular}}
\hfil
\subfloat[Less Error by SPVNAS]{\begin{tabular}[b]{c}%
\includegraphics[width=0.22\linewidth]{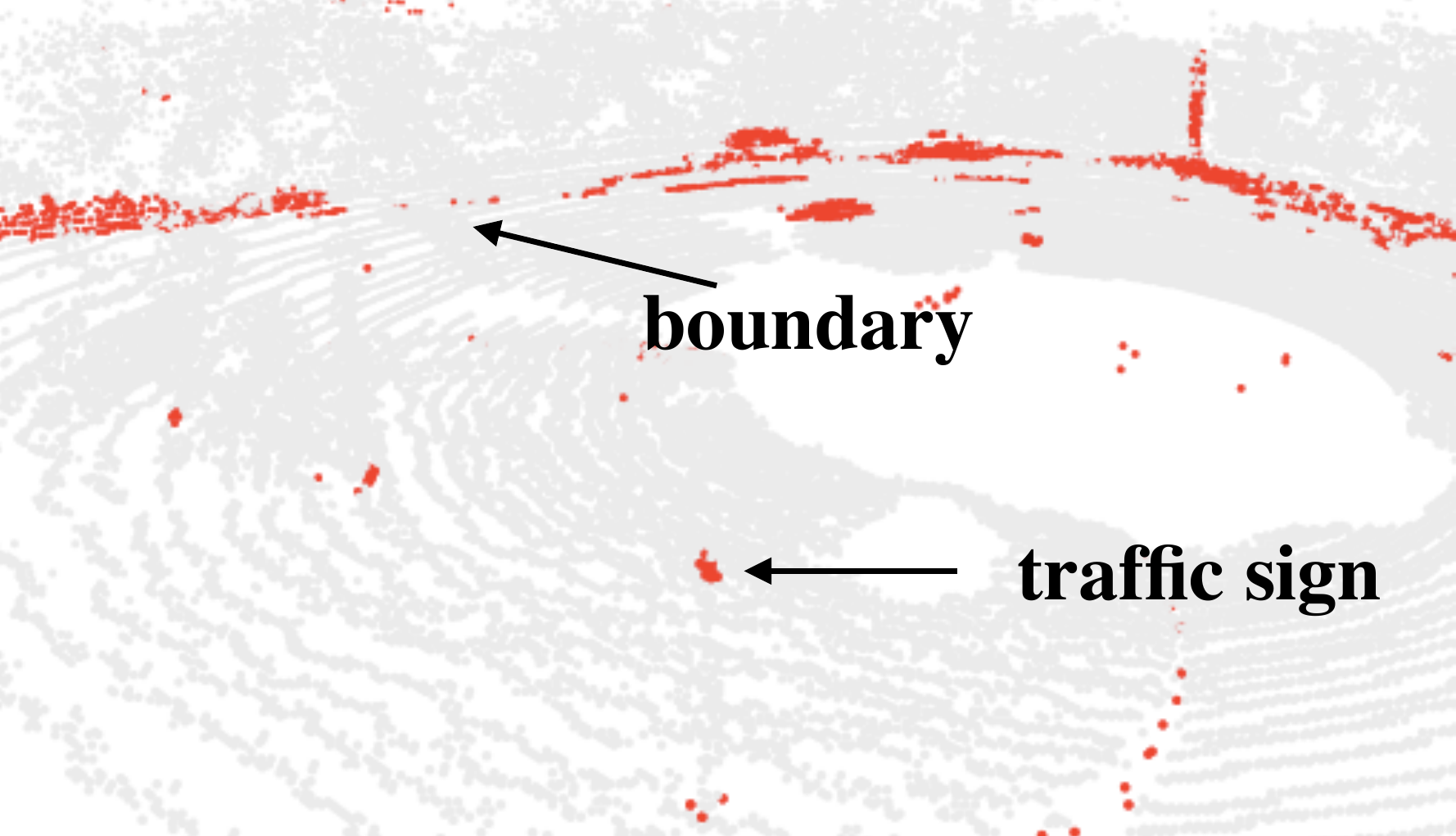}\\
\includegraphics[width=0.22\linewidth]{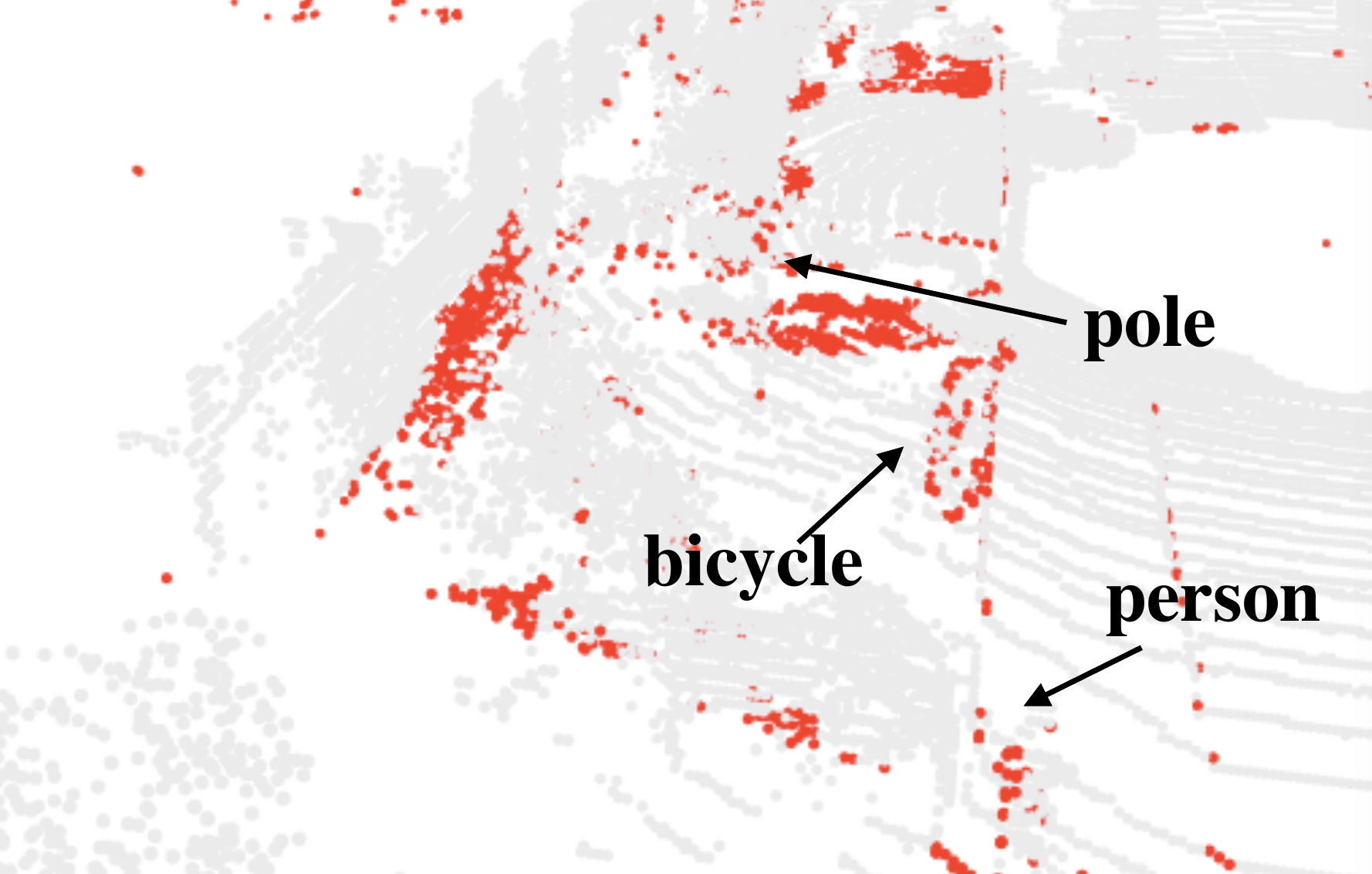}%
\end{tabular}}
\hfil
\subfloat[Ground Truth]{\begin{tabular}[b]{c}%
\includegraphics[width=0.22\linewidth]{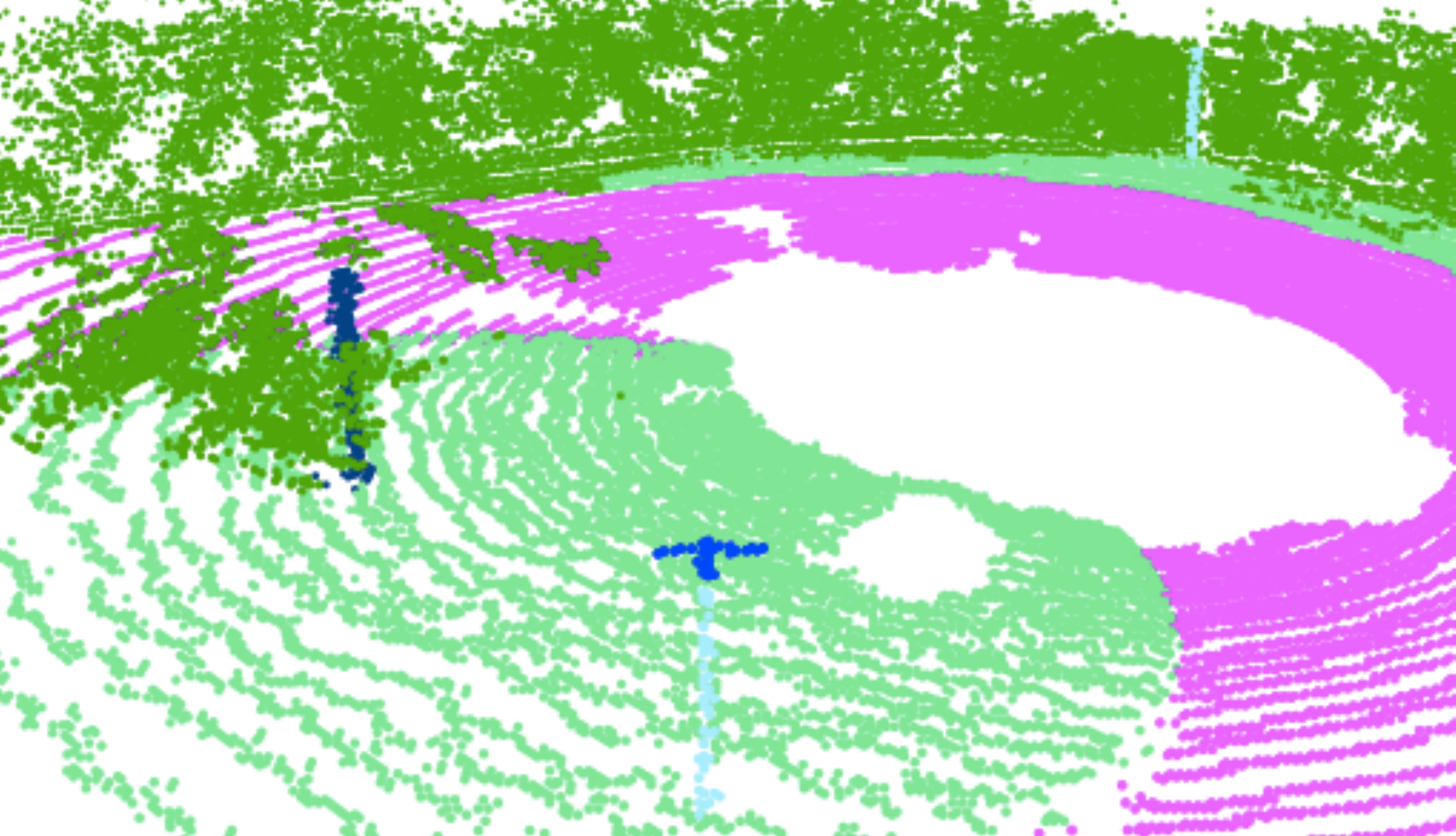}\\
\includegraphics[width=0.22\linewidth]{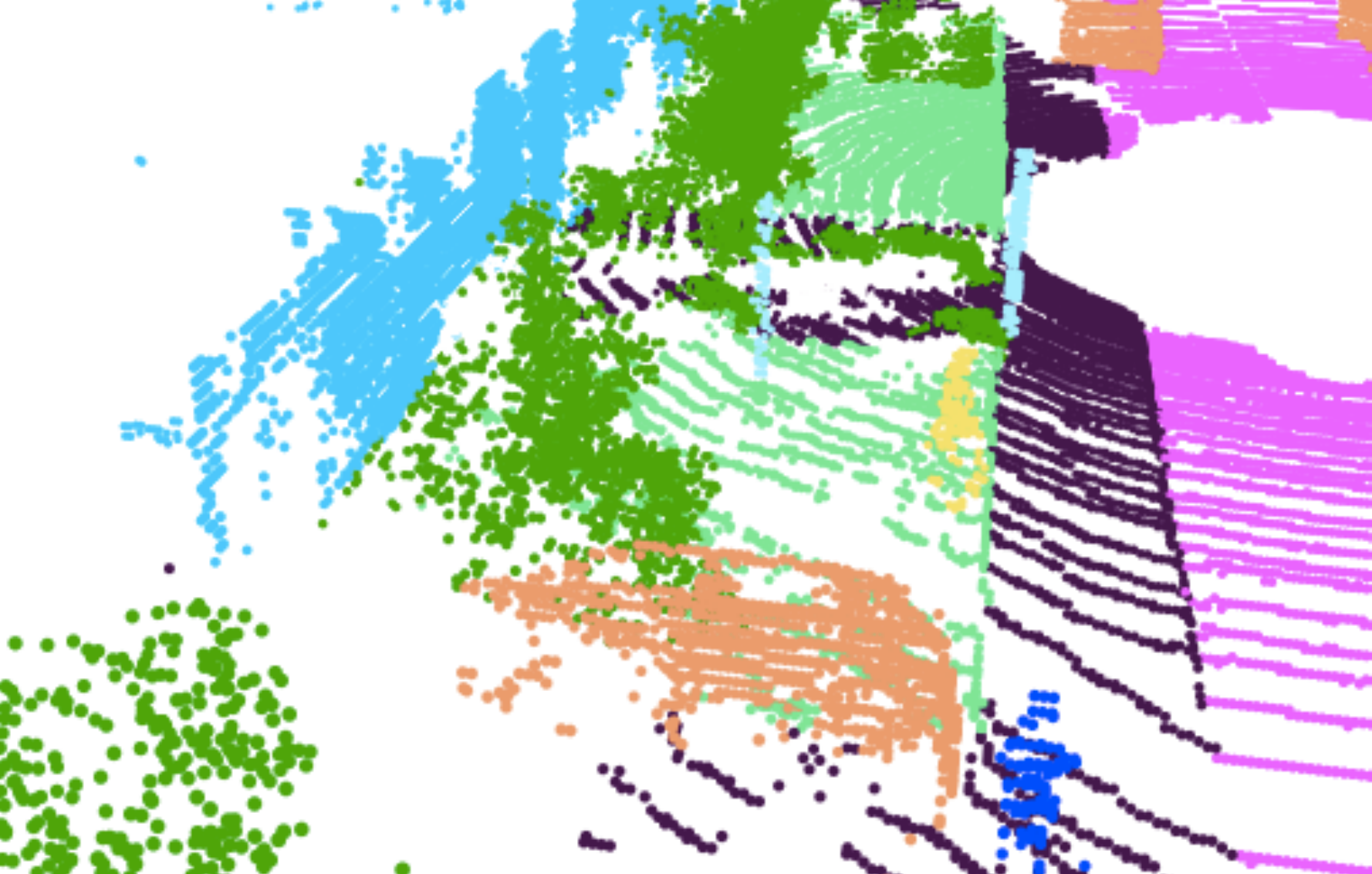}%
\end{tabular}}
\caption{MinkowskiNet usually has a higher error recognizing small objects and region boundaries, while our SPVNAS recognizes small objects better thanks to the high-resolution point-based branch.}
\label{fig:exp:semantickitti:visualizations}
\end{figure*}

We evaluate our method on 3D semantic segmentation and conduct experiments on the large-scale outdoor scene dataset, SemanticKITTI~\cite{behley2019semantickitti}. This dataset contains 23,201 LiDAR point clouds for training and 20,351 for testing, and it is annotated from all 22 sequences in the KITTI~\cite{geiger2013vision} Odometry benchmark. We train all models on the entire training set and report the mean intersection-over-union (mIoU) on the official test set under the single scan setting. We measure the latency on a single NVIDIA GTX 1080 Ti GPU with batch size of 1. We use \texttt{TorchSparse v1.0.0}\footnote{\texttt{\url{https://github.com/mit-han-lab/torchsparse}}} as our inference backend.

\myparagraph{Results.}
As in \tab{tab:semantickitti:results:3d}, SPVNAS outperforms the previous state-of-the-art MinkowskiNet~\cite{choy20194d} by \textbf{3.3\%} in mIoU with 1.7$\times$ model size reduction, 1.5$\times$ computation reduction and 1.1$\times$ measured speedup. Further, we downscale our SPVNAS by setting the resource constraint to 15G MACs. This offers us with a much smaller model that outperforms MinkowskiNet with \textbf{8.3$\times$} model size reduction, \textbf{7.6$\times$} computation reduction, and \textbf{2.7$\times$} measured speedup. In \fig{fig:exp:semantickitti:visualizations}, we also provide qualitative comparisons between SPVNAS and MinkowskiNet: our SPVNAS makes fewer errors for small instances.

\begin{figure*}[!t]
\centering
\captionsetup[subfigure]{position=top}
\subfloat[SECOND]{\begin{tabular}[b]{c}%
\includegraphics[width=0.31\linewidth]{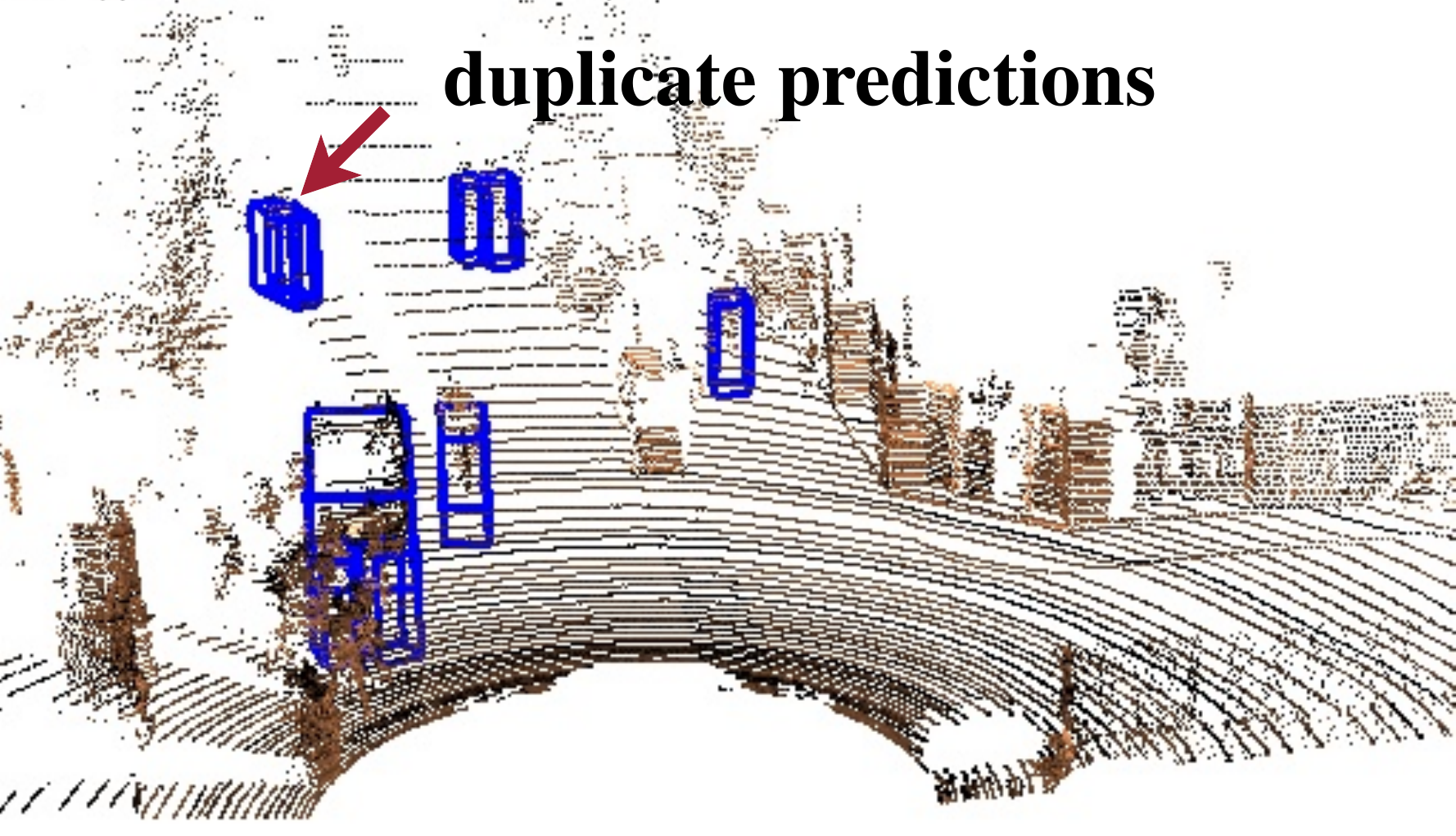}\\
\includegraphics[width=0.31\linewidth]{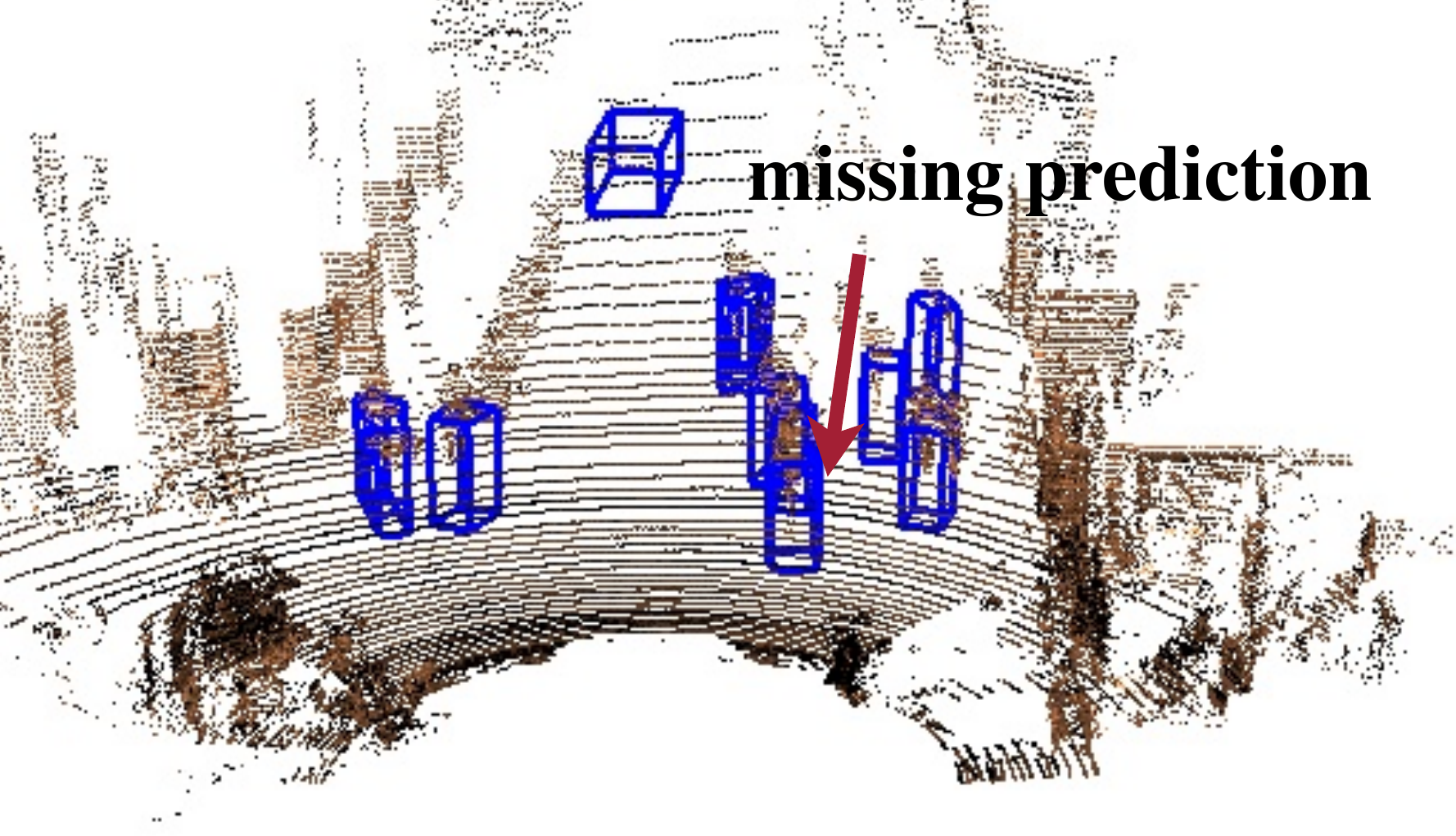}%
\end{tabular}}
\hfil
\subfloat[SPVCNN]{\begin{tabular}[b]{c}%
\includegraphics[width=0.31\linewidth]{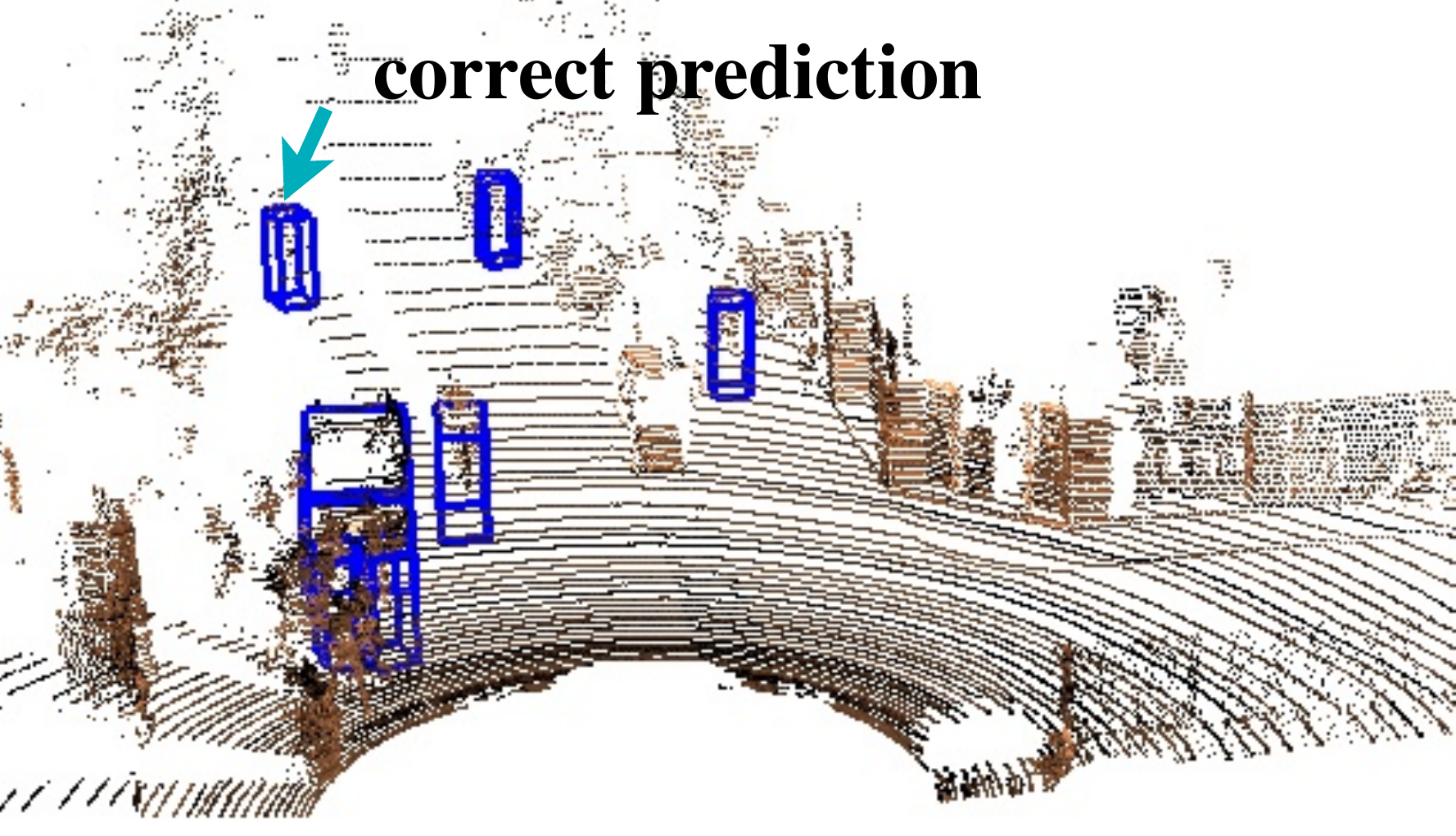}\\
\includegraphics[width=0.31\linewidth]{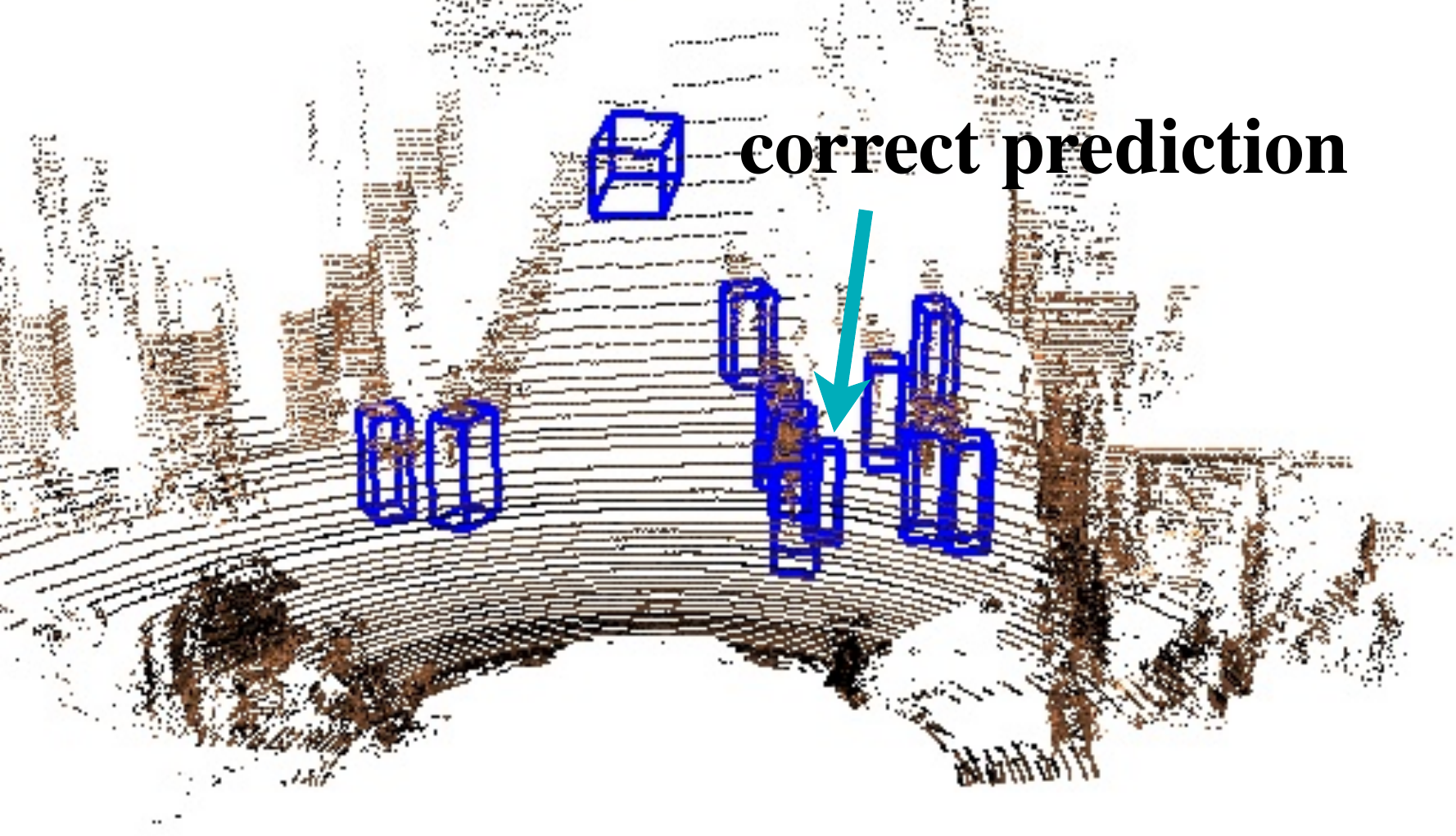}%
\end{tabular}}
\hfil
\subfloat[Ground Truth]{\begin{tabular}[b]{c}%
\includegraphics[width=0.31\linewidth]{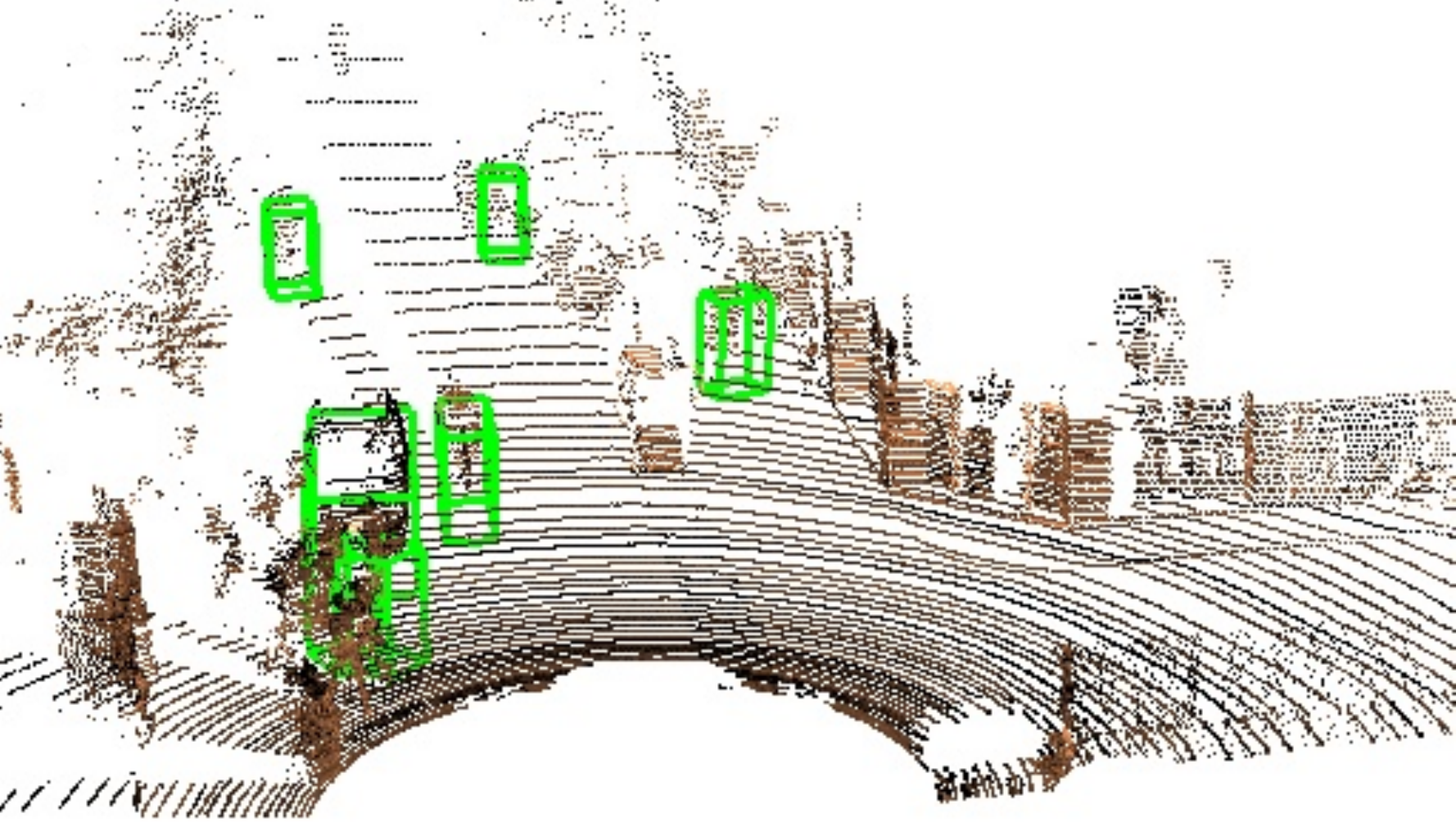}\\
\includegraphics[width=0.31\linewidth]{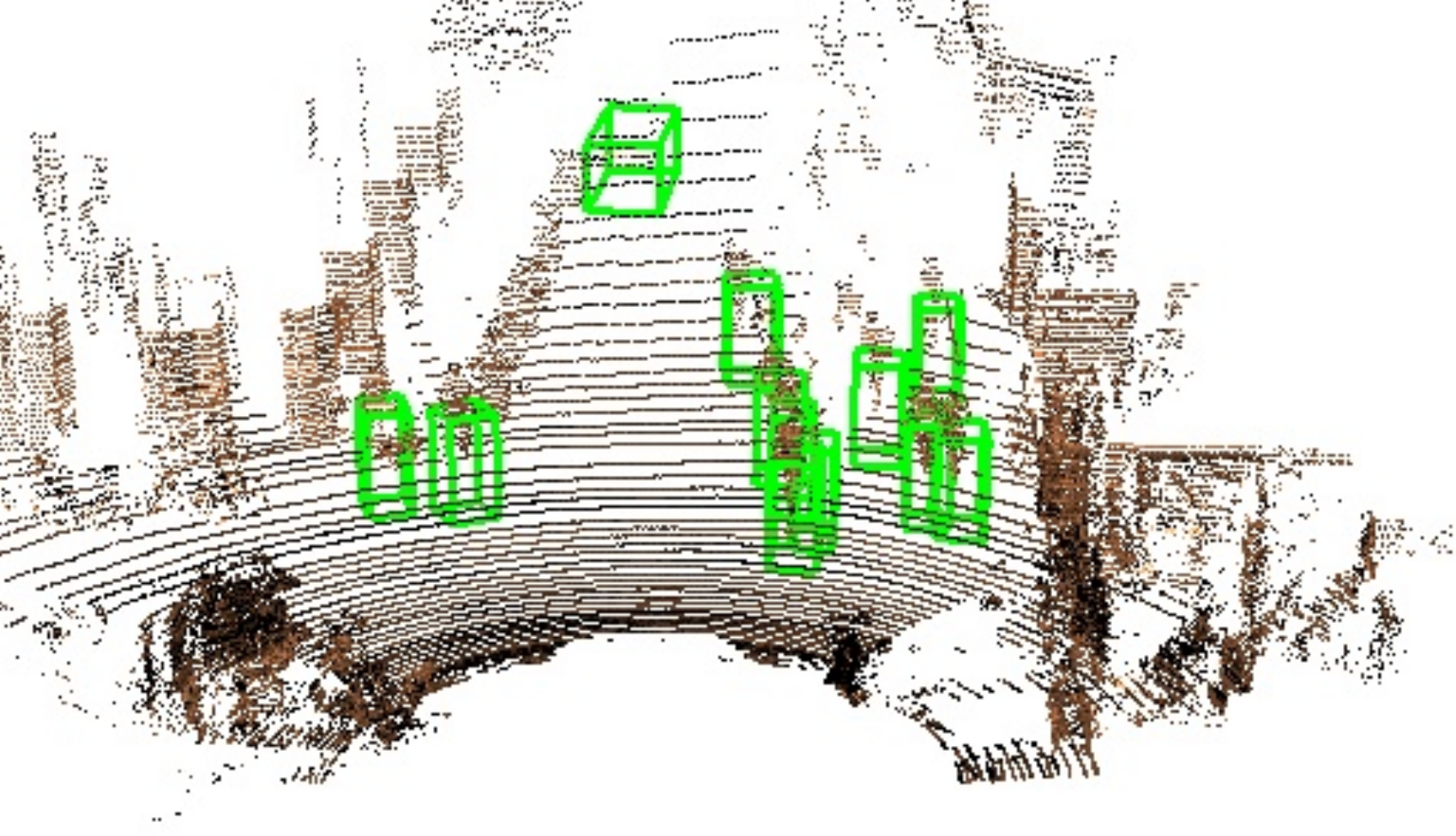}%
\end{tabular}}
\caption{Visualizations of outdoor object detection on KITTI. SPVCNN performs better than SECOND in detecting small objects in crowded scenes.}
\label{fig:exp:kitti:visualizations}
\end{figure*}

In \tab{tab:semantickitti:results:2d}, we compare SPVNAS with 2D projection-based models. With a smaller backbone (by removing the decoder layers), SPVNAS outperforms DarkNets~\cite{behley2019semantickitti} by more than \textbf{10\%} in mIoU with 1.2$\times$ speedup even though 2D CNNs are much better optimized by modern deep learning libraries. Compared with other 2D methods, our SPVNAS achieves at least \textbf{8.5}$\times$ model size reduction and \textbf{15.2}$\times$ computation reduction while being much more accurate. Furthermore, SPVNAS achieves higher mIoU than KPConv~\cite{thomas2019kpconv}, which is the previous state-of-the-art point-based model, with \textbf{17$\times$} model size reduction and \textbf{23$\times$} computation reduction.

\myparagraph{Analysis.}
In \fig{fig:semantickitti:tradeoffs}, we present both mIoU \vs \#MACs and mIoU \vs latency trade-offs, where we uniformly scale the channel numbers in MinkowskiNet and SPVCNN down as our baselines. It can be observed that a better 3D module (SPVNAS) and a well-designed 3D network architecture (3D-NAS) are equally important to the final performance boost. Remarkably, SPVNAS outperforms MinkowskiNet by more than 6\% in mIoU at 110 ms latency. Such a large improvement comes from non-uniform channel scaling and elastic network depth. In these manually-designed models (MinkowskiNet and SPVCNN), 77\% of the total computation is distributed to the upsampling stage. With 3D-NAS, this ratio is reduced to 47-63\%, making the computation more balanced and the downsampling stage (feature extraction) more emphasized.

\begin{table}[t]
\renewcommand{\arraystretch}{1.3}
\setlength{\tabcolsep}{5.5pt}
\small\centering

\caption{Results of Outdoor Scene Segmentation on nuScenes}
\label{tab:exp:nuscenes:results}

\vspace{-6pt}
\scalebox{0.825}{
\begin{tabular}{lcccc}
    \toprule
     & \#Params (M) & \#MACs (G) & Latency (ms) & mIoU \\
    \midrule
    PolarSeg~\cite{zhang2020polarnet}  & 13.6 & 45.6 & \textbf{42.8} &  47.8 \\
    MinkowskiNet~\cite{choy20194d} & 21.7 &  22.2 & 81.0 & 53.7 \\
    \midrule
    \textbf{SPVCNN} & 21.8 & 23.1 & 89.9 &  54.7\\
    \textbf{SPVNAS} & \textbf{9.6}  & \textbf{10.6} & 59.1 &  \textbf{54.7}\\
    \bottomrule
\end{tabular}
}

\vspace{8pt}\justifying\noindent
{\footnotesize\itshape{SPVNAS outperforms MinkowskiNet with \textbf{2.1}$\times$ computation reduction and \textbf{1.4$\times$} measured speedup.}}
\end{table}

\subsubsection{nuScenes}

The distribution of point clouds varies significantly across different sensors. To verify the generalizability of our SPVCNN and SPVNAS, we also conduct experiments on a recent 3D segmentation benchmark, nuScenes~\cite{caesar2020nuscenes}. The dataset contains 34,149 LiDAR point clouds for training and 6,008 for testing, and provides point-level semantic annotations for each point cloud. We follow the official instruction to split the training set into \texttt{train} split (consisting of 28,130 LiDAR point clouds) and \texttt{val} split (consisting of 6,019 LiDAR point clouds). We train all manually-designed models on the \texttt{train} split and evaluate them on the \texttt{val} split. For our SPVNAS, we search the best model on a 20\% holdout \texttt{minival} split from the \texttt{train} split and perform evaluation on the \texttt{val} split. We report the official 31-class mIoU as the evaluation metric. Similar to SemanticKITTI, we measure the latency on an NVIDIA GTX 1080 Ti GPU with batch size of 1.

\myparagraph{Results.}
As in \tab{tab:exp:nuscenes:results}, SPVNAS achieves a better accuracy \vs efficiency trade-off than both PolarSeg~\cite{zhang2020polarnet} and MinkowskiNet~\cite{choy20194d}. It outperforms MinkowskiNet by \textbf{1.0\%} in mIoU with \textbf{2.3}$\times$ model size reduction, \textbf{2.1}$\times$ computation reduction, and \textbf{1.4}$\times$ speedup. It achieves \textbf{6.9\%} mIoU improvement over the projection-based PolarSeg. Similar to our observations on SemanticKITTI, both efficient 3D primitive design (SPVConv, \textbf{+1.0} mIoU) and network architecture search (3D-NAS, \textbf{2.2}$\times$ computation reduction, \textbf{1.5}$\times$ measured speedup) contribute to the accuracy and efficiency boost of SPVNAS.

\begin{table}[!t]
\renewcommand{\arraystretch}{1.3}
\setlength{\tabcolsep}{1.8pt}
\small\centering

\caption{Results of Outdoor Object Detection on KITTI (Two-Stage)}
\label{tab:exp:kitti:pvcnn}

\vspace{-6pt}
\scalebox{0.825}{
\begin{tabular}{lccccccccccc}
    \toprule
    & \multirow{2.5}{*}{\shortstack[m]{Mem. \\ (G)}} & \multirow{2.5}{*}{\shortstack[m]{Lat. \\ (ms)}} & \multicolumn{3}{c}{Car} & \multicolumn{3}{c}{Cyclist} & \multicolumn{3}{c}{Pedestrian} \\
    \cmidrule(lr){4-6}\cmidrule(lr){7-9}\cmidrule(lr){10-12}
    & & & Easy & Mod. & Hard & Easy & Mod. & Hard & Easy & Mod. & Hard \\
    \midrule
    F-PN & 1.3 & 29.1 & 85.2 & 71.6 & 63.8 & 77.1 & 56.5 & 52.8 & 66.4 & 56.9 & 50.4 \\
    F-PN++ & 2.0 & 105.2 & 84.7 & 72.0 & 64.2 & 75.6 & 56.7 & 53.3 & 68.4 & 60.0 & 52.6 \\
    \midrule
    \textbf{PVCNN} & 1.4 & 58.9 & \textbf{85.3} & \textbf{72.1} & \textbf{64.2} & \textbf{78.1} & \textbf{57.5} & \textbf{53.7} & \textbf{70.6} & \textbf{61.2} & \textbf{56.3} \\
    \bottomrule
\end{tabular}
}

\vspace{8pt}\justifying\noindent
{\footnotesize\itshape{PVCNN outperforms F-PointNet++ in all categories significantly with \textbf{1.8$\times$} measured speedup and \textbf{1.4$\times$} memory reduction.}}
\end{table}

\begin{table}[!t]
\renewcommand{\arraystretch}{1.3}
\setlength{\tabcolsep}{3pt}
\small\centering

\caption{Results of Outdoor Object Detection on KITTI (One-Stage)}
\label{tab:exp:kitti:spvcnn}

\vspace{-6pt}
\scalebox{0.825}{
\begin{tabular}{lccccccccc}
    \toprule
     & \multicolumn{3}{c}{Car} & \multicolumn{3}{c}{Cyclist} & \multicolumn{3}{c}{Pedestrian} \\
     \cmidrule(lr){2-4}\cmidrule(lr){5-7}\cmidrule(lr){8-10}
     & Easy & Mod. & Hard & Easy & Mod. & Hard & Easy & Mod. & Hard \\
    \midrule
    SECOND~\cite{yan2018second} & 89.8 & 80.9 & 78.4 & 82.5 & 62.8 & 58.9 & \textbf{68.3} & 60.8 & 55.3 \\
    \midrule
    \textbf{SPVCNN} & \textbf{90.9} & \textbf{81.8} & \textbf{79.2} & \textbf{85.1} & \textbf{63.8} & \textbf{60.1} & 68.2 & \textbf{61.6} & \textbf{55.9} \\
    \bottomrule
\end{tabular}
}

\vspace{8pt}\justifying\noindent
{\footnotesize\itshape{SPVCNN outperforms SECOND in most categories especially in cyclists.}}
\end{table}
\begin{figure*}[!t]
\centering
\captionsetup[subfigure]{position=top}
\subfloat[MIT Driverless]{\begin{tabular}[b]{c}%
\includegraphics[width=0.47\linewidth]{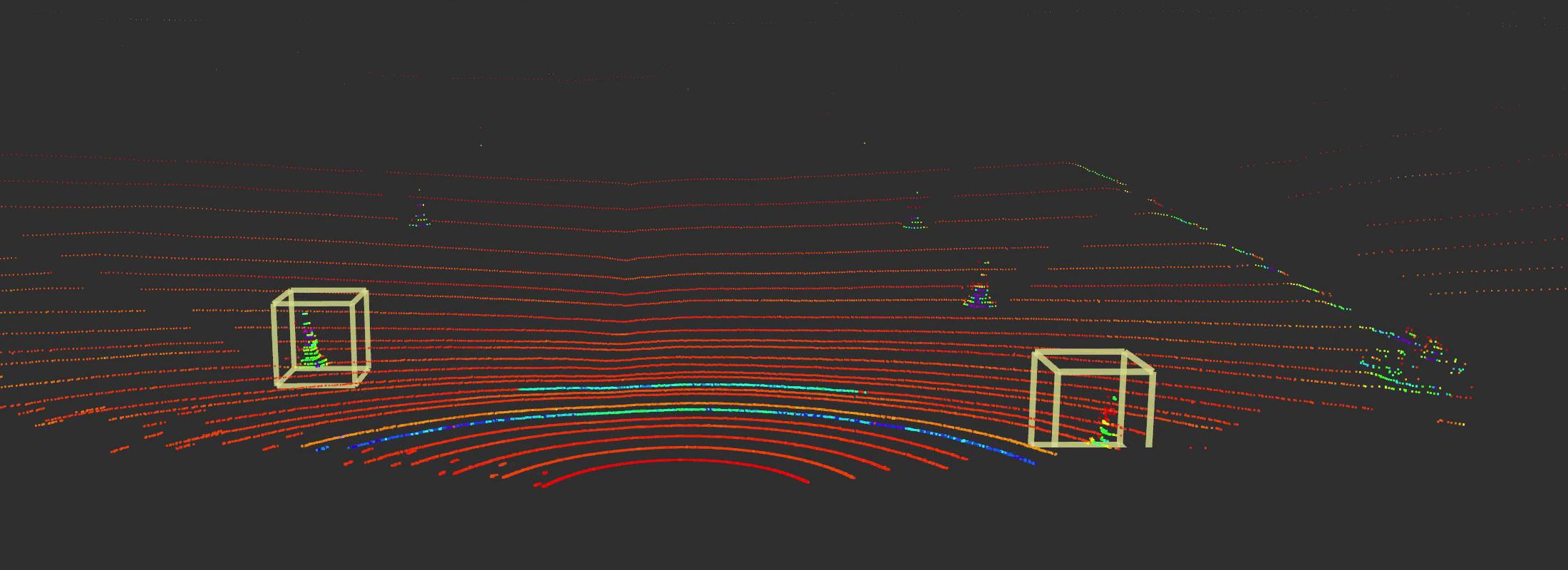}\\
\includegraphics[width=0.47\linewidth]{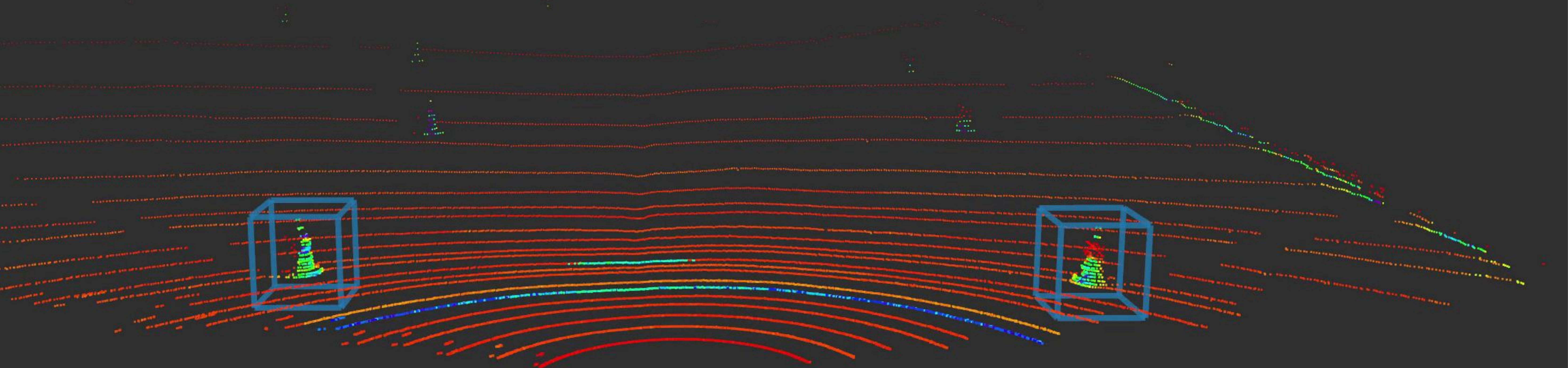}\\
\includegraphics[width=0.47\linewidth]{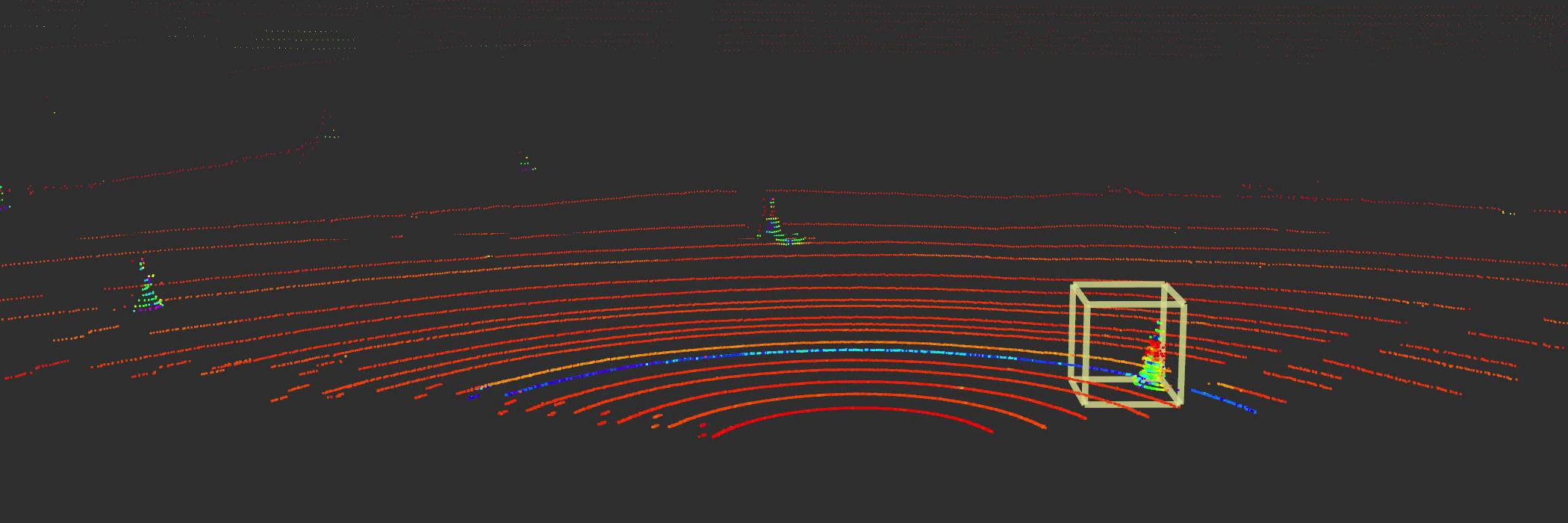}%
\end{tabular}}
\hfil
\subfloat[PVCNN]{\begin{tabular}[b]{c}%
\includegraphics[width=0.47\linewidth]{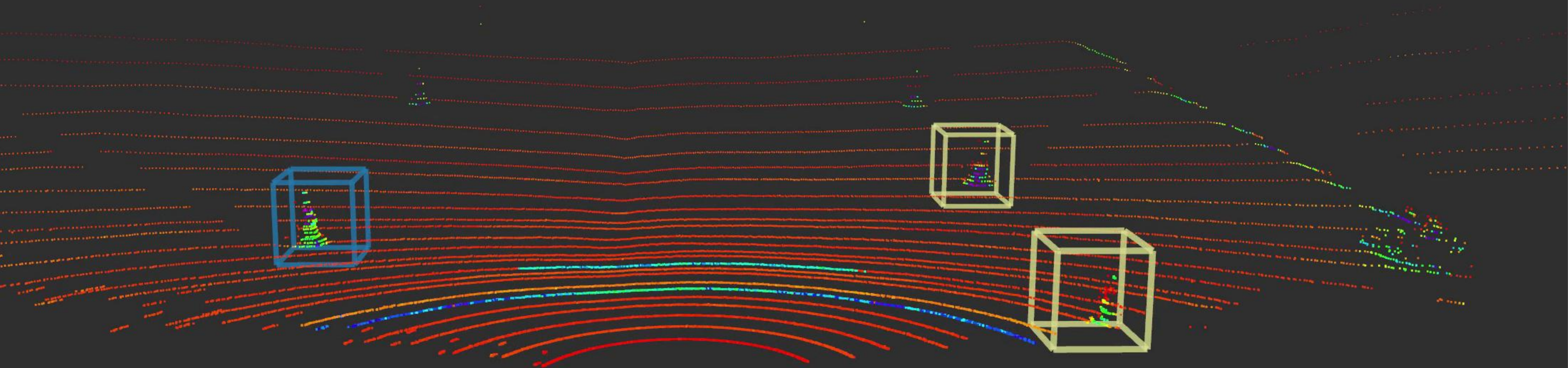}\\
\includegraphics[width=0.47\linewidth]{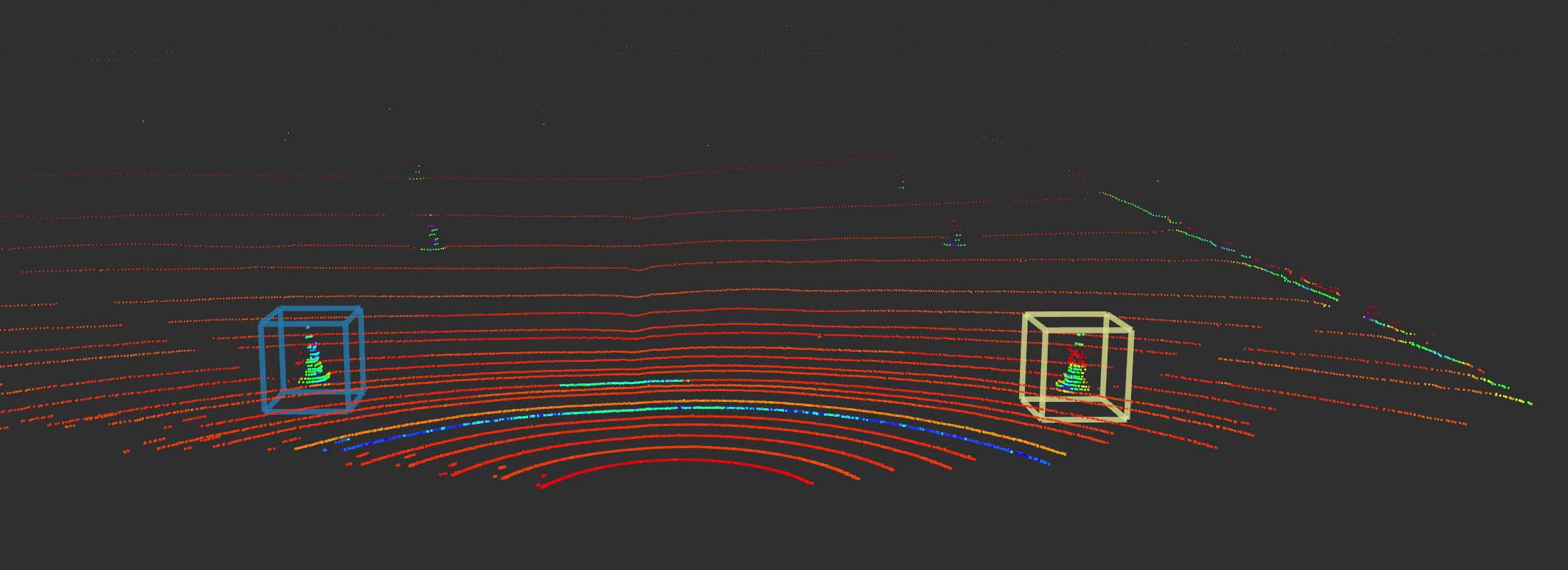}\\
\includegraphics[width=0.47\linewidth]{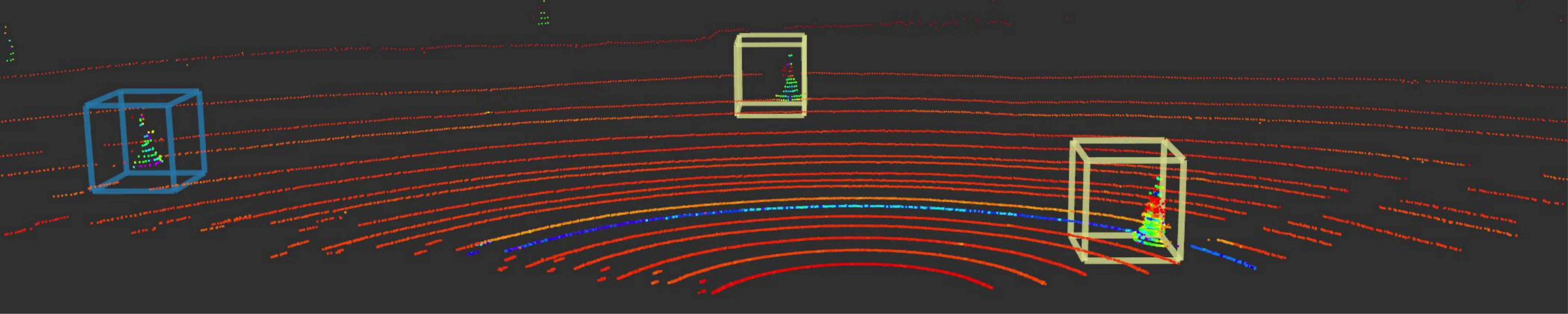}%
\end{tabular}}
\caption{PVCNN is much more accurate: \eg, MIT Driverless' model misclassifies the right traffic cone in the second example as blue. PVCNN also supports larger detection range (see the first and the third example), enabling the racing vehicle to confidently run at a higher speed.}
\label{fig:applications:driving:visualizations}
\end{figure*}
\begin{figure}[!t]
\centering
\includegraphics[width=0.48\textwidth]{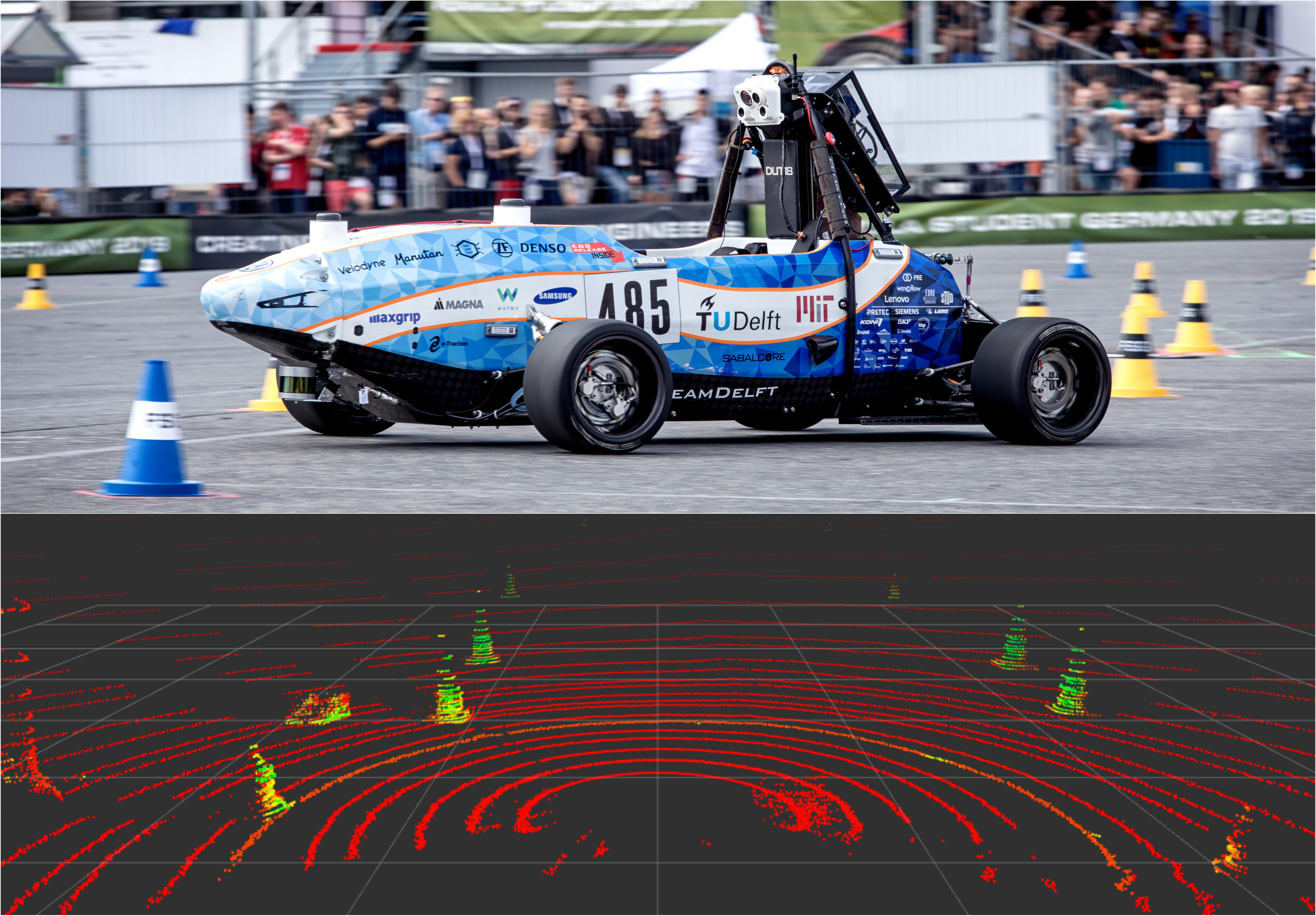}
\caption{Autonomous racing vehicle of MIT Driverless (at the Formula Student competition). Top: vehicle on the racing track (\copyright FSG - Elena Schulz). Bottom: 3D data collected by the Velodyne 32-channel LiDAR sensor mounted on the car.}
\label{fig:applications:driving:teaser}
\end{figure}

\subsection{3D Outdoor Object Detection}

We evaluate our method on 3D object detection and conduct experiments on the large-scale outdoor scene dataset, KITTI~\cite{geiger2013vision}. We follow the conventional training-validation split, where 3,712 samples are used for training and 3,769 samples are left for validation. We report the 3D average precision (with 40 recall positions) on the validation set under IoU thresholds of 0.7 for car, 0.5 for cyclist and pedestrian.

\myparagraph{Results.}
We first apply our method to F-PointNet~\cite{qi2018frustum}, which is a two-stage 3D object detection framework. It first leverages the off-the-shelf 2D object detector to produce frustrum proposals. For each frustrum, it then applies PointNets to classify the content. As the frustrums are relatively small, it is suitable to process them using PVConv. Thus, we build our PVCNN based on F-PointNet by replacing the MLP layers within its instance segmentation network with PVConv and keep its box proposal and refinement networks unchanged. We compare our PVCNN with F-PointNet (whose backbone is PointNet) and F-PointNet++ (whose backbone is PointNet++). In \tab{tab:exp:kitti:pvcnn}, even if our PVCNN does not aggregate neighboring features in the box estimation network while F-PointNet++ does, PVCNN still outperforms it in all classes with \textbf{1.8$\times$} lower latency. Remarkably, our model achieves \textbf{2.4\%} average mAP improvement in the most challenging pedestrian class. Compared with F-PointNet, our PVCNN obtains up to \textbf{4-5\%} mAP improvement in pedestrians, which indicates that our model is both efficient and expressive.

Apart from the frustrum-based framework, we also apply our method to SECOND~\cite{yan2018second}, which is a single-stage 3D object detection framework. It consists of a sparse encoder using 3D sparse convolutions and a region proposal network that performs 2D convolutions after projecting the encoded features into the bird's-eye view (BEV). Unlike F-PointNet, SECOND is applied to the entire outdoor scenes, which are of large scale. Therefore, we choose to replace the sparse encoder with SPVCNN. As a fair comparison, we reimplement and retrain SECOND, which already outperforms the results claimed in the original paper~\cite{yan2018second}. As summarized in \tab{tab:exp:kitti:spvcnn}, our SPVCNN achieves consistent improvements in all categories, especially in cyclist and pedestrian. This is because the high-resolution point-based branch carries more information for small objects. We also provide qualitative results in \fig{fig:exp:kitti:visualizations}, where SPVCNN excels in detecting small objects in crowded scenes.

\begin{table}[!t]
\renewcommand{\arraystretch}{1.3}
\setlength{\tabcolsep}{4pt}
\small\centering

\caption{Results of Cone Classification for Autonomous Racing Vehicle}
\label{tab:applications:driving:results}

\vspace{-6pt}
\scalebox{0.825}{
\begin{tabular}{lcccc}
\toprule
& \#Params (M) & \#MACs (G) & Latency (ms) & Accuracy (\%)\\
\midrule
{MIT Driverless} & {--} & {--} & {14.0} & {95.0}\\
{PointNet~\cite{qi2017pointnet}} & {2.0} & {0.07} & {\textbf{1.4}} & {96.6}\\
{PointNet++~\cite{qi2017pointnet++}} & {1.7} & {4.0} & {53.5} & {97.0}\\
{DGCNN~\cite{wang2018dynamic}} & {0.6} & {0.03} & {2.1} & {98.0}\\
{\textbf{PVCNN}} & {\textbf{0.5}} & {\textbf{0.03}} & {2.3} & {\textbf{98.8}}\\
\bottomrule
\end{tabular}
}

\vspace{8pt}\justifying\noindent
{\footnotesize\itshape{{Compared with existing learning-based solutions, PVCNN achieves higher accuracy with low latency. Here, we measure the GPU latency on NVIDIA GTX 1080 Ti based on an average of 8 cones per scene.}}}
\end{table}

\section{Applications}

The autonomous racing vehicle is an ideal testbed of autonomous driving systems. Given the image and LiDAR inputs, the system needs to operate the car to drive on the racing track as fast as possible (and avoid the obstacles). As in \fig{fig:applications:driving:teaser}, the racing track is surrounded by traffic cones in the Formula Student competition. In particular, there are three types of traffic cones, the blue ones representing the left boundary of the track, the yellow ones defining the right boundary of the track, and the orange ones marking the start and finish of the track. A typical solution is composed of perception, state estimation, path planning and control. Among these stages, the perception model serves as the eyes of the racing vehicle: it needs to localize and recognize the traffic cones accurately in order to segment the (drivable) racing track. The model needs to be very fast as its latency limits the velocity of the racing car; it also needs to be very accurate in order to prevent the car from driving outside the racing track and hitting the landmarks (2 seconds of penalty for each cone hit by the rules of the Formula Student competition). Therefore, we aim to improve the efficiency and accuracy of the LiDAR perception model with our PVCNN.

For the LiDAR perception pipeline, MIT Driverless adopts the clustering algorithm to localize the traffic cones on the racing track and then classifies the type of each traffic cone based on their LiDAR features. Due to the limited hardware resources on the car, their team previously designed a simple 1D feature-based neural network as their cone classification model. Though simple, it achieves moderate classification accuracy of 95\%, which helped their team win the second and third places in two of the largest 2019 Formula Student competitions. However, this simple perception model can only support the detection range of 8 meters, limiting the velocity of their car to only 5 meters per second. Together with MIT Driverless, we replace their traffic cone classification model with our PVCNN. As in \tab{tab:applications:driving:results}, {our PVCNN accelerates their previous pipeline by \textbf{6$\times$} while achieving almost perfect classification accuracy (\textbf{98.8\%}). PVCNN also outperforms other learning-based solutions such as PointNet (+2.2\%), PointNet++ (+1.8\%) and DGCNN (+0.8\%).} In \fig{fig:applications:driving:visualizations}, we also visualize the traffic cone classification results of their original solution and our PVCNN, from which, our PVCNN is indeed more accurate and supports larger detection range, enabling the racing vehicle to run at a higher speed. MIT Driverless has deployed our PVCNN into their latest system.

\section{Conclusion}

In this paper, we studied 3D deep learning from the hardware efficiency perspective. We systematically analyzed the bottlenecks of previous point-based and voxel-based models and proposed a novel hardware-efficient 3D primitive to combine the best from both. We further enhanced this primitive with the sparse convolution to make it more effective for large (outdoor) scenes. Based on our designed 3D primitive, we then introduced the 3D neural architecture search framework to automatically explore the best network architecture that satisfies a given resource constraint. Extensive experiments on multiple 3D benchmark datasets validate the efficiency and effectiveness of our proposed method. Finally, we also deployed our model to the autonomous racing vehicle of MIT Driverless, achieving larger detection range, higher accuracy and lower latency. We hope that this paper will enable more real-world applications and inspire future research in the direction of efficient 3D deep learning.

\section*{Acknowledgments}

This work is supported by NSF CAREER Award \#1943349, MIT Quest for Intelligence, MIT-IBM Watson AI Lab, Samsung, Hyundai and SONY. We thank AWS Machine Learning Research Awards for providing the computational resource.

\bibliographystyle{IEEEtran}
\bibliography{reference}

\begin{thebibliography}{10}
\providecommand{\url}[1]{#1}
\csname url@samestyle\endcsname
\providecommand{\newblock}{\relax}
\providecommand{\bibinfo}[2]{#2}
\providecommand{\BIBentrySTDinterwordspacing}{\spaceskip=0pt\relax}
\providecommand{\BIBentryALTinterwordstretchfactor}{4}
\providecommand{\BIBentryALTinterwordspacing}{\spaceskip=\fontdimen2\font plus
\BIBentryALTinterwordstretchfactor\fontdimen3\font minus
  \fontdimen4\font\relax}
\providecommand{\BIBforeignlanguage}[2]{{%
\expandafter\ifx\csname l@#1\endcsname\relax
\typeout{** WARNING: IEEEtran.bst: No hyphenation pattern has been}%
\typeout{** loaded for the language `#1'. Using the pattern for}%
\typeout{** the default language instead.}%
\else
\language=\csname l@#1\endcsname
\fi
#2}}
\providecommand{\BIBdecl}{\relax}
\BIBdecl

\bibitem{riegler2017octnet}
G.~Riegler, A.~O. Ulusoy, and A.~Geiger, ``{OctNet: Learning Deep 3D
  Representations at High Resolutions},'' in \emph{Proc. IEEE Conf. Comput.
  Vis. Pattern Recognit.}, 2017.

\bibitem{cicek20163d}
O.~Cicek, A.~Abdulkadir, S.~S. Lienkamp, T.~Brox, and O.~Ronneberger, ``{3D
  U-Net: Learning Dense Volumetric Segmentation from Sparse Annotation},'' in
  \emph{Proc. Medical Image Computing and Computer Assisted Intervention},
  2016.

\bibitem{qi2017pointnet}
C.~R. Qi, H.~Su, K.~Mo, and L.~J. Guibas, ``{PointNet: Deep Learning on Point
  Sets for 3D Classification and Segmentation},'' in \emph{Proc. IEEE Conf.
  Comput. Vis. Pattern Recognit.}, 2017.

\bibitem{qi2017pointnet++}
C.~R. Qi, L.~Yi, H.~Su, and L.~J. Guibas, ``{PointNet++: Deep Hierarchical
  Feature Learning on Point Sets in a Metric Space},'' in \emph{Proc. Adv.
  Neural Inf. Process. Syst.}, 2017.

\bibitem{klokov2017escape}
R.~Klokov and V.~S. Lempitsky, ``{Escape from Cells: Deep Kd-Networks for the
  Recognition of 3D Point Cloud Models},'' in \emph{Proc. IEEE Int. Conf.
  Comput. Vis.}, 2017.

\bibitem{li2018pointcnn}
Y.~Li, R.~Bu, M.~Sun, W.~Wu, X.~Di, and B.~Chen, ``{PointCNN: Convolution on
  $\mathcal{X}$-Transformed Points},'' in \emph{Proc. Adv. Neural Inf. Process.
  Syst.}, 2018.

\bibitem{wu20153d}
Z.~Wu, S.~Song, A.~Khosla, F.~Yu, L.~Zhang, X.~Tang, and J.~Xiao, ``{3D
  ShapeNets: A Deep Representation for Volumetric Shapes},'' in \emph{Proc.
  IEEE Conf. Comput. Vis. Pattern Recognit.}, 2015.

\bibitem{chang2015shapenet}
A.~X. Chang, T.~Funkhouser, L.~Guibas, P.~Hanrahan, Q.~Huang, Z.~Li,
  S.~Savarese, M.~Savva, S.~Song, H.~Su, J.~Xiao, L.~Yi, and F.~Yu,
  ``{ShapeNet: An Information-Rich 3D Model Repository},''
  \emph{arXiv:1512.03012}, 2015.

\bibitem{maturana2015voxnet}
D.~Maturana and S.~Scherer, ``{VoxNet: A 3D Convolutional Neural Network for
  Real-Time Object Recognition},'' in \emph{Proc. IEEE/RSJ Int. Conf. Intell.
  Robot. Syst.}, 2015.

\bibitem{qi2016volumetric}
C.~R. Qi, H.~Su, M.~Niessner, A.~Dai, M.~Yan, and L.~J. Guibas, ``{Volumetric
  and Multi-View CNNs for Object Classification on 3D Data},'' in \emph{Proc.
  IEEE Conf. Comput. Vis. Pattern Recognit.}, 2016.

\bibitem{wang2019voxsegnet}
Z.~Wang and F.~Lu, ``{VoxSegNet: Volumetric CNNs for Semantic Part Segmentation
  of 3D Shapes},'' \emph{IEEE Trans. Vis. Comput. Graph}, 2019.

\bibitem{zhou2018voxelnet}
Y.~Zhou and O.~Tuzel, ``{VoxelNet: End-to-End Learning for Point Cloud Based 3D
  Object Detection},'' in \emph{Proc. IEEE Conf. Comput. Vis. Pattern
  Recognit.}, 2018.

\bibitem{wang2017cnn}
P.-S. Wang, Y.~Liu, Y.-X. Guo, C.-Y. Sun, and X.~Tong, ``{O-CNN: Octree-based
  Convolutional Neural Networks for 3D Shape Analysis},'' \emph{ACM Trans.
  Graph.}, 2017.

\bibitem{tatarchenko2018tangent}
M.~Tatarchenko, J.~Park, V.~Koltun, and Q.-Y. Zhou, ``{Tangent Convolutions for
  Dense Prediction in 3D},'' in \emph{Proc. IEEE Conf. Comput. Vis. Pattern
  Recognit.}, 2018.

\bibitem{le2018pointgrid}
T.~Le and Y.~Duan, ``{PointGrid: A Deep Network for 3D Shape Understanding},''
  in \emph{Proc. IEEE Conf. Comput. Vis. Pattern Recognit.}, 2018.

\bibitem{wang2018dynamic}
Y.~Wang, Y.~Sun, Z.~Liu, S.~E. Sarma, M.~M. Bronstein, and J.~M. Solomon,
  ``{Dynamic Graph CNN for Learning on Point Clouds},'' \emph{ACM Trans.
  Graph.}, 2019.

\bibitem{lan2019modeling}
S.~Lan, R.~Yu, G.~Yu, and L.~S. Davis, ``{Modeling Local Geometric Structure of
  3D Point Clouds using Geo-CNN},'' in \emph{Proc. IEEE Conf. Comput. Vis.
  Pattern Recognit.}, 2019.

\bibitem{xu2018spidercnn}
Y.~Xu, T.~Fan, M.~Xu, L.~Zeng, and Y.~Qiao, ``{SpiderCNN: Deep Learning on
  Point Sets with Parameterized Convolutional Filters},'' in \emph{Proc. Eur.
  Conf. Comput. Vis.}, 2018.

\bibitem{su2018splatnet}
H.~Su, V.~Jampani, D.~Sun, S.~Maji, E.~Kalogerakis, M.-H. Yang, and J.~Kautz,
  ``{SPLATNet: Sparse Lattice Networks for Point Cloud Processing},'' in
  \emph{Proc. IEEE Conf. Comput. Vis. Pattern Recognit.}, 2018.

\bibitem{li2018so}
J.~Li, B.~M. Chen, and G.~H. Lee, ``{SO-Net: Self-Organizing Network for Point
  Cloud Analysis},'' in \emph{Proc. IEEE Conf. Comput. Vis. Pattern Recognit.},
  2018.

\bibitem{tchapmi2017segcloud}
L.~P. Tchapmi, C.~B. Choy, I.~Armeni, J.~Gwak, and S.~Savarese, ``{SEGCloud:
  Semantic Segmentation of 3D Point Clouds},'' in \emph{Proc. Int. Conf. 3D
  Vis.}, 2017.

\bibitem{wang2018sgpn}
W.~Wang, R.~Yu, Q.~Huang, and U.~Neumann, ``{SGPN: Similarity Group Proposal
  Network for 3D Point Cloud Instance Segmentation},'' in \emph{Proc. IEEE
  Conf. Comput. Vis. Pattern Recognit.}, 2018.

\bibitem{landrieu2018large}
L.~Landrieu and M.~Simonovsky, ``{Large-Scale Point Cloud Semantic Segmentation
  With Superpoint Graphs},'' in \emph{Proc. IEEE Conf. Comput. Vis. Pattern
  Recognit.}, 2018.

\bibitem{wang2018deep}
S.~Wang, S.~Suo, W.-C. Ma, A.~Pokrovsky, and R.~Urtasun, ``{Deep Parametric
  Continuous Convolutional Neural Networks},'' in \emph{Proc. IEEE Conf.
  Comput. Vis. Pattern Recognit.}, 2018.

\bibitem{graham20183d}
B.~Graham, M.~Engelcke, and L.~van~der Maaten, ``{3D Semantic Segmentation With
  Submanifold Sparse Convolutional Networks},'' in \emph{Proc. IEEE Conf.
  Comput. Vis. Pattern Recognit.}, 2018.

\bibitem{huang2018recurrent}
Q.~Huang, W.~Wang, and U.~Neumann, ``{Recurrent Slice Networks for 3D
  Segmentation on Point Clouds},'' in \emph{Proc. IEEE Conf. Comput. Vis.
  Pattern Recognit.}, 2018.

\bibitem{qi2018frustum}
C.~R. Qi, W.~Liu, C.~Wu, H.~Su, and L.~J. Guibas, ``{Frustum PointNets for 3D
  Object Detection from RGB-D Data},'' in \emph{Proc. IEEE Conf. Comput. Vis.
  Pattern Recognit.}, 2018.

\bibitem{shi2019pointrcnn}
S.~Shi, X.~Wang, and H.~Li, ``{PointRCNN: 3D Object Proposal Generation and
  Detection from Point Cloud},'' in \emph{Proc. IEEE Conf. Comput. Vis. Pattern
  Recognit.}, 2019.

\bibitem{hu2020rand}
Q.~Hu, B.~Yang, L.~Xie, S.~Rosa, Y.~Guo, Z.~Wang, N.~Trigoni, and A.~Markham,
  ``{RandLA-Net: Efficient Semantic Segmentation of Large-Scale Point
  Clouds},'' in \emph{Proc. IEEE Conf. Comput. Vis. Pattern Recognit.}, 2020.

\bibitem{wang2018adaptive}
P.-S. Wang, Y.~Liu, Y.-X. Guo, C.-Y. Sun, and X.~Tong, ``{Adaptive O-CNN: A
  Patch-based Deep Representation of 3D Shapes},'' in \emph{SIGGRAPH Asia},
  2018.

\bibitem{lei2019octree}
H.~Lei, N.~Akhtar, and A.~Mian, ``{Octree Guided CNN With Spherical Kernels for
  3D Point Clouds},'' in \emph{Proc. IEEE Conf. Comput. Vis. Pattern
  Recognit.}, 2019.

\bibitem{choy20194d}
C.~Choy, J.~Gwak, and S.~Savarese, ``{4D Spatio-Temporal ConvNets: Minkowski
  Convolutional Neural Networks},'' in \emph{Proc. IEEE Conf. Comput. Vis.
  Pattern Recognit.}, 2019.

\bibitem{han2015learning}
S.~Han, J.~Pool, J.~Tran, and W.~J. Dally, ``{Learning both Weights and
  Connections for Efficient Neural Networks},'' in \emph{Proc. Adv. Neural Inf.
  Process. Syst.}, 2015.

\bibitem{han2016deep}
S.~Han, H.~Mao, and W.~J. Dally, ``{Deep Compression: Compressing Deep Neural
  Networks with Pruning, Trained Quantization and Huffman Coding},'' in
  \emph{Proc. Int. Conf. Learn. Representations}, 2016.

\bibitem{he2018amc}
Y.~He, J.~Lin, Z.~Liu, H.~Wang, L.-J. Li, and S.~Han, ``{AMC: AutoML for Model
  Compression and Acceleration on Mobile Devices},'' in \emph{Proc. Eur. Conf.
  Comput. Vis.}, 2018.

\bibitem{zhou2017incremental}
A.~Zhou, A.~Yao, Y.~Guo, L.~Xu, and Y.~Chen, ``{Incremental Network
  Quantization: Towards Lossless CNNs with Low-Precision Weights},'' in
  \emph{Proc. Int. Conf. Learn. Representations}, 2017.

\bibitem{wang2019haq}
K.~Wang, Z.~Liu, Y.~Lin, J.~Lin, and S.~Han, ``{HAQ: Hardware-Aware Automated
  Quantization with Mixed Precision},'' in \emph{Proc. IEEE Conf. Comput. Vis.
  Pattern Recognit.}, 2019.

\bibitem{iandola2016squeezenet}
F.~N. Iandola, S.~Han, M.~W. Moskewicz, K.~Ashraf, W.~J. Dally, and K.~Keutzer,
  ``{SqueezeNet: AlexNet-Level Accuracy with 50x Fewer Parameters and $<$ 0.5MB
  Model Size},'' \emph{arXiv}, 2016.

\bibitem{howard2017mobilenets}
A.~G. Howard, M.~Zhu, B.~Chen, D.~Kalenichenko, W.~Wang, T.~Weyand,
  M.~Andreetto, and H.~Adam, ``{MobileNets: Efficient Convolutional Neural
  Networks for Mobile Vision Applications},'' \emph{arXiv:1704.04861}, 2017.

\bibitem{sandler2018mobilenetv2}
M.~Sandler, A.~Howard, M.~Zhu, A.~Zhmoginov, and L.-C. Chen, ``{MobileNetV2:
  Inverted Residuals and Linear Bottlenecks},'' in \emph{Proc. IEEE Conf.
  Comput. Vis. Pattern Recognit.}, 2018.

\bibitem{howard2019searching}
A.~Howard, M.~Sandler, G.~Chu, L.-C. Chen, B.~Chen, M.~Tan, W.~Wang, Y.~Zhu,
  R.~Pang, V.~Vasudevan, Q.~V. Le, and H.~Adam, ``{Searching for
  MobileNetV3},'' in \emph{Proc. IEEE Int. Conf. Comput. Vis.}, 2019.

\bibitem{zhang2018shufflenet}
X.~Zhang, X.~Zhou, M.~Lin, and J.~Sun, ``{ShuffleNet: An Extremely Efficient
  Convolutional Neural Network for Mobile Devices},'' in \emph{Proc. IEEE Conf.
  Comput. Vis. Pattern Recognit.}, 2018.

\bibitem{ma2018shufflenet}
N.~Ma, X.~Zhang, H.-T. Zheng, and J.~Sun, ``{ShuffleNet V2: Practical
  Guidelines for Efficient CNN Architecture Design},'' in \emph{Proc. Eur.
  Conf. Comput. Vis.}, 2018.

\bibitem{zoph2017neural}
B.~Zoph and Q.~V. Le, ``{Neural Architecture Search with Reinforcement
  Learning},'' in \emph{Proc. Int. Conf. Learn. Representations}, 2017.

\bibitem{zoph2018learning}
B.~Zoph, V.~Vasudevan, J.~Shlens, and Q.~V. Le, ``{Learning Transferable
  Architectures for Scalable Image Recognition},'' in \emph{Proc. IEEE Conf.
  Comput. Vis. Pattern Recognit.}, 2018.

\bibitem{liu2019progressive}
C.~Liu, B.~Zoph, M.~Neumann, J.~Shlens, W.~Hua, L.-J. Li, L.~Fei-Fei,
  A.~Yuille, J.~Huang, and K.~Murphy, ``{Progressive Neural Architecture
  Search},'' in \emph{Proc. Eur. Conf. Comput. Vis.}, 2018.

\bibitem{tan2019mnasnet}
M.~Tan, B.~Chen, R.~Pang, V.~Vasudevan, M.~Sandler, A.~Howard, and Q.~V. Le,
  ``{MnasNet: Platform-Aware Neural Architecture Search for Mobile},'' in
  \emph{Proc. IEEE Conf. Comput. Vis. Pattern Recognit.}, 2019.

\bibitem{wu2019fbnet}
B.~Wu, X.~Dai, P.~Zhang, Y.~Wang, F.~Sun, Y.~Wu, Y.~Tian, P.~Vajda, Y.~Jia, and
  K.~Keutzer, ``{FBNet: Hardware-aware Efficient Convnet Design via
  Differentiable Neural Architecture Search},'' in \emph{Proc. IEEE Conf.
  Comput. Vis. Pattern Recognit.}, 2019.

\bibitem{tan2019efficientnet}
M.~Tan and Q.~V. Le, ``{EfficientNet: Rethinking Model Scaling for
  Convolutional Neural Networks},'' in \emph{Proc. Int. Conf. Mach. Learn.},
  2019.

\bibitem{wang2020hat}
H.~Wang, Z.~Wu, Z.~Liu, H.~Cai, L.~Zhu, C.~Gan, and S.~Han, ``{HAT:
  Hardware-Aware Transformers for Efficient Natural Language Processing},'' in
  \emph{Proc. Annu. Meet. Assoc. Comput. Linguistics}, 2020.

\bibitem{strubell2019energy}
E.~Strubell, A.~Ganesh, and A.~McCallum, ``{Energy and Policy Considerations
  for Deep Learning in NLP},'' in \emph{Proc. Annu. Meet. Assoc. Comput.
  Linguistics}, 2019.

\bibitem{liu2019darts}
H.~Liu, K.~Simonyan, and Y.~Yang, ``{DARTS: Differentiable Architecture
  Search},'' in \emph{Proc. Int. Conf. Learn. Representations}, 2019.

\bibitem{cai2019proxylessnas}
H.~Cai, L.~Zhu, and S.~Han, ``{ProxylessNAS: Direct Neural Architecture Search
  on Target Task and Hardware},'' in \emph{Proc. Int. Conf. Learn.
  Representations}, 2019.

\bibitem{guo2019single}
Z.~Guo, X.~Zhang, H.~Mu, W.~Heng, Z.~Liu, Y.~Wei, and J.~Sun, ``{Single Path
  One-Shot Neural Architecture Search with Uniform Sampling},'' in \emph{Proc.
  Eur. Conf. Comput. Vis.}, 2020.

\bibitem{chen2019detnas}
Y.~Chen, T.~Yang, X.~Zhang, G.~Meng, X.~Xiao, and J.~Sun, ``{DetNAS: Backbone
  Search for Object Detection},'' in \emph{Proc. Adv. Neural Inf. Process.
  Syst.}, 2019.

\bibitem{cai2020once}
H.~Cai, C.~Gan, T.~Wang, Z.~Zhang, and S.~Han, ``{Once for All: Train One
  Network and Specialize it for Efficient Deployment},'' in \emph{Proc. Int.
  Conf. Learn. Representations}, 2020.

\bibitem{stamoulis2019single}
D.~Stamoulis, R.~Ding, D.~Wang, D.~Lymberopoulos, B.~Priyantha, J.~Liu, and
  D.~Marculescu, ``{Single-Path NAS: Designing Hardware-Efficient ConvNets in
  Less Than 4 Hours},'' \emph{arXiv:1904.02877}, 2019.

\bibitem{yang2018netadapt}
T.-J. Yang, A.~Howard, B.~Chen, X.~Zhang, A.~Go, M.~Sandler, V.~Sze, and
  H.~Adam, ``{NetAdapt: Platform-Aware Neural Network Adaptation for Mobile
  Applications},'' in \emph{Proc. Eur. Conf. Comput. Vis.}, 2018.

\bibitem{liu2019metapruning}
Z.~Liu, H.~Mu, X.~Zhang, Z.~Guo, X.~Yang, K.-T. Cheng, and J.~Sun,
  ``{MetaPruning: Meta Learning for Automatic Neural Network Channel
  Pruning},'' in \emph{Proc. IEEE Int. Conf. Comput. Vis.}, 2019.

\bibitem{cai2019automl}
H.~Cai, J.~Lin, Y.~Lin, Z.~Liu, K.~Wang, T.~Wang, L.~Zhu, and S.~Han, ``{AutoML
  for Architecting Efficient and Specialized Neural Networks},'' \emph{IEEE
  Micro}, 2019.

\bibitem{li2020gan}
M.~Li, J.~Lin, Y.~Ding, Z.~Liu, J.-Y. Zhu, and S.~Han, ``{GAN Compression:
  Efficient Architectures for Interactive Conditional GANs},'' in \emph{Proc.
  IEEE Conf. Comput. Vis. Pattern Recognit.}, 2020.

\bibitem{wang2020hardware}
K.~Wang, Z.~Liu, Y.~Lin, J.~Lin, and S.~Han, ``{Hardware-Centric AutoML for
  Mixed-Precision Quantization},'' \emph{Int. J. Comput. Vis.}, 2020.

\bibitem{wang2020apq}
T.~Wang, K.~Wang, H.~Cai, J.~Lin, Z.~Liu, H.~Wang, Y.~Lin, and S.~Han, ``{APQ:
  Joint Search for Network Architecture, Pruning and Quantization Policy},'' in
  \emph{Proc. IEEE Conf. Comput. Vis. Pattern Recognit.}, 2020.

\bibitem{radosavovic2019on}
I.~Radosavovic, J.~Johnson, S.~Xie, W.-Y. Lo, and P.~Dollar, ``{On Network
  Design Spaces for Visual Recognition},'' in \emph{Proc. IEEE Int. Conf.
  Comput. Vis.}, 2019.

\bibitem{zhu2019vnas}
Z.~Zhu, C.~Liu, D.~Yang, A.~Yuille, and D.~Xu, ``{V-NAS: Neural Architecture
  Search for Volumetric Medical Image Segmentation},'' in \emph{Proc. Int.
  Conf. 3D Vis.}, 2019.

\bibitem{kim2019scalable}
S.~Kim, I.~Kim, S.~Lim, W.~Baek, C.~Kim, H.~Cho, B.~Yoon, and T.~Kim,
  ``{Scalable Neural Architecture Search for 3D Medical Image Segmentation},''
  in \emph{Proc. Medical Image Computing and Computer Assisted Intervention},
  2019.

\bibitem{yang2019searching}
D.~Yang, H.~Roth, Z.~Xu, F.~Milletari, L.~Zhang, and D.~Xu, ``{Searching
  Learning Strategy with Reinforcement Learning for 3D Medical Image
  Segmentation},'' in \emph{Proc. Medical Image Computing and Computer Assisted
  Intervention}, 2019.

\bibitem{bae2019resource}
W.~Bae, S.~Lee, Y.~Lee, B.~Park, M.~Chung, and K.-H. Jung, ``{Resource
  Optimized Neural Architecture Search for 3D Medical Image Segmentation},'' in
  \emph{Proc. Medical Image Computing and Computer Assisted Intervention},
  2019.

\bibitem{wong2019segnas3d}
K.~C. Wong and M.~Moradi, ``{SegNAS3D: Network Architecture Search with
  Derivative-Free Global Optimization for 3D Image Segmentation},'' in
  \emph{Proc. Medical Image Computing and Computer Assisted Intervention},
  2019.

\bibitem{yu2020c2fnas}
Q.~Yu, D.~Yang, H.~Roth, Y.~Bai, Y.~Zhang, A.~Yuille, and D.~Xu, ``{C2FNAS:
  Coarse-to-Fine Neural Architecture Search for 3D Medical Image
  Segmentation},'' in \emph{Proc. IEEE Conf. Comput. Vis. Pattern Recognit.},
  2020.

\bibitem{ma2020auto}
Z.~Ma, Z.~Zhou, Y.~Liu, Y.~Lei, and H.~Yan, ``{Auto-ORVNet: Orientation-Boosted
  Volumetric Neural Architecture Search for 3D Shape Classification},''
  \emph{IEEE Access}, 2020.

\bibitem{li2020sgas}
G.~Li, G.~Qian, I.~C. Delgadillo, M.~Muller, A.~Thabet, and B.~Ghanem, ``{SGAS:
  Sequential Greedy Architecture Search},'' in \emph{Proc. IEEE Conf. Comput.
  Vis. Pattern Recognit.}, 2020.

\bibitem{ioffe2015batch}
S.~Ioffe and C.~Szegedy, ``{Batch Normalization: Accelerating Deep Network
  Training by Reducing Internal Covariate Shift},'' in \emph{Proc. Int. Conf.
  Mach. Learn.}, 2015.

\bibitem{maas2013rectifier}
A.~L. Maas, A.~Y. Hannun, and A.~Y. Ng, ``{Rectifier Nonlinearities Improve
  Neural Network Acoustic Models},'' in \emph{Proc. Int. Conf. Mach. Learn.},
  2013.

\bibitem{behley2019semantickitti}
J.~Behley, M.~Garbade, A.~Milioto, J.~Quenzel, S.~Behnke, C.~Stachniss, and
  J.~Gall, ``{SemanticKITTI: A Dataset for Semantic Scene Understanding of
  LiDAR Sequences},'' in \emph{Proc. IEEE Int. Conf. Comput. Vis.}, 2019.

\bibitem{deng2009imagenet}
J.~Deng, W.~Dong, R.~Socher, L.-J. Li, K.~Li, and L.~Fei-Fei, ``{ImageNet: A
  Large-Scale Hierarchical Image Database},'' in \emph{Proc. IEEE Conf. Comput.
  Vis. Pattern Recognit.}, 2009.

\bibitem{wu2019pointconv}
W.~Wu, Z.~Qi, and L.~Fuxin, ``{PointConv: Deep Convolutional Networks on 3D
  Point Clouds},'' in \emph{Proc. IEEE Conf. Comput. Vis. Pattern Recognit.},
  2019.

\bibitem{mo2019partnet}
K.~Mo, S.~Zhu, A.~X. Chang, L.~Yi, S.~Tripathi, L.~J. Guibas, and H.~Su,
  ``{PartNet: A Large-Scale Benchmark for Fine-Grained and Hierarchical
  Part-Level 3D Object Understanding},'' in \emph{Proc. IEEE Conf. Comput. Vis.
  Pattern Recognit.}, 2019.

\bibitem{armeni20163d}
I.~Armeni, O.~Sener, A.~R. Zamir, H.~Jiang, I.~Brilakis, M.~Fischer, and
  S.~Savarese, ``{3D Semantic Parsing of Large-Scale Indoor Spaces},'' in
  \emph{Proc. IEEE Conf. Comput. Vis. Pattern Recognit.}, 2016.

\bibitem{armeni2017joint}
I.~Armeni, A.~Sax, A.~R. Zamir, and S.~Savarese, ``{Joint 2D-3D-Semantic Data
  for Indoor Scene Understanding},'' \emph{arXiv:1702.01105}, 2017.

\bibitem{caesar2020nuscenes}
H.~Caesar, V.~Bankiti, A.~H. Lang, S.~Vora, V.~E. Liong, Q.~Xu, A.~Krishnan,
  Y.~Pan, G.~Baldan, and O.~Beijbom, ``{nuScenes: A Multimodal Dataset for
  Autonomous Driving},'' in \emph{Proc. IEEE Conf. Comput. Vis. Pattern
  Recognit.}, 2020.

\bibitem{geiger2012kitti}
A.~Geiger, P.~Lenz, and R.~Urtasun, ``{Are we ready for Autonomous Driving? The
  KITTI Vision Benchmark Suite},'' in \emph{Proc. IEEE Conf. Comput. Vis.
  Pattern Recognit.}, 2012.

\bibitem{le2020going}
E.-T. Le, I.~Kokkinos, and N.~J. Mitra, ``{Going Deeper with Lean Point
  Networks},'' in \emph{Proc. IEEE Conf. Comput. Vis. Pattern Recognit.}, 2020.

\bibitem{li2019deepgcns}
G.~Li, M.~Müller, G.~Qian, I.~C. Delgadillo, A.~Abualshour, A.~Thabet, and
  B.~Ghanem, ``{DeepGCNs: Can GCNs Go as Deep as CNNs?}'' in \emph{Proc. IEEE
  Int. Conf. Comput. Vis.}, 2019.

\bibitem{li2019sgas}
G.~Li, G.~Qian, I.~C. Delgadillo, M.~M{\"u}ller, A.~Thabet, and B.~Ghanem,
  ``{SGAS: Sequential Greedy Architecture Search},'' in \emph{Proc. IEEE Conf.
  Comput. Vis. Pattern Recognit.}, 2020.

\bibitem{li2020lcnas}
G.~Li, M.~Xu, S.~Giancola, A.~Thabet, and B.~Ghanem, ``{LC-NAS: Latency
  Constrained Neural Architecture Search for Point Cloud Networks},''
  \emph{arXiv:2008.10309}, 2020.

\bibitem{hu2019randla}
Q.~Hu, B.~Yang, L.~Xie, S.~Rosa, Y.~Guo, Z.~Wang, N.~Trigoni, and A.~Markham,
  ``{RandLA-Net: Efficient Semantic Segmentation of Large-Scale Point
  Clouds},'' in \emph{Proc. IEEE Conf. Comput. Vis. Pattern Recognit.}, 2020.

\bibitem{thomas2019kpconv}
H.~Thomas, C.~R. Qi, J.-E. Deschaud, B.~Marcotegui, F.~Goulette, and L.~J.
  Guibas, ``{KPConv: Flexible and Deformable Convolution for Point Clouds},''
  in \emph{Proc. IEEE Int. Conf. Comput. Vis.}, 2019.

\bibitem{xu2020squeezesegv3}
C.~Xu, B.~Wu, Z.~Wang, W.~Zhan, P.~Vajda, K.~Keutzer, and M.~Tomizuka,
  ``{SqueezeSegV3: Spatially-Adaptive Convolution for Efficient Point-Cloud
  Segmentation},'' in \emph{Proc. Eur. Conf. Comput. Vis.}, 2020.

\bibitem{alonso20203d}
I.~Alonso, L.~Riazuelo, L.~Montesano, and A.~C. Murillo, ``{3D-MiniNet:
  Learning a 2D Representation from Point Clouds for Fast and Efficient 3D
  LIDAR Semantic Segmentation},'' in \emph{Proc. IEEE/RSJ Int. Conf. Intell.
  Robot. Syst.}, 2020.

\bibitem{zhang2020polarnet}
Y.~Zhang, Z.~Zhou, P.~David, X.~Yue, Z.~Xi, B.~Gong, and H.~Foroosh,
  ``{PolarNet: An Improved Grid Representation for Online LiDAR Point Clouds
  Semantic Segmentation},'' in \emph{Proc. IEEE Conf. Comput. Vis. Pattern
  Recognit.}, 2020.

\bibitem{cortinhal2020salsanext}
T.~Cortinhal, G.~Tzelepis, and E.~E. Aksoy, ``{SalsaNext: Fast,
  Uncertainty-aware Semantic Segmentation of LiDAR Point Clouds for Autonomous
  Driving},'' \emph{arXiv:2003.03653}, 2020.

\bibitem{geiger2013vision}
A.~Geiger, P.~Lenz, C.~Stiller, and R.~Urtasun, ``{Vision meets Robotics: The
  KITTI Dataset},'' \emph{Int. J. Robot. Res.}, 2013.

\bibitem{yan2018second}
Y.~Yan, Y.~Mao, and B.~Li, ``{SECOND: Sparsely Embedded Convolutional
  Detection},'' \emph{Sensors (Basel)}, 2018.

\end{thebibliography}

\begin{IEEEbiography}[{\includegraphics[width=1in,height=1.25in,clip,keepaspectratio]{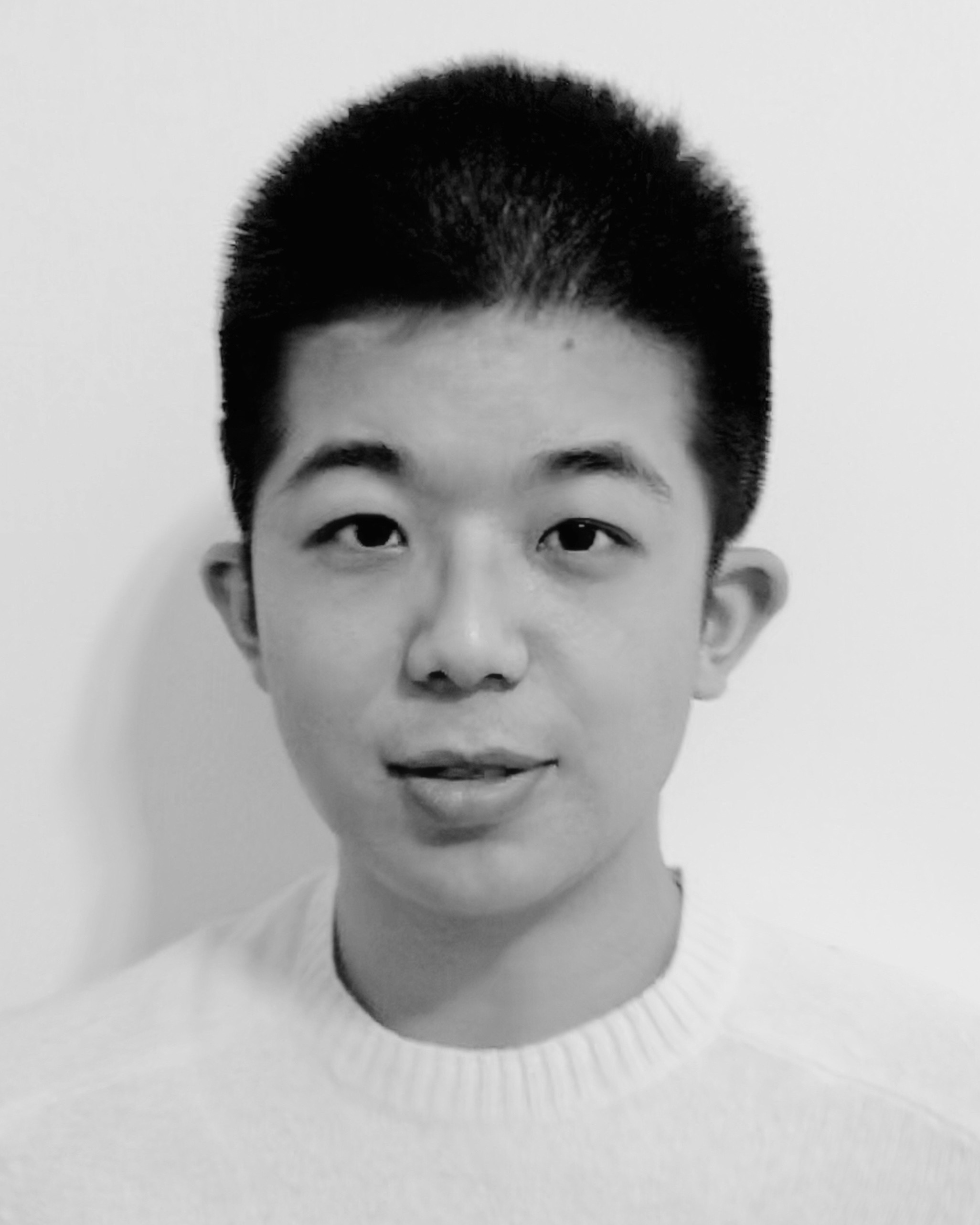}}]{Zhijian Liu}
received the B.Eng. degree in computer science from Shanghai Jiao Tong University, China, in 2018, and the S.M. degree in electrical engineering and computer science from MIT, in 2020. He is working toward the Ph.D. degree at MIT, under the supervision of Prof. Song Han. His research focuses on efficient deep learning and its applications in computer vision and robotics.
\end{IEEEbiography}

\begin{IEEEbiography}[{\includegraphics[width=1in,height=1.25in,clip,keepaspectratio]{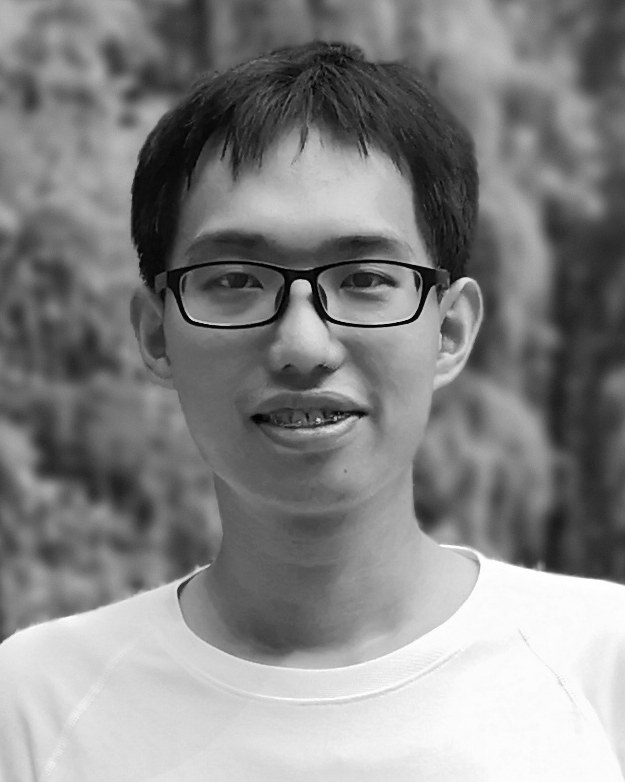}}]{Haotian Tang}
is a first-year Ph.D. student at MIT EECS, advised by Prof. Song Han. Previously, he received his B.Eng. degree from Department of Computer Science and Engineering, Shanghai Jiao Tong University, China, in 2020. His research interest is co-designing efficient machine learning algorithms and systems.
\end{IEEEbiography}

\begin{IEEEbiography}[{\includegraphics[width=1in,height=1.25in,clip,keepaspectratio]{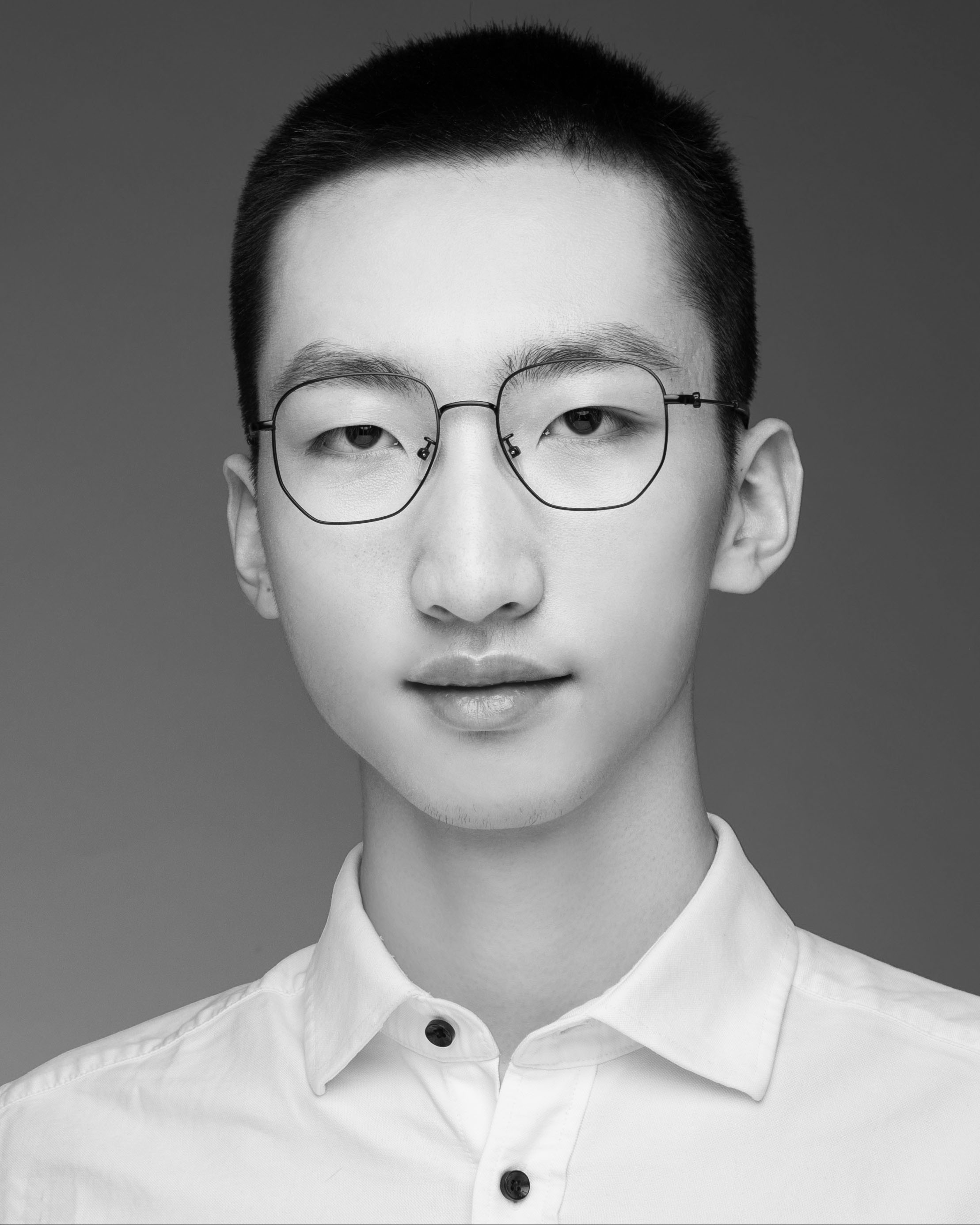}}]{Shengyu Zhao}
is an undergraduate student at Institute for Interdisciplinary Information Sciences, Tsinghua University. He was a visiting researcher at MIT under the supervision of Prof. Song Han. His research focuses on efficient deep learning, autonomous driving, and generative models.
\end{IEEEbiography}

\begin{IEEEbiography}[{\includegraphics[width=1in,height=1.25in,clip,keepaspectratio]{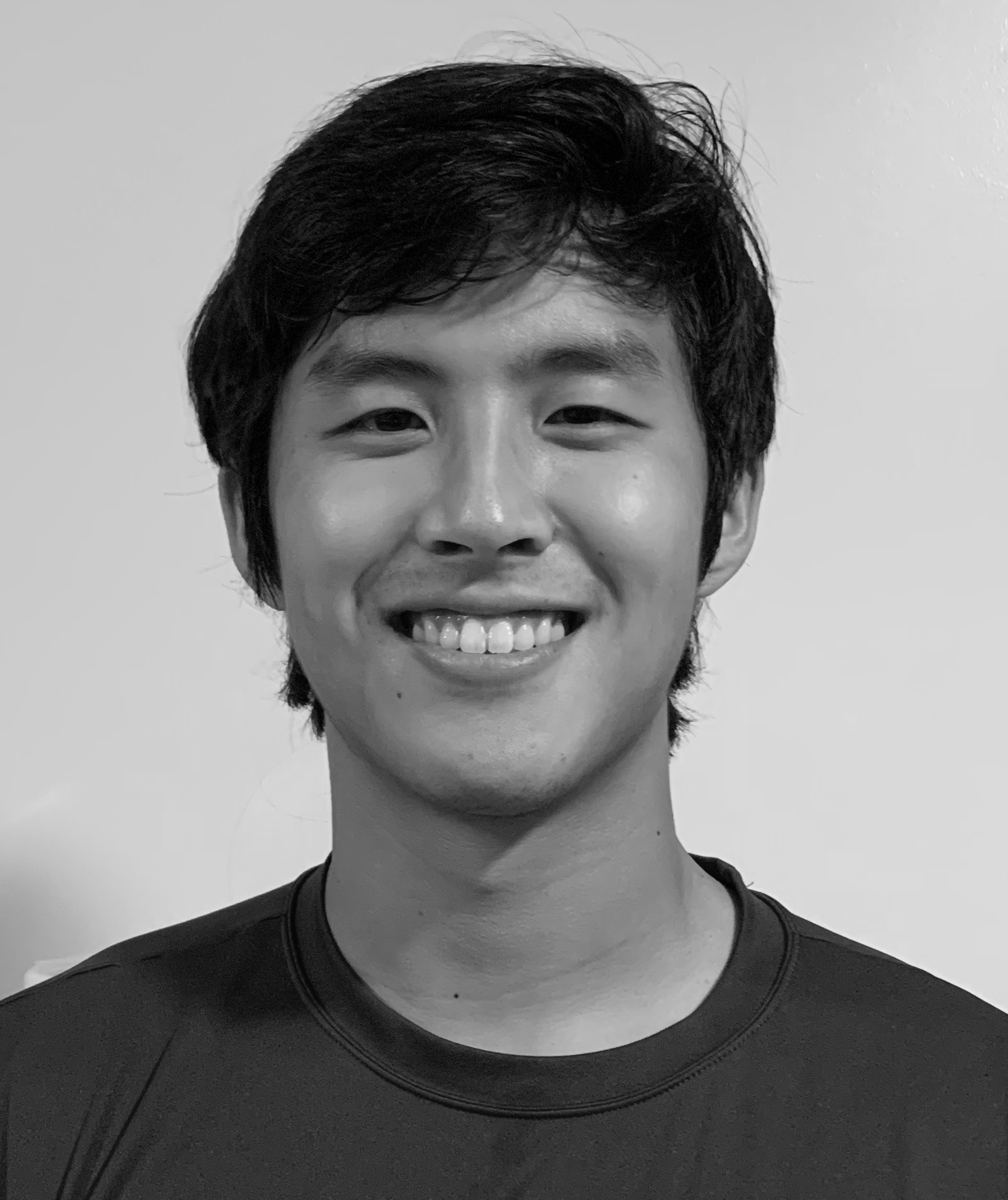}}]{Kevin Shao}
is a second-year undergraduate student at MIT EECS. He is an undergraduate researcher working under Prof. Song Han, and his research interest is computer vision for autonomous vehicles.
\end{IEEEbiography}

\begin{IEEEbiography}[{\includegraphics[width=1in,height=1.25in,clip,keepaspectratio]{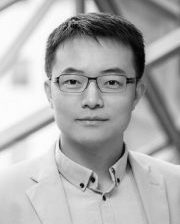}}]{Song Han}
is an assistant professor at MIT EECS Department. Dr. Han received the Ph.D. degree in Electrical Engineering from Stanford University and B.S. degree in Electrical Engineering from Tsinghua University. Dr. Han's research focuses on efficient deep learning computing at the intersection between machine learning and computer architecture. He proposed ``Deep Compression'' and the ``Efficient Inference Engine'' that impacted the industry. He is a recipient of NSF CAREER Award, MIT Technology Review Innovators Under 35, best paper awards at ICLR 2016 and FPGA 2017, Facebook Faculty Award, SONY Faculty Award, AWS Machine Learning Research Award.
\end{IEEEbiography}

\end{document}